\documentclass[10pt,twocolumn,letterpaper]{article}

\usepackage{wacv}              

\definecolor{wacvblue}{rgb}{0.21,0.49,0.74}
\usepackage[pagebackref,breaklinks,colorlinks,allcolors=wacvblue]{hyperref}

\def\wacvPaperID{2623} 
\def\confName{WACV}
\def\confYear{2027}

\usepackage{graphicx}
\usepackage{amsmath,amssymb, bm}
\usepackage{booktabs}
\usepackage{multirow}
\usepackage{xcolor}
\usepackage{algorithm}
\usepackage{algpseudocode}

\usepackage{subcaption}
\usepackage{tabularx}
\usepackage{wrapfig}
\usepackage{rotating}
\usepackage{colortbl}

\newcommand{\method}{PLAP-LCA}
\newcommand{\plap}{PLAP}

\newcommand{\lca}{LCA}

\makeatletter\long\def\revised#1{{\color{black}#1}}\makeatother
\usepackage[symbol]{footmisc}
\title{Lighting-aware Unified Model for Instance Segmentation}

\author{
Qisai Liu\thanks{These authors contributed equally.}\qquad
Alloy Das\footnotemark[1]\qquad
Zhanhong Jiang\qquad
Joshua R. Waite\\
Aditya Balu\qquad
Adarsh Krishnamurthy\qquad
Soumik Sarkar\\[3pt]
Iowa State University\\[3pt]
{\tt\footnotesize
\{supersai, alloydas, zhjiang, jrwaite, baditya, adarsh, soumiks\}@iastate.edu}
}

\newcommand{\cref}[3]{\hyperref[#2]{#1~\ref*{#2}{#3}}}
\newcommand{\colref}[2]{\hyperref[#2]{#1~\ref*{#2}}}
\newcommand{\algoref}[1]{\colref{Algorithm}{#1}}

\newcommand{\figref}[1]{\colref{Fig.}{#1}}

\newcommand{\secref}[1]{\colref{Sec.}{#1}}
\newcommand{\tableref}[1]{\colref{Table}{#1}}
\newcommand{\coloredref}[2]{\hyperref[#2]{#1~\ref*{#2}}}
\newcommand{\coloredsubref}[3]{\hyperref[#2]{#1~\ref*{#2}{#3}}}

\begin{document}
\maketitle
\begin{abstract}
Foundation models such as the Segment Anything Model (SAM) generalize well but degrade under
real-world illumination changes. We address this with Lighting Convolutional Attention (LCA), a lightweight attention module that equips a frozen encoder with trainable illumination-aware feature adaptation. LCA combines a channel, a spatial, and a fixed Laplacian contrast gate by
conjunctive multiplicative fusion, so a feature survives only where global, local, and physical
edge cues agree. We train it with paired clean and re-lit images under a prediction
consistency loss, and release a Unity-based synthetic dataset with physically accurate
lighting. We study prompted mask prediction under illumination change, where
every mask is decoded from a ground-truth prompt. Across COCO, VOC, Cityscapes, Unity, and real
adverse-weather imagery from ACDC, LCA improves over its frozen backbone on both SAM and
SAM~3, improves robustness to illumination changes, and transfers to real adverse conditions with no
target-domain training. 
\end{abstract}

\section{Introduction}
\label{sec:intro}

Instance segmentation has achieved remarkable progress through foundation models such as the Segment Anything Model (SAM)~\cite{kirillov2023segment}, which demonstrates impressive zero-shot generalization across diverse visual domains. However, because these models are predominantly trained on well-lit and high-quality imagery, they frequently suffer significant performance degradation when deployed in environments with adverse illumination. This creates a critical gap in real-world applications such as nighttime driving, indoor low-light scenes, overexposed outdoor settings, and mixed artificial and natural lighting conditions.

In safety-critical settings such as autonomous driving~\cite{cordts2016cityscapes}, robotic manipulation and medical imaging~\cite{chankhachon2025deep, chowdhury2024deep}, illumination varies continuously and cannot be controlled. A model that segments accurately at noon but fails at dusk is difficult to deploy reliably, yet collecting and annotating data across diverse lighting conditions is costly.

We argue that addressing lighting robustness requires intervention at three complementary levels. The first is the \textit{data level}, where training images must systematically span the complex space of realistic lighting conditions. The second is the \textit{architectural level}, where the model must possess explicit mechanisms for reasoning about structural integrity independent of illumination. The third is the \textit{training level}, where the learning objective must explicitly force the extraction of lighting-invariant representations.

\begin{figure}[h]
    \centering
    \includegraphics[width=1\linewidth]{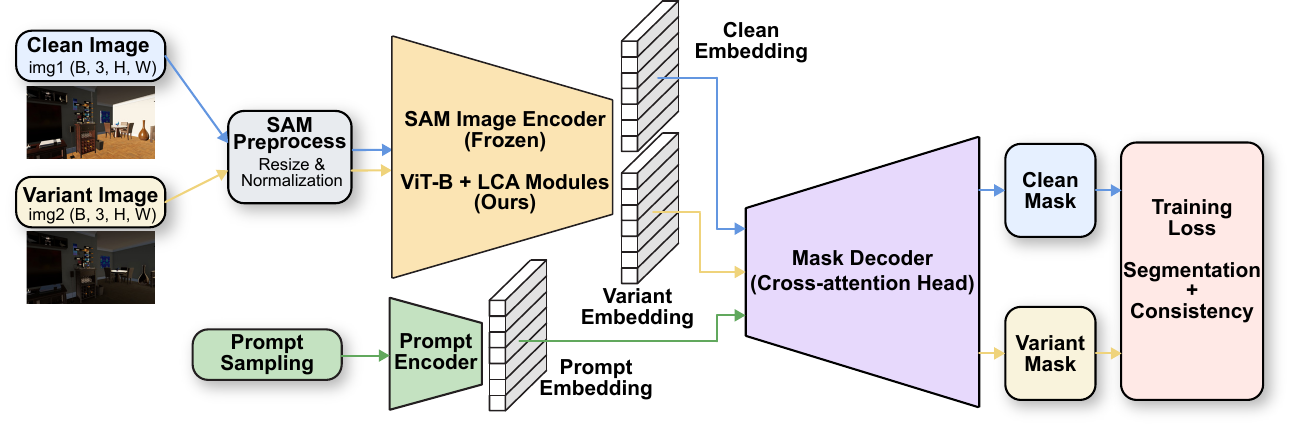}
    \caption{\revised{PLAP-LCA. Clean and variant images share a frozen SAM's image encoder with LCA modules on the final four blocks. Only the LCA weights, the scalar gates and the mask decoder are trained, while the encoder stays frozen.}}
    \label{fig:training_pipeline}
\end{figure}

\noindent\textbf{Contributions.} To this end, we propose \textit{Pairwise Lighting Augmentation Pipe\-line with Lighting Convolutional Attention (\method{})}, a framework that equ\-ips the Segment Anything Model (SAM) with robust lighting invariance without requiring computationally prohibitive fine-tuning of its heavy backbone, as shown in Figure~\ref{fig:training_pipeline}. At the core of our approach is the \lca{} module, \revised{inserted into} SAM's frozen encoder. \revised{\lca{} employs a three-gate architecture including channel, spatial, and Laplacian contrast whose conjunctive multiplicative fusion preserves features only when global, local, and physical edge cues all agree. We train with a dual-stream pairwise strategy using standard segmentation supervision alongside a targeted \textit{prediction consistency loss} that penalizes mask discrepancies between clean and variant image streams.}
This strategy is evaluated on both real-world public benchmarks and a new synthetic dataset generated from a Unity environment, covering a wide range of physically accurate illumination.
In summary, our main contributions are as follows:
(1) We introduce the \lca{} module, \revised{a lightweight three-gate attention module that equips
    foundation models with illumination-robust structural representations.} (2) We propose \revised{a severity sampling pairwise training strategy} and a novel \revised{prediction
    consistency loss} to explicitly bridge the representational gap between ideal and
    adverse lighting conditions.
    (3) We present a comprehensive empirical study on lighting vulnerabilities, supported by a novel Unity-based synthetic dataset for physically accurate illumination evaluation as well as three diverse public benchmarks, namely COCO~\cite{lin2014microsoft}, VOC~\cite{everingham2015pascal}, and Cityscapes~\cite{cordts2016cityscapes}, demonstrating consistent improvements across all illumination conditions.


\section{Related Work}
\label{sec:related}
In this section, we review primarily related work on foundation models for segmentation and refer interested readers to \secref{additional_related_work} for attention mechanisms and data augmentation.

\noindent\textbf{Segmentation Models.}
The Segment Anything Model (SAM)~\cite{kirillov2023segment} represents a paradigm shift in image segmentation, leveraging a dataset of over 1 billion masks to achieve strong zero-shot generalization through a promptable architecture. SAM's core architecture, consisting of a transformer image encoder, a lightweight prompt encoder, and a lightweight mask decoder, is highly effective. Subsequent work has explored SAM adaptation for specialized domains. For instance, SAM-Adapter~\cite{chen2023sam} incorporates task-specific tuning, while HQ-SAM~\cite{ke2023segment} introduces an additional token to enhance mask boundary details. More recently, the community has expanded SAM's capabilities. EfficientSAM~\cite{xiong2024efficientsam} drastically reduces computational overhead via masked image pretraining, PerSAM~\cite{zhang2023personalize} enables training-free, one-shot personalization for specific visual concepts, and SAM~2~\cite{ravi2024sam} extends the foundation model into real-time video segmentation using a streaming memory architecture.
However, despite these rapid architectural evolutions, existing adaptations primarily optimize for computational efficiency, semantic tracking, or boundary precision under standard conditions. They leave the foundation model vulnerable to low-level physical degradations like extreme lighting. Our work complements these efforts by addressing the distinct challenge of illumination invariance, introducing a lightweight adapter that secures lighting resilience without updating SAM's pre-trained weights. \revised{Methodologically \lca{} belongs to the parameter-efficient adaptation family, alongside adapters~\cite{houlsby2019parameter}, low-rank updates~\cite{hu2022lora} and prompt tuning~\cite{jia2022visual}. It differs from them in what the inserted module computes. Those methods insert a task-generic bottleneck. \lca{} instead applies a fixed physical contrast prior chosen for the degradation being corrected. The closest work in objective is low-light instance segmentation~\cite{chen2023instance}, which we contrast with in \secref{sec:experiments}.}

\noindent\textbf{Robustness to Environmental Conditions.}
Segmentation under adverse conditions has been studied mainly in autonomous driving, through unsupervised domain adaptation~\cite{hoyer2022daformer}, style transfer~\cite{huang2017arbitrary} and night-specific architectures~\cite{wu2021dannet}.  These approaches retrain or adapt a full segmentation network for a target domain. Beyond driving, multi-domain training has likewise been used to make scene text spotting robust to noisy capture, including underwater light attenuation and color scatter~\cite{das2024diving}. We instead keep a general-purpose foundation model frozen and correct its features for illumination at inference cost of a few percent. LLNet~\cite{LORE2017650} enhances and denoises low-light images, whereas LCA adapts segmentation features.

\section{Dataset Generation}
\label{sec:dataset_generation}
A main contribution of our approach is that segmentation models can be trained to be lighting-invariant, with structured pairwise training on clean and lighting-variant images that share identical ground-truth. This requires robust methods for generating realistic lighting variants. We employ two complementary strategies. The first is image-space augmentation by the Pairwise Lighting Augmentation Pipeline (\plap{}), which is applicable to any existing dataset, and (2)~physically rendered synthetic data generated within the Unity engine, which provides complex illumination variations with pixel-level perfect annotations. Both strategies develop the key insight that lighting variations do not change object geometry masks, bounding boxes, and category labels remain valid across all different illumination conditions. The clean-variant pairs share annotations and provide a direct training signal for lighting invariance.

\subsection{Image-Space Augmentation: \plap{}}
\label{sec:plap}

\plap{} generates paired image sets $(I_{\text{clean}}, I_{\text{variant}})$ by applying controlled lighting variant techniques to existing benchmark images (COCO, VOC, Cityscapes). This still preserves all instance-level annotations. The pipeline is defined by three design elements, namely physically motivated operations, conflict-aware sampling, and structured severity levels.
\plap{} implements multiple (in this work we select nine to cover generic scenarios) lighting operations, each corresponding to physical models of image formation or sensor response, namely \textit{exposure, directional shadow, color temperature (warm/cool), vignetting, contrast, Gamma correction, brightness, film grain, and atmospheric haze}. Please see \secref{image_space_operations} for more details about the definitions of these operations.

\noindent\textbf{Conflict-Aware Sampling.}
\label{sec:conflict}
As noted above, certain operation pairs are contradictory or redundant. We encode these as conflict groups:
\begin{equation}
\mathcal{C} = \big\{\{\text{warm}, \text{cool}\},\;\{\text{exposure}, \text{brightness}\},\;\{\text{haze}, \text{contrast}\}\big\}
\end{equation}
Once an operation is selected, everything else in its conflict group is removed from the candidate pool for that image. In the sampling stage, this helps ensure variants remain within the manifold of realistic lighting conditions.

\noindent\textbf{Three-Level Severity Protocol.}
\label{sec:severity}
Rather than treating augmentation strength as a single continuous factor, we discretize it into three severity tiers. Each tier constrains both the parameter range of individual operations and the maximum number of operations that can be composed. Severity~1 applies a single subtle change that is barely visible on casual inspection. Severity~2 allows up to two operations with wider parameter ranges, producing clearly noticeable lighting variants. Severity~3 pushes up to three simultaneous operations into extreme, which is the kind of lighting you encounter in a parking garage at night or in an overexposed outdoor scene at midday. Please see \tableref{tab:severity} in \secref{severity_levels} for more details.

\subsection{Unity-Lighting Synthetic Dataset}
\label{sec:unity_generation}
\begin{figure*}[ht]
  \centering
  \includegraphics[width=0.75\textwidth]{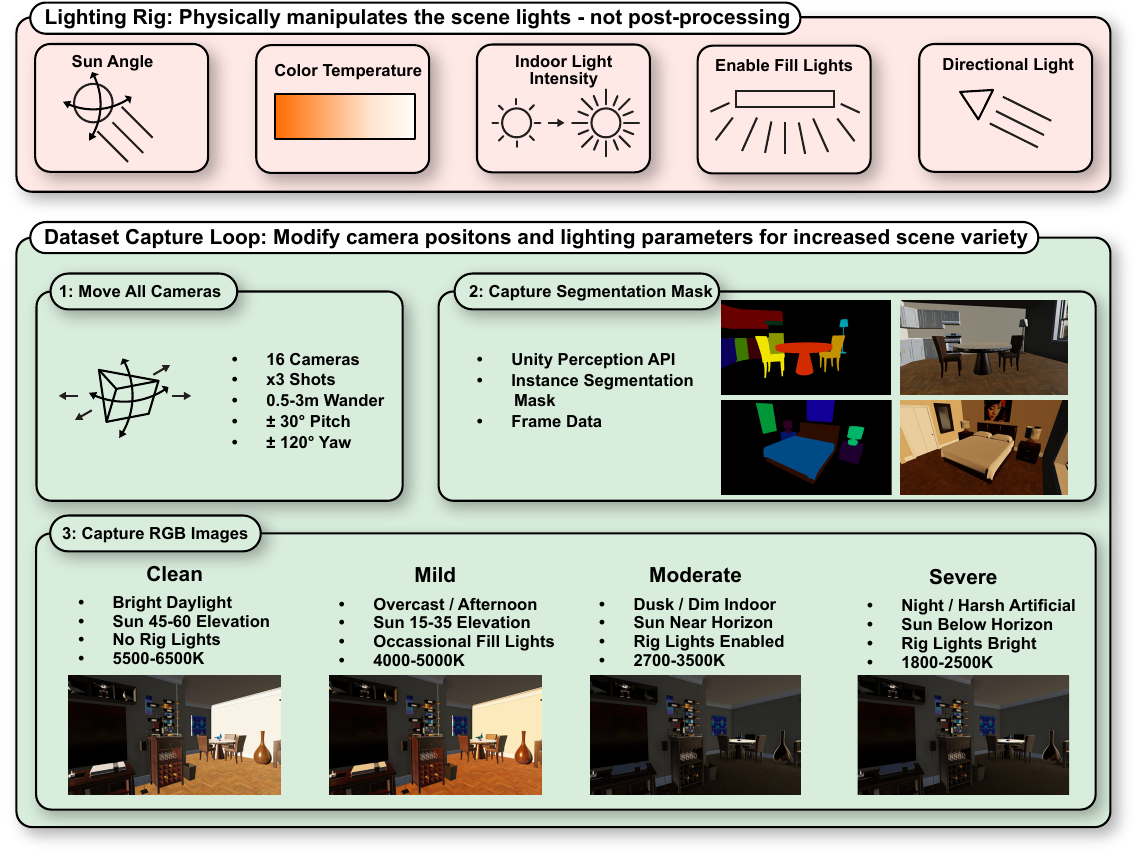}
  \caption{%
    \textbf{Unity synthetic data pipeline overview.}
    \emph{Top:}~The lighting rig exposes five physically-grounded
    parameters manipulated directly in the 3D scene.
    \emph{Middle:}~Each iteration randomises 16 camera positions  then records instance-segmentation masks.
    \emph{Bottom:}~Four available lighting settings are illustrated. Each training pair contains a clean image and one randomly selected lighting variant with shared ground-truth masks.}
  \label{fig:unity-pipeline}
\end{figure*}

While \plap{} operates in 2D image space and can augment any existing dataset, it fundamentally cannot reproduce illumination that simulates the actual 3D light transport. In Image-space operations, it cannot reproduce complex physical phenomena like sunlight reflections, the interplay of mixed natural and artificial illuminants, or light refracting through windows. To capture these physically grounded effects, we construct a complementary synthetic dataset directly inside the Unity engine. The synthetic data pipeline is shown in \figref{fig:unity-pipeline}. Please see \secref{unity_dataset} for the property summary and the specific generation method of the dataset.

\noindent
\section{Methodology}
\label{sec:method}
Our framework consists of three components. The first is a pairwise lighting augmentation pipeline for generating training pairs, \plap{}, which was detailed in \secref{sec:dataset_generation}. The second is an~\lca{} module for illumination-aware feature processing. The third is a pairwise training strategy driven by a novel \revised{prediction consistency loss} function.

\subsection{Lighting Convolutional Attention}
\label{sec:lsa}

\begin{figure}[ht]
    \centering
    \includegraphics[width=\linewidth]{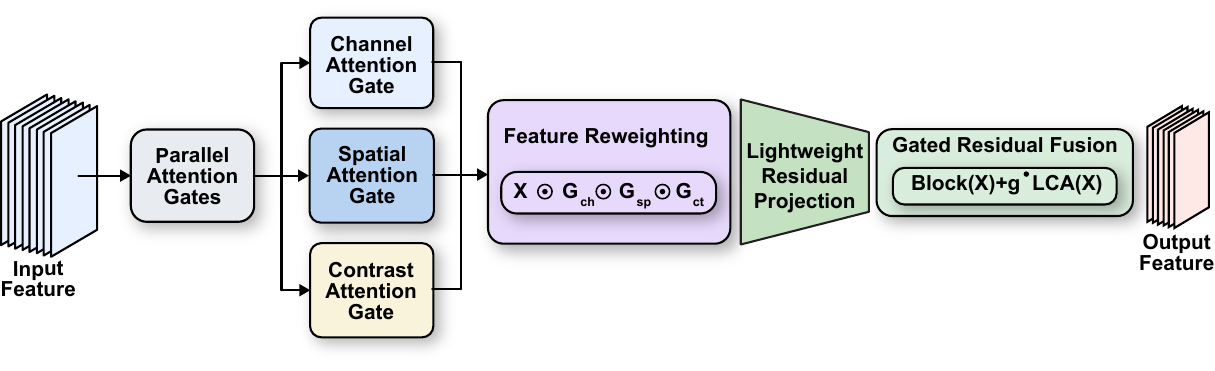}
    \caption{Lighting Convolutional Attention (LCA) module. Channel, spatial, and Laplacian contrast gates are multiplied element-wise and projected separable convolution, then fused through a learned scalar gate before the residual addition.}
    \label{fig:lca_arch}
\end{figure}

The \lca{} module (as shown in \figref{fig:lca_arch}) is designed to inject illumination awareness into SAM's frozen image encoder. Rather than fine-tuning the full encoder, we attach a lightweight \lca{} module to the last multiple transformer blocks. This framework also introduces a set of additional trainable parameters dedicated to a lighting-sensitive feature. 
Before going into the full details of the architecture, we first motivate its specific formulation. Our design comes from the observation that illumination changes corrupt feature maps across three distinct dimensions. \textbf{Globally}, uniform shifts in brightness or color temperature bias entire feature channels. \textbf{Locally}, spatial effects like cast shadows and specular highlights disrupt specific regions. \textbf{Structurally}, degraded edge contrast obscures the true physical boundaries of objects.

To address each of these dimensions, the \lca{} module employs three parallel attention gates. \revised{While the channel and spatial gates share lineage with SE-Net~\cite{hu2018squeeze} and CBAM~\cite{woo2018cbam} respectively, \lca{}'s architectural commitment is their \emph{conjunctive multiplicative fusion}. A feature is preserved only when global channel reweighting, spatial localisation, \emph{and} a fixed physical Laplacian edge prior all agree simultaneously. This differs fundamentally from the additive or sequential stacking in CBAM/SE, where no single cue can override the others, and the contrast gate incorporates a fixed Laplacian kernel as a structural prior alongside the learned channel and spatial gates. \tableref{tab:lsa_version_ablation} compares the overall performance of design variants with different fusion strategies, projections, and parameter counts, while \tableref{tab:gate_ablation} evaluates the contribution of each gate within the chosen architecture.}

The three gates operate simultaneously on the original block input $\mathbf{x} \in \mathbb{R}^{B \times C \times H \times W}$ ($B$: batch size, $C$: number of channels, $H$: height, $W$: width): a \textbf{channel attention gate} $\mathbf{G}_{\text{ch}}: \mathbb{R}^{B \times C \times H \times W} \to \mathbb{R}^{B \times C \times 1 \times 1}$ that recalibrates feature channels to counteract global illumination bias, a \textbf{spatial attention gate} $\mathbf{G}_{\text{sp}}: \mathbb{R}^{B \times C \times H \times W} \to \mathbb{R}^{B \times 1 \times H \times W}$ that identifies locally affected regions, and a \textbf{contrast attention gate} $\mathbf{G}_{\text{ct}}: \mathbb{R}^{B \times C \times H \times W} \to \mathbb{R}^{B \times 1 \times H \times W}$ based on a Laplacian edge detector that reinforces degraded structural boundaries. The three gates are fused together and projected through a residual path, yielding the complete module:
\begin{equation}
\label{eq:lca_full}
\begin{aligned}
\tilde{\mathbf{x}} &= \mathbf{x} \odot \mathbf{G}_{\text{ch}}(\mathbf{x}) \odot
  \mathbf{G}_{\text{sp}}(\mathbf{x}) \odot \mathbf{G}_{\text{ct}}(\mathbf{x}), \\[2pt]
\Phi(\mathbf{x}) &= \text{PW}\!\left(\text{ReLU}\!\left(\text{GN}\!\left(
  \text{DW}(\tilde{\mathbf{x}})\right)\right)\right),
\end{aligned}
\end{equation}
where $\Phi(\mathbf{x}):\mathbb R^{B \times C \times H \times W}\to\mathbb R^{B \times C' \times H \times W}$ denotes the output feature map.\\
$\text{DW}:\mathbb R^{B \times C \times H \times W}\to\mathbb R^{B \times C \times H \times W}$ and $\text{PW}:\mathbb R^{B \times C \times H \times W}\to\mathbb R^{B \times C' \times H \times W}$ denote depthwise and pointwise convolutions, respectively. The depthwise convolution performs spatial filtering independently for each channel, while the pointwise ($1\times 1$) convolution mixes information across channels. Together they form a lightweight depthwise-separable convolution used for efficient feature projection. GN here is the group normalization to stabilize the training.

\noindent\textbf{Channel Attention Gate.}
\label{sec:channel_gate}
Illumination shifts affect some feature channels more than others. Channels encoding brightness or colour temperature fluctuate strongly, while those capturing shape or texture stay comparatively stable. The channel gate $\mathbf{G}_{\text{ch}}$ separates the two.
We first summarize each channel using two complementary statistics. An average captures the typical activation energy across the spatial extent, and a maximum that captures the strongest localized response. Both are obtained directly from global adaptive pooling operations:
$
\mathbf{d}_{\text{avg}} = \text{AvgPool}(\mathbf{x}), 
\mathbf{d}_{\text{max}} = \text{MaxPool}(\mathbf{x}) 
\in \mathbb{R}^{B \times C \times 1 \times 1}.
$
%

\noindent\textbf{Spatial Attention Gate.}
\label{sec:spatial_gate}
Cast shadows, specular spots and illumination gradients are spatially localised, so the spatial gate asks \emph{where} to adapt.
We condense the channel dimension into two spatial summary maps, a mean and a max across all $C$ features at each location:
$
\mathbf{s}_{\text{avg}} = \text{Mean}(\mathbf{x}, \text{dim}{=}1),
\mathbf{s}_{\text{max}} = \text{Max}(\mathbf{x}, \text{dim}{=}1)
\in \mathbb{R}^{B \times 1 \times H \times W}
$.
These are concatenated into a 2-channel tensor and passed through a single convolution with a deliberately large $7{\times}7$ kernel:
$
\mathbf{G}_{\text{sp}}(\mathbf{x}) = \sigma\!\left(\text{Conv}_{7 \times 7}\!\left([\mathbf{s}_{\text{avg}} \,;\, \mathbf{s}_{\text{max}}]\right)\right)
$
The $7{\times}7$ receptive field matches the spatial scale of shadow boundaries, penumbra gradients and highlight falloffs in the feature map at this scale.

\noindent\textbf{Contrast Attention Gate.}
\label{sec:contrast_gate}
Neither of the first two gates models the physical signature of lighting variation, which is edges. Illumination changes are most visible at boundaries, where shadows and highlights alter local contrast while object structure is unchanged.
The gate operates in three stages. First, a learned $1{\times}1$ convolution projects the $C$-dimensional feature space down to a single-channel intensity map:
$
\label{eq:ct_gray}
\mathbf{g} = \text{Conv}_{1 \times 1}^{\text{proj}}(\mathbf{x}) \in \mathbb{R}^{B \times 1 \times H \times W}
$.
The input is a 768-dimensional deep feature map, so this projection learns which combination of channels best exposes the structure that lighting disturbs.
Next, we apply a fixed Laplacian kernel~\cite{kondor2016multiscale} to detect edges in this intensity map:
\begin{equation}
\label{eq:ct_lap}
\mathbf{e} = \mathbf{g} * \mathbf{K}_{\text{lap}}, \quad
\mathbf{K}_{\text{lap}} = \begin{bmatrix} 0 & 1 & 0 \\ 1 & -4 & 1 \\ 0 & 1 & 0 \end{bmatrix}
\end{equation}
The Laplacian approximates $\nabla^2 \mathbf{g}$ and responds wherever intensity changes rapidly. We keep it fixed as a non-trainable buffer, since it is already an optimal isotropic edge detector and freezing it guarantees the gate stays a physical prior rather than drifting into another learned filter. The learned projection and refinement layers adapt this fixed structural prior to the encoder's feature space. Finally, the raw Laplacian response, which can be negative and has arbitrary magnitude, is normalized per sample and refined by a small learnable convolution:
\begin{equation}
\label{eq:ct_gate}
\hat{\mathbf{e}} = \frac{\mathbf{e} - \min(\mathbf{e})}{\max(\mathbf{e}) - \min(\mathbf{e}) + \epsilon}, \quad
\mathbf{G}_{\text{ct}}(\mathbf{x}) = \sigma\!\left(\text{Conv}_{3 \times 3}^{\text{refine}}(\hat{\mathbf{e}})\right)
\end{equation}
where min and max operate over the spatial dimensions of each sample independently, and $\epsilon = 10^{-6}$. Per-sample normalization ensures consistent input range regardless of absolute edge magnitude. The $3{\times}3$ refinement (10 parameters) allows the gate to sharpen relevant edges and suppress noise, making this an \textit{adaptive} edge detector rather than a purely classical one.

\noindent\textbf{Gate Fusion.}
\label{sec:fusion}
The three gates are combined via element-wise multiplication:
\begin{equation}
\label{eq:fusion}
\mathbf{x}_{\text{att}} = \mathbf{x} \odot \underbrace{\mathbf{G}_{\text{ch}}(\mathbf{x})}_{(B,C,1,1)} \odot \underbrace{\mathbf{G}_{\text{sp}}(\mathbf{x})}_{(B,1,H,W)} \odot \underbrace{\mathbf{G}_{\text{ct}}(\mathbf{x})}_{(B,1,H,W)}
\end{equation}
Broadcasting handles the shape mismatch naturally. The channel gate scales features globally per channel, while the spatial and contrast gates modulate each location uniformly across channels. Multiplication means that if \textit{any} gate outputs a near-zero value, the corresponding feature is suppressed, regardless of how strongly the other gates activate. In other words, a feature survives only when the channel \textit{is} important, the location \textit{is} relevant, and a lighting edge \textit{is} present. This conjunctive behavior is strictly more selective than additive or weighted-sum alternatives, and it comes at zero parameter cost.

\noindent\textbf{Depthwise-Separable Projection.}
\label{sec:dwsep}
The gated features $\mathbf{x}_{\text{att}}$ are projected into the residual space through a depthwise-separable convolution. The depthwise stage applies an independent $3{\times}3$ filter to each channel:
$
\label{eq:dw}
\mathbf{h} = \text{DW}(\mathbf{x}_{\text{att}}), \quad
\text{DW} = \text{Conv2d}(C, C, 3{\times}3, \text{groups}{=}C)
$.
This gives every channel its own local spatial kernel, enabling the module to capture per-feature patterns like lighting gradients and penumbra structures, which cannot be achieved by using a purely pointwise ($1{\times}1$) projection. After group normalization and ReLU activation, a pointwise $1{\times}1$ convolution mixes information across channels:
$
\label{eq:pw}
\Phi(\mathbf{x}) = \text{PW}\!\left(\text{ReLU}(\text{GN}(\mathbf{h}))\right), \quad
\text{PW} = \text{Conv2d}(C, C, 1{\times}1, \text{bias}{=}\text{False})
$.

The pointwise convolution is initialized to all zeros: $\mathbf{W}_{\text{PW}} \leftarrow \mathbf{0}$. This is a deliberate choice with a strong guarantee. At the start of training, regardless of what the gates compute and regardless of the input, the pointwise layer maps everything to zero, so $\Phi(\mathbf{x}) \equiv \mathbf{0}$. The network begins as an \textit{exact} copy of pretrained SAM, not an approximation. As gradients flow back through the loss, the weights grow from zero in directions that reduce the segmentation error, and the \lca{} module ``fades in'' progressively. This avoids the representation damage that randomly-initialized injections can cause in pretrained models.

\subsection{Integration into SAM}
\label{sec:integration}

SAM's image encoder consists of 12 transformer blocks. We keep all of them frozen and attach an \lca{} module to each of the last $N$ blocks \revised{(we use $N{=}4$ for SAM, the last third of its 12-block encoder, see \tableref{tab:nblocks}, and $N{=}10$ for HQ-SAM and SAM~3)}. For a patched block $\mathbf{B}_\ell$, the output becomes:
\begin{equation}
\label{eq:integration}
\mathbf{x}_{\ell+1} = \mathbf{B}_\ell(\mathbf{x}_\ell) + \sigma(\gamma_\ell) \cdot \Phi(\mathbf{x}_\ell)
\end{equation}
where $\gamma_\ell \in \mathbb{R}$ is a learnable gate scalar, initialized to $-1.0$ so that $\sigma(\gamma_\ell) \approx 0.269$ at the start. Note that $\Phi$ takes the block's \textit{input} $\mathbf{x}_\ell$, not its output, so the frozen block runs exactly as before.
Combined with the zero-initialized pointwise convolution (\secref{sec:dwsep}), the initial contribution of \lca{} is exactly:
$
\label{eq:init_zero}
\sigma(-1.0) \cdot \Phi(\mathbf{x}_\ell) = 0.269 \times \mathbf{0} = \mathbf{0}
$.
The network thus starts training as unmodified SAM. Each block then independently learns how much adaptation it needs. Blocks processing more lighting-sensitive features tend to develop larger gate values over the course of training, while blocks that already produce stable features keep their gates small.

Our framework trains three parameter groups. These are (i) the \lca{} modules attached to the patched blocks, (ii) the per-block gate scalars $\gamma_\ell$, and (iii) SAM's mask decoder. 

\subsection{Dual-Stream Training}
\label{sec:training}

The key insight is that a lighting-invariant model should produce identical outputs for the same scene under different illumination. We implement this via a dual-stream setup (\algoref{alg:lca} in the Supplementary). Each training sample is a pair $(I_{\text{clean}}, I_{\text{var}})$ depicting the same scene under clean and variant lighting, sharing ground-truth masks $\mathcal{M} = \{m_1, \ldots, m_K\}$. Both images pass independently through the encoder, generating two sets of spatial embeddings. For each ground-truth mask, we sample a random foreground point and tight bounding box as prompts. Crucially, the same prompts decode masks from both streams, ensuring any prediction difference stems solely from illumination-induced feature variation.
\textit{Supervised loss.}
For each predicted mask $p_k$ and its corresponding ground truth $m_k$, we use a standard combination of binary cross-entropy (BCE) and Dice loss (induced by Dice Similarity Coefficient):
$
\label{eq:seg_loss}
\mathcal{L}_{\text{seg}}(p, m) = \text{BCE}(p, m) + 1 - \text{Dice}(p, m)
$.
Both streams contribute to the supervised signal through a weighted blend:
$
\label{eq:sup_loss}
\mathcal{L}_{\text{sup}} = \lambda_s \cdot \mathcal{L}_{\text{seg}}(p_{\text{clean}}, m) + (1 - \lambda_s) \cdot \mathcal{L}_{\text{seg}}(p_{\text{var}}, m)
$,
where $\lambda_s = 0.5$ by default, weighting both streams equally. This ensures the model learns to segment correctly under \textit{both} lighting conditions, not just the clean one.
\revised{\textit{Prediction consistency loss.}
While the supervised loss tells the model \textit{what} to predict, the prediction consistency loss tells it to predict the \textit{same thing} regardless of lighting. We enforce this with a simple L1 penalty between the two streams' sigmoid-activated mask outputs (not feature maps):}
$
\label{eq:cons_loss}
\mathcal{L}_{\text{cons}} = \left\| \sigma(p_{\text{clean}}) - \sigma(p_{\text{var}}) \right\|_1
$.
This loss requires no additional annotation, only the paired images themselves. \revised{It provides a self-supervised signal that is complementary to, though not independent of, the supervised objective. Both act on the same mask outputs. The consistency term is indifferent to errors the two streams share, and penalises only disagreement between them.}
\noindent
\textit{Total loss.}
The per-instance losses are accumulated across all $K$ instances in a batch:
\begin{equation}
\label{eq:total_loss}
\mathcal{L} = \frac{1}{K} \sum_{k=1}^{K} \left[ \mathcal{L}_{\text{sup}}^{(k)} + \lambda_c \cdot \mathcal{L}_{\text{cons}}^{(k)} \right]
\end{equation}
\revised{We set $\lambda_c = 5$ by default (\tableref{tab:cons_weight_ablation}).} This weight balances segmentation supervision and prediction consistency, encouraging agreement between the clean and variant streams while maintaining mask accuracy. As training progresses and the supervised loss decreases, the relative influence of consistency naturally grows, pushing the model toward illumination invariance.

\section{Experiments}
\label{sec:experiments}

\revised{We evaluate \method{} across five benchmarks spanning general object segmentation, urban driving, controlled synthetic lighting, and real adverse-condition driving scenes. We ask whether \lca{} improves segmentation quality and lighting robustness over existing adaptation strategies, then ablate resolution, gate design, block count and loss weights.}

\subsection{Dataset}
\label{sec:datasets}

\begin{table}[t]
\centering
\caption{\revised{Dataset statistics. Counts are unique source images in our \plap{} splits, and each yields one re-lit variant. The ``Val inst.'' column gives the number of prompted instances the metrics average over. COCO and Cityscapes are class-balanced subsets. ACDC provides real imagery captured under adverse conditions.}}
\label{tab:datasets}
\small
\setlength{\tabcolsep}{3.5pt}
\begin{tabular}{@{}lcccrl@{}}
\toprule
\textbf{Dataset} & \textbf{Cls.} & \textbf{Train} & \textbf{Val} & \textbf{Val inst.} & \textbf{Domain} \\
\midrule
COCO~\cite{lin2014microsoft}       & 80  & \revised{4,209} & \revised{743}   & \revised{5,456} & General \\
VOC 2012~\cite{everingham2015pascal} & 20  & 1,464           & 1,449           & \revised{3,406} & General \\
Cityscapes~\cite{cordts2016cityscapes} & \revised{19} & \revised{2,036} & \revised{361} & \revised{2,255} & Urban \\
Unity (Ours)                       & \revised{164} & \revised{1,398} & 251   & \revised{2,096} & Synthetic \\
\revised{ACDC~\cite{sakaridis2021acdc}} & \revised{8} & \revised{1,573} & \revised{398} & \revised{2,793} & \revised{Real adverse} \\
\bottomrule
\end{tabular}
\end{table}
We evaluate \method{} on \revised{three established benchmarks, one synthetic dataset,
and ACDC~\cite{sakaridis2021acdc}. ACDC supplies real fog, night, rain and snow imagery, which lets us verify that improvements obtained under synthetic re-lighting persist on genuine adverse captures}
(\tableref{tab:datasets}). Full details appear in \secref{additional_results}.


\subsection{Comparative Study}
\label{sec:main_results}












\revised{\tableref{tab:main_results} reports mask quality on the four core datasets under
clean (\textbf{C}) and lighting-variant (\textbf{V}) streams. Rather than listing every
adaptation side by side, we pair each released backbone, fully frozen, with the same
backbone carrying \lca{}, so that each pair measures the overall adaptation gain. The SAM and SAM~2.1 baselines, the decoder-only and \lca{}-only variants, and the four ACDC conditions appear in the full grid in the appendix.}

\revised{\noindent\textbf{Adapter gain.} \lca{} improves upon its own frozen backbone on every dataset, metric and stream. Clean mIoU gains range from $+0.143$ on VOC to $+0.311$ on Cityscapes for HQ-SAM, and from $+0.060$ on COCO to $+0.135$ on Unity for SAM~3. The improvement is thus consistent but diminishes as the encoder strengthens. A stronger pre-trained backbone resolves part of the illumination gap on its own, approximately halving the residual error available to the adapter. We expect this trend to persist as backbones continue to improve.}

\revised{\noindent\textbf{Backbone dependence.} HQ-SAM with \lca{} is strongest on
Cityscapes and Unity, while SAM~3 with \lca{} leads on COCO and VOC, with the largest single
margin on VOC ($0.787$ against $0.756$ mIoU). Batch size differs across model families, since memory requirements differ by an order of magnitude between SAM and SAM~3. Cross-family rows therefore report the best performance each model attains rather than a controlled comparison, and we draw our conclusions from the within-family pairs, which measure the overall gain.}

\begin{table*}[t]
\centering
\caption{\revised{Main results. \textbf{C}\,=\,clean, \textbf{V}\,=\,lighting variant. Each backbone is paired with the same backbone carrying \lca{} and a tuned decoder, so each pair measures the overall gain. \lca{} columns are mean\,$\pm$\,sd over three seeds. \textbf{Bold} marks the best per row and stream. The remaining configurations and the ACDC conditions are in \tableref{tab:t2full_miou}.}$^{\dagger}$\revised{YOLOv11s mIoU (C/V): 0.245/0.238 City, 0.364/0.346 COCO, 0.542/0.518 VOC, 0.052/0.038 Unity.}}
\label{tab:main_results}
\footnotesize
\renewcommand{\arraystretch}{0.9}
\setlength{\tabcolsep}{2pt}
\newcolumntype{Y}{>{\centering\arraybackslash}X}
\begin{tabularx}{\linewidth}{ll|YY|YY|YY|YY}
\toprule
 & & \multicolumn{2}{c|}{HQ-SAM} & \multicolumn{2}{c|}{\textbf{HQ-SAM+\lca{}}} & \multicolumn{2}{c|}{SAM\,3} & \multicolumn{2}{c}{\textbf{SAM\,3+\lca{}}} \\
\cmidrule(lr){3-4}\cmidrule(lr){5-6}\cmidrule(lr){7-8}\cmidrule(lr){9-10}
\textbf{Metric} & \textbf{Dataset} & \textbf{C} & \textbf{V} & \textbf{C} & \textbf{V} & \textbf{C} & \textbf{V} & \textbf{C} & \textbf{V} \\
\midrule
\multirow{4}{*}{mIoU}
  & Cityscapes & 0.4737 & 0.4713 & \textbf{0.7851}{\tiny$\pm$.0006} & \textbf{0.7820}{\tiny$\pm$.0008} & 0.6496 & 0.6486 & 0.7662{\tiny$\pm$.0042} & 0.7634{\tiny$\pm$.0042} \\
  & COCO & 0.6585 & 0.6539 & 0.8178{\tiny$\pm$.0011} & 0.8153{\tiny$\pm$.0012} & 0.7607 & 0.7602 & \textbf{0.8207}{\tiny$\pm$.0006} & \textbf{0.8196}{\tiny$\pm$.0006} \\
  & VOC & 0.6131 & 0.6078 & 0.7561{\tiny$\pm$.0012} & 0.7498{\tiny$\pm$.0012} & 0.6918 & 0.6909 & \textbf{0.7873}{\tiny$\pm$.0015} & \textbf{0.7838}{\tiny$\pm$.0014} \\
  & Unity & 0.6047 & 0.6018 & \textbf{0.8816}{\tiny$\pm$.0017} & \textbf{0.8722}{\tiny$\pm$.0003} & 0.7325 & 0.7284 & 0.8679{\tiny$\pm$.0007} & 0.8591{\tiny$\pm$.0004} \\
\midrule
\multirow{4}{*}{pmAP$_{50}$}
  & Cityscapes & 0.5224 & 0.5166 & \textbf{0.9143}{\tiny$\pm$.0026} & \textbf{0.9112}{\tiny$\pm$.0021} & 0.7623 & 0.7627 & 0.9076{\tiny$\pm$.0053} & 0.9030{\tiny$\pm$.0090} \\
  & COCO & 0.7751 & 0.7692 & \textbf{0.9674}{\tiny$\pm$.0011} & 0.9661{\tiny$\pm$.0003} & 0.9264 & 0.9242 & 0.9663{\tiny$\pm$.0004} & \textbf{0.9666}{\tiny$\pm$.0010} \\
  & VOC & 0.6952 & 0.6841 & 0.8653{\tiny$\pm$.0006} & 0.8578{\tiny$\pm$.0027} & 0.7930 & 0.7927 & \textbf{0.9000}{\tiny$\pm$.0018} & \textbf{0.8939}{\tiny$\pm$.0025} \\
  & Unity & 0.6427 & 0.6398 & \textbf{0.9493}{\tiny$\pm$.0022} & \textbf{0.9440}{\tiny$\pm$.0018} & 0.8192 & 0.8177 & 0.9358{\tiny$\pm$.0004} & 0.9303{\tiny$\pm$.0028} \\
\midrule
\multirow{4}{*}{pmAP$_{75}$}
  & Cityscapes & 0.2483 & 0.2412 & \textbf{0.6782}{\tiny$\pm$.0048} & \textbf{0.6718}{\tiny$\pm$.0044} & 0.4049 & 0.4053 & 0.6262{\tiny$\pm$.0057} & 0.6213{\tiny$\pm$.0070} \\
  & COCO & 0.4512 & 0.4399 & 0.7790{\tiny$\pm$.0031} & 0.7695{\tiny$\pm$.0026} & 0.6342 & 0.6389 & \textbf{0.7895}{\tiny$\pm$.0026} & \textbf{0.7873}{\tiny$\pm$.0020} \\
  & VOC & 0.4339 & 0.4313 & 0.6410{\tiny$\pm$.0010} & 0.6328{\tiny$\pm$.0024} & 0.5948 & 0.5916 & \textbf{0.6945}{\tiny$\pm$.0053} & \textbf{0.6895}{\tiny$\pm$.0045} \\
  & Unity & 0.4399 & 0.4447 & \textbf{0.8569}{\tiny$\pm$.0040} & \textbf{0.8408}{\tiny$\pm$.0013} & 0.6202 & 0.6088 & 0.8360{\tiny$\pm$.0023} & 0.8259{\tiny$\pm$.0017} \\
\midrule
\multirow{4}{*}{pmAP}
  & Cityscapes & 0.2757 & 0.2711 & \textbf{0.6417}{\tiny$\pm$.0004} & \textbf{0.6356}{\tiny$\pm$.0008} & 0.4244 & 0.4222 & 0.6097{\tiny$\pm$.0077} & 0.6052{\tiny$\pm$.0080} \\
  & COCO & 0.4473 & 0.4404 & 0.6934{\tiny$\pm$.0018} & 0.6882{\tiny$\pm$.0019} & 0.5895 & 0.5894 & \textbf{0.7007}{\tiny$\pm$.0008} & \textbf{0.6984}{\tiny$\pm$.0008} \\
  & VOC & 0.4318 & 0.4244 & 0.6123{\tiny$\pm$.0013} & 0.6019{\tiny$\pm$.0021} & 0.5584 & 0.5563 & \textbf{0.6606}{\tiny$\pm$.0023} & \textbf{0.6549}{\tiny$\pm$.0025} \\
  & Unity & 0.4356 & 0.4328 & \textbf{0.8264}{\tiny$\pm$.0034} & \textbf{0.8089}{\tiny$\pm$.0006} & 0.5935 & 0.5862 & 0.8043{\tiny$\pm$.0012} & 0.7903{\tiny$\pm$.0012} \\
\bottomrule
\end{tabularx}
\end{table*}

\revised{\noindent\textbf{Clean-to-variant gaps.} Each validation image contributes one re-lit variant, with severity sampled uniformly across the three tiers. Averaged over a dataset, this produces small clean-to-variant gaps for every model, including the frozen backbones, which receive no lighting-specific training. These aggregate gaps therefore characterise the perturbation rather than the illumination robustness of any particular model. The robustness claim rests instead on the severity sweep and on real adverse imagery, both examined in \secref{sec:robustness}.}

\revised{\noindent\textbf{Per-instance distribution.} On the Cityscapes lighting-variant stream \lca{} raises mean instance IoU from $0.471$ to $0.782$ and improves $2{,}152$ of $2{,}252$ instances ($95.6\%$). The frozen backbone spreads instances across the entire IoU range, whereas \lca{} concentrates them near $1.0$. The improvement is therefore distributed across the dataset rather than confined to a small number of cases (\figref{fig:iou_analysis}).}

\revised{Individual cases follow the same pattern, with the baseline flooding low-contrast regions while \lca{} recovers a compact mask (\secref{sec:supp_qualitative}). GradCAM maps in the appendix show the contrast gate refocusing encoder activation onto high-contrast object boundaries.}




\revised{\subsection{Computational Overhead}
\label{sec:compute}
\tableref{tab:compute_main} reports deployment cost on a single RTX~4090 at
$1024{\times}1024$. We patch $N{=}4$ blocks for SAM and $N{=}10$ for HQ-SAM and SAM~3.
Our full model, \lca{}+Dec, adds 2.50\,M parameters to SAM (2.8\%), 24.4\,GFLOPs (5.0\%), 2.6\,ms of latency and 50.4\,MB of extra activation memory. On HQ-SAM, the deeper $N{=}10$ configuration adds 5.99\,M parameters (6.3\%), 24.4\,GFLOPs (4.9\%), 2.3\,ms and 52.2\,MB. On SAM~3 the same $N{=}10$ module adds 10.61\,M parameters, but against a far larger model this is only 2.3\%, with 54.9\,GFLOPs (2.0\%), 6.2\,ms and 69.8\,MB.

\noindent At the same depth ($N{=}10$), the absolute cost grows with the backbone while the relative cost falls, from 6.3\% of a 94.8\,M model to 2.3\% of a 458\,M one. \lca{} therefore scales with the encoder rather than imposing a constant cost, which is the property that matters when the adapter is applied to a larger backbone. The image encoder remains frozen in all configurations, reducing the number of trainable parameters. Dec updates the 4.06\,M mask decoder, \lca{} updates the adapter alone, and \lca{}+Dec updates both.}

\begin{table}[t]
\centering
\caption{\revised{Deployment cost on one RTX~4090, fp16, batch~1 at $1024{\times}1024$. Median of 100 iterations after 20 warm-up, with \textbf{IQR} over those iterations. \textbf{Params} denotes parameter count (M).}}
\label{tab:compute_main}
\small
\setlength{\tabcolsep}{3pt}
\begin{tabular}{@{}l|ccccc@{}}
\toprule
\revised{Model} & \revised{Params} & \revised{GFLOPs} & \revised{Lat.} & \revised{IQR} & \revised{Mem.} \\
 & \revised{(M)} & & \revised{(ms)} & \revised{(ms)} & \revised{(MB)} \\
\midrule
\revised{SAM}                & \revised{89.67}  & \revised{487.9}  & \revised{41.61} & \revised{0.03} & \revised{1585.7} \\
\revised{~~+ \lca{}}         & \revised{92.17}  & \revised{512.3}  & \revised{44.87} & \revised{0.03} & \revised{1628.4} \\
\revised{~~+ \lca{}+Dec}     & \revised{92.17}  & \revised{512.3}  & \revised{44.25} & \revised{0.07} & \revised{1636.1} \\
\midrule
\revised{HQ-SAM}             & \revised{94.81}  & \revised{494.5}  & \revised{43.38} & \revised{0.29} & \revised{1625.8} \\
\revised{~~+ \lca{}+Dec}     & \revised{100.80}  & \revised{518.9}  & \revised{45.67} & \revised{0.04} & \revised{1678.0} \\
\midrule
\revised{SAM~3}              & \revised{458.26} & \revised{2810.1} & \revised{65.94} & \revised{0.30} & \revised{2069.0} \\
\revised{~~+ \lca{}+Dec}     & \revised{468.87} & \revised{2865.0} & \revised{72.13} & \revised{0.21} & \revised{2138.8} \\
\bottomrule
\end{tabular}
\end{table}

\subsection{Ablation Studies}
\label{sec:ablation}

\revised{\noindent\textbf{Input resolution.}} \revised{Resolution enters through two independent settings. 
The data axis varies the loader while the encoder still receives $4{,}096$ tokens, so no embedding is interpolated. 
The encoder axis varies the encoder input itself, shrinking the token grid and interpolating the position embedding. 
With validation held at a fixed resolution, higher input resolutions generally improve mask quality:
 on the encoder axis \lca{} goes from $0.567$ to $0.746$ mIoU and the frozen backbone from $0.385$ to $0.613$.
  The data axis rises to 512 and then flattens, 
  so image detail beyond 512 adds little once the encoder receives a full token grid.
   \lca{} holds a margin of $0.13$ to $0.18$ mIoU at every resolution on both axes. The complete results are shown in \secref{sec:supp_resolution}.}

\revised{\noindent\textbf{Gate ablation.}} \revised{\tableref{tab:gate_ablation} removes each gate in turn from the three-gate module.
The comparison of gate-composition strategies is in \secref{sec:supp_design}.}

\begin{table}[t]
\centering
\caption{\revised{Gate ablation. One gate removed at a time, VOC, $N{=}4$, mean over three seeds, with paired $t$ against the protocol-matched baseline that keeps all three gates.}}
\label{tab:gate_ablation}
\footnotesize
\setlength{\tabcolsep}{3.5pt}
\begin{tabular}{@{}l|cc|cc@{}}
\toprule
 & \multicolumn{2}{c|}{\revised{Clean}} & \multicolumn{2}{c}{\revised{Variant}} \\
\cmidrule(lr){2-3}\cmidrule(lr){4-5}
\revised{Configuration} & \revised{mIoU} & \revised{$\Delta$ ($t$)} & \revised{mIoU} & \revised{$\Delta$ ($t$)} \\
\midrule
\revised{Three gates}     & \revised{\textbf{0.7129}} & \revised{n/a} & \revised{\textbf{0.7056}} & \revised{n/a} \\
\revised{~~$-$ channel}   & \revised{0.7127} & \revised{$-$0.0001 ($-$0.2)} & \revised{0.7050} & \revised{$-$0.0007 ($-$0.9)} \\
\revised{~~$-$ spatial}   & \revised{0.7099} & \revised{$-$0.0030 ($-$1.4)} & \revised{0.7024} & \revised{$-$0.0032 ($-$1.7)} \\
\revised{~~$-$ Laplacian} & \revised{0.7058} & \revised{$\bm{-}$\textbf{0.0071} ($\bm{-}$\textbf{3.1})} & \revised{0.6989} & \revised{$\bm{-}$\textbf{0.0067} ($\bm{-}$\textbf{3.0})} \\
\bottomrule
\end{tabular}
\end{table}

\revised{All three removals reduce accuracy on both streams. The fixed-Laplacian contrast gate has the largest effect, costing $0.0071$ clean and $0.0067$ variant mIoU. The spatial gate accounts for $0.0030$ clean and $0.0032$ variant, and the channel gate for $0.0001$ clean and $0.0007$ variant. The three gates therefore operate at different scales, with the grounded contrast prior carrying the largest share, which is consistent with the illumination-specific role it is designed to play.}

\revised{\noindent\textbf{Loss weights.}} \revised{The consistency weight has a clear optimum on the stream it targets. Moving from $\lambda_{\text{cons}}{=}0$ to $5$ gains $0.0051$ mIoU on the lighting-variant stream against $0.0022$ on clean, and very large values degrade the model as stream agreement overrides segmentation accuracy. The supervised blend varies by less than $0.004$, since the consistency term already couples the streams. We adopt $\lambda_{\text{cons}}{=}5$ and $\lambda_{\text{sup}}{=}0.5$, swept in \secref{sec:supp_design}.}

\revised{\noindent\textbf{Number of patched blocks.}} \revised{\tableref{tab:nblocks}
sweeps the number of patched blocks $N$. Almost the entire gain arrives by $N{=}4$. The adapter
lifts VOC from $0.6135$ to $0.7029$ clean mIoU, and extending to $N{=}10$ adds only $0.0043$
while multiplying the adapter by $2.5\times$. Returns are flat from $N{=}6$ onward, with
$N{=}6$ to $N{=}10$ spanning $0.0018$ mIoU. These results support $N{=}4$ as a practical balance between mask quality and adapter size.}

\begin{table}[t]
\centering
\caption{\revised{Number of patched blocks $N$, VOC, \lca{}-only, mean\,$\pm$\,sd over three seeds. $N{=}0$ is the frozen backbone.}}
\label{tab:nblocks}
\small
\setlength{\tabcolsep}{5pt}
\begin{tabular}{@{}c|c|cc@{}}
\toprule
\revised{$N$} & \revised{\lca{} params} & \revised{Clean} & \revised{Variant} \\
\midrule
\revised{0}   & \revised{n/a}      & \revised{0.6135} & \revised{0.6091} \\
\revised{2}   & \revised{1.25\,M} & \revised{0.6950\,{\tiny$\pm$.0058}} & \revised{0.6886\,{\tiny$\pm$.0045}} \\
\revised{\textbf{4}} & \revised{\textbf{2.49\,M}} & \revised{0.7029\,{\tiny$\pm$.0053}} & \revised{0.6947\,{\tiny$\pm$.0056}} \\
\revised{6}   & \revised{3.74\,M} & \revised{\textbf{0.7090}\,{\tiny$\pm$.0025}} & \revised{\textbf{0.7017}\,{\tiny$\pm$.0026}} \\
\revised{8}   & \revised{4.99\,M} & \revised{0.7079\,{\tiny$\pm$.0011}} & \revised{0.7011\,{\tiny$\pm$.0021}} \\
\revised{10}  & \revised{6.24\,M} & \revised{0.7072\,{\tiny$\pm$.0030}} & \revised{0.7000\,{\tiny$\pm$.0037}} \\
\bottomrule
\end{tabular}
\vspace{-6pt}
\end{table}

\begin{table}[t]
\centering
\caption{\revised{\textbf{(a)} Relative mIoU degradation at severity~3,
$100\cdot(\text{clean}-\text{sev3})/\text{clean}$, lower is better; absolute scores in
\tableref{tab:severity_full}. \textbf{(b)} Synthetic-to-real transfer: \lca{}+Dec on SAM~3
trained on COCO and evaluated on real ACDC with no ACDC gradient steps, mIoU on the unmodified photograph, averaged over three seeds.}}
\label{tab:robustness}
\small
\setlength{\tabcolsep}{3pt}
\begin{minipage}[t]{0.53\linewidth}
\centering
\revised{\textbf{(a)} Degradation at sev3}\\[3pt]
\begin{tabular}{@{}l|cccc@{}}
\toprule
\revised{Data} & \revised{SAM} & \revised{Dec} & \revised{\lca{}} & \revised{+Dec} \\
\midrule
\revised{City}  & \revised{1.87} & \revised{0.95} & \revised{0.76} & \revised{\textbf{0.74}} \\
\revised{COCO}  & \revised{1.76} & \revised{0.90} & \revised{0.96} & \revised{\textbf{0.83}} \\
\revised{VOC}   & \revised{1.80} & \revised{2.33} & \revised{1.82} & \revised{\textbf{1.69}} \\
\revised{Unity} & \revised{1.02} & \revised{1.10} & \revised{1.07} & \revised{\textbf{0.91}} \\
\bottomrule
\end{tabular}
\end{minipage}
\hfill
\begin{minipage}[t]{0.45\linewidth}
\centering
\revised{\textbf{(b)} COCO to real ACDC}\\[3pt]
\setlength{\tabcolsep}{2pt}
\begin{tabular}{@{}l|cc|c@{}}
\toprule
\revised{Cond.} & \revised{SAM\,3} & \revised{+\lca{}} & \revised{$\Delta$} \\
\midrule
\revised{Fog}   & \revised{0.7393} & \revised{\textbf{0.8212}} & \revised{\textbf{+.082}} \\
\revised{Night} & \revised{0.6623} & \revised{\textbf{0.7431}} & \revised{\textbf{+.081}} \\
\revised{Rain}  & \revised{0.6961} & \revised{\textbf{0.7777}} & \revised{\textbf{+.082}} \\
\revised{Snow}  & \revised{0.7027} & \revised{\textbf{0.8021}} & \revised{\textbf{+.099}} \\
\bottomrule
\end{tabular}
\end{minipage}
\vspace{-6pt}
\end{table}

\revised{\subsection{Robustness and Transfer}}
\label{sec:robustness}

\revised{\noindent\textbf{Severity degradation.}} \revised{\tableref{tab:robustness}(a) reports relative mIoU degradation at the most severe setting. \lca{}+Dec degrades least on all four core datasets, and the ordering is not inherited from clean accuracy. On VOC, decoder tuning degrades more than the frozen backbone ($2.33$ against $1.80$) while \lca{}+Dec degrades least at $1.69$. Decoder tuning alone therefore improves accuracy without improving robustness, which is precisely the distinction this work targets.}

\revised{\noindent\textbf{Synthetic-to-real transfer.}} \revised{\plap{} variants and Unity are both synthetic, so we assess whether the improvements persist on real imagery. \tableref{tab:robustness}(b) trains on COCO with \plap{} and evaluates on real ACDC with no ACDC gradient steps and no scene overlap. \lca{}+Dec improves over frozen SAM~3 by $0.081$ to $0.099$ mIoU on all four conditions. The night condition is hardest for the frozen model at $0.6623$ mIoU and gains $0.081$ after adaptation. Fog, rain, and snow additionally involve scattering and occlusion, and we report them as evidence of breadth rather than of illumination robustness specifically. Training under synthetic re-lighting therefore transfers to genuine adverse capture.}

\noindent\textbf{Limitations.} \revised{The fixed Laplacian prior is sensitive to the signal-to-noise ratio. Under genuine low-light capture, sensor noise can obscure structural edges, and high-pass filtering may amplify this noise, limiting the effectiveness of the contrast gate (\secref{sec:supp_positioning}). \plap{} re-lighting is milder than real adverse capture, so
the clean-to-variant gaps understate real degradation, and the transfer experiment is the
direct measure of it. Batch size differs across model families, so cross-family rows report
the best each model reaches rather than a controlled comparison.}

\section{Conclusion}
\label{sec:conclusion}
We introduced PLAP-LCA to improve SAM's robustness to illumination changes through three contributions. The first is the LCA module, \revised{a lightweight three-gate attention module whose channel, spatial, and Laplacian contrast gates are fused conjunctively to target the three principal dimensions of illumination corruption}. \revised{The second is a stratified severity sampling pairwise training strategy with a prediction consistency loss.} The third is a comprehensive empirical study across benchmarks and a new Unity-based synthetic dataset. \revised{Experiments demonstrate consistent improvements under diverse lighting on both SAM and SAM~3 encoders, and transfer to real adverse-weather imagery with no target-domain training.} \revised{A three-seed ablation shows that removing any gate reduces mean performance, with the largest drop from removing the fixed Laplacian contrast gate. These results support the three-gate design and highlight the value of its lighting-specific prior.} \revised{Future work should address the low-SNR regime, where a fixed Laplacian prior amplifies noise rather than structure.}

\section*{Acknowledgments}
This work was supported by the AI Institute for Resilient
Agriculture (USDA-NIFA Grant No.~2021-67021-35329) and
COALESCE: COntext Aware LEarning for Sustainable
CybEr-Agricultural Systems (NSF CPS Frontier Grant
No.~1954556).

{
    \small
    \bibliographystyle{ieeenat_fullname}
    \bibliography{main}
}
\clearpage
\appendix
\section{Supplementary Materials}



\subsection{Additional Related Work}
\label{additional_related_work}

\noindent\textbf{Attention Mechanisms for Visual Robustness.}
Attention mechanisms have been widely adopted to enhance visual recognition,
evolving from channel (SE-Net~\cite{hu2018squeeze}) and spatial attention
(CBAM~\cite{woo2018cbam}) to the self-attention paradigms driving Vision
Transformers~\cite{dosovitskiy2020image}. Despite these advancements,
attention mechanisms explicitly designed for environmental robustness remain
underexplored. The closest parallel to our objective involves feature
modulation and normalization strategies~\cite{pan2018two}, which adapt batch
statistics to align different domains. However, these methods remain
implicitly learned and agnostic to the physical properties of the
degradation. Our \lca{} module distinguishes itself by using contrast maps
as an explicit, physically-grounded illumination signal. Combined with a
learned gating mechanism, this design isolates structural cues from lighting
artifacts while retaining the foundational knowledge embedded within the
pre-trained backbone.

\noindent\textbf{Data Augmentation for Robustness.}
Standard augmentation pipelines~\cite{cubuk2019autoaugment,cubuk2020randaugment}
apply photometric and geometric transformations without explicit physical
motivation. Corruption benchmarks~\cite{hendrycks2019benchmarking} evaluate
robustness but do not provide training-time augmentation strategies, while
AugMax~\cite{wang2021augmax} and adversarial augmentation
approaches~\cite{wang2025comprehensive} improve worst-case robustness but
lack structured severity control. More recently, the community has pivoted
toward illumination-centric data synthesis. Recent frameworks employ inverse
rendering for physically-based lighting augmentation~\cite{jin2025physically},
utilize diffusion models for diverse illumination
synthesis~\cite{chaturvedi2025synthlight}, or apply color constancy
principles for varied illuminant
simulation~\cite{chankhachon2025deep}. While these approaches
inject realistic lighting variation, they often incur high generation cost
and do not organize synthesized data into progressive difficulty tiers. Our
\method{} provides physically motivated operations with conflict-aware
sampling and structured severity levels, complemented by a Unity-based
synthetic dataset for physically accurate evaluation.

\subsection{Full Architecture Details}
\label{sec:supp_arch}
The main paper provides a high-level overview of PLAP-LCA in \figref{fig:training_pipeline}. Here we present the detailed diagrams of the training pipeline \figref{fig:supp_training_pipeline} and the LCA module internals \figref{fig:supp_lca_arch}, followed by the complete forward-pass pseudocode \algoref{alg:lca}.

\subsubsection{Training Pipeline}
\label{sec:supp_training_pipeline}

\begin{figure*}[tp]
    \centering
    \includegraphics[width=\textwidth]{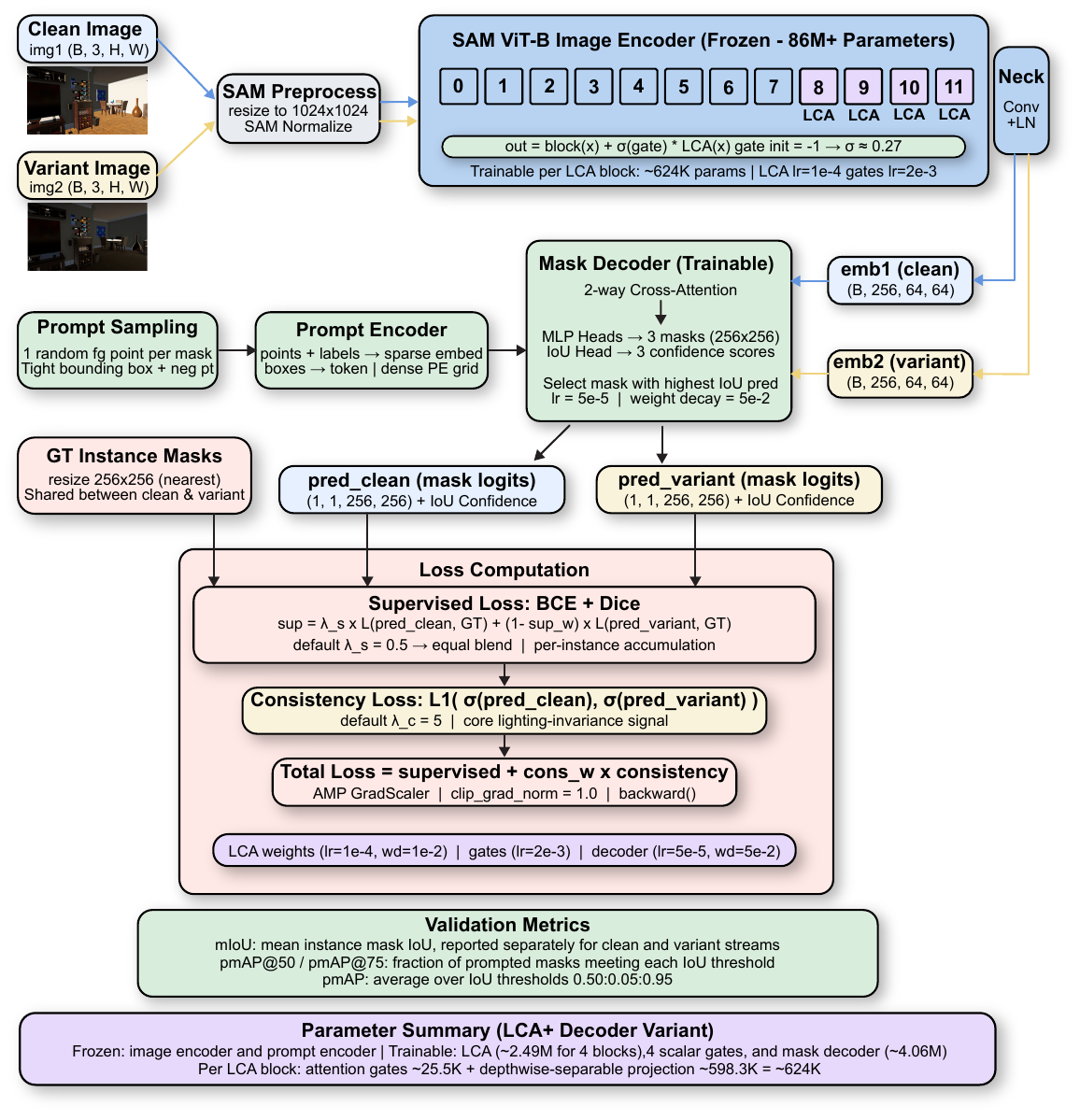}
    \caption{%
        Detailed PLAP-LCA training pipeline.
        Clean and variant images are processed by a shared, frozen SAM encoder with LCA modules at the last \revised{four} blocks. The prompts remain unchanged and are shared with both streams. The supervised loss trains against shared ground-truth masks, while the consistency loss penalizes any prediction divergence caused purely by illumination.%
    }
    \label{fig:supp_training_pipeline}
\end{figure*}

\figref{fig:supp_training_pipeline} shows the full dual stream training setup. Both the clean and variant images share the same frozen SAM encoder with LCA modules included into the last \revised{four} blocks. The prompt encoder and mask decoder are shared across both streams. Only the LCA
weights, the per-block gate scalars, and the mask decoder are updated during training.

\subsubsection{LCA Module Architecture}
\label{sec:supp_lca_arch}

\begin{figure*}[tp]
    \centering
    \includegraphics[width=0.9\textwidth]{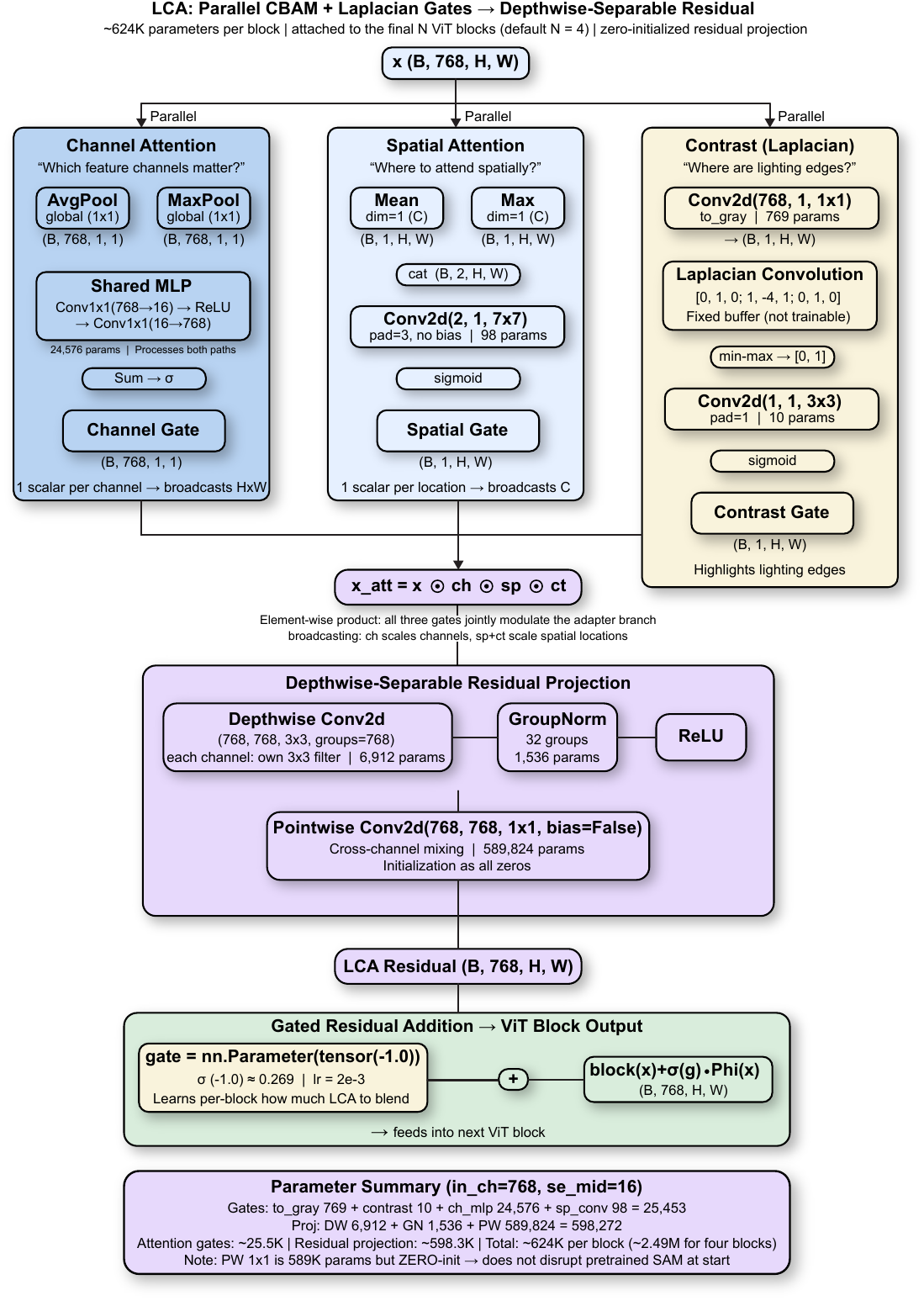}
    \caption{%
        LCA module architecture.The three gates run in parallel. The \textbf{channel gate} (left) uses global average and max pool statistics
        fed through a shared MLP to produce a per-channel scalar mask. The \textbf{spatial gate} (center) concatenates channel-wise mean and max maps
        and convolves them with a $7{\times}7$ kernel to produce a per-location scalar mask. The \textbf{contrast gate} (right) projects the feature map to a single-channel grayscale, applies a fixed Laplacian kernel, min-max normalizes the result, and refines it with a trainable convolution to highlight lighting edges. All three gates are multiplied onto the input feature tensor, the gated features then pass through a projection.
        }
    \label{fig:supp_lca_arch}
\end{figure*}

\figref{fig:supp_lca_arch} displays the internal structure of the LCA module. Three attention gates (channel, spatial, contrast) run in parallel and are fused by element-wise multiplication before a depthwise-separable residual projection.

\begin{algorithm}[t]
\caption{Forward Pass of the \lca{} Module}
\label{alg:lca}
\begin{algorithmic}[1]
\Require Block input $\mathbf{x} \in \mathbb{R}^{B \times C \times H \times W}$,
         gate scalar $\gamma$
\Ensure  Adapted output $\mathbf{x}_{\text{out}} \in \mathbb{R}^{B \times C \times H \times W}$

\Statex \textcolor{gray}{\textit{\# Channel Attention Gate}}
\State $\mathbf{d}_{\text{avg}} \gets \text{AdaptiveAvgPool2d}(\mathbf{x})$ \Comment{$(B, C, 1, 1)$}
\State $\mathbf{d}_{\text{max}} \gets \text{AdaptiveMaxPool2d}(\mathbf{x})$ \Comment{$(B, C, 1, 1)$}
\State $\mathbf{G}_{\text{ch}} \gets \sigma\!\big(\text{MLP}(\mathbf{d}_{\text{avg}}) + \text{MLP}(\mathbf{d}_{\text{max}})\big)$ \Comment{shared MLP, $(B, C, 1, 1)$}

\Statex
\Statex \textcolor{gray}{\textit{\# Spatial Attention Gate}}
\State $\mathbf{s}_{\text{avg}} \gets \text{Mean}(\mathbf{x}, \text{dim}{=}1)$         \Comment{$(B, 1, H, W)$}
\State $\mathbf{s}_{\text{max}} \gets \text{Max}(\mathbf{x}, \text{dim}{=}1)$          \Comment{$(B, 1, H, W)$}
\State $\mathbf{G}_{\text{sp}} \gets \sigma\!\big(\text{Conv}_{7 \times 7}([\mathbf{s}_{\text{avg}} \,;\, \mathbf{s}_{\text{max}}])\big)$ \Comment{$(B, 1, H, W)$}

\Statex
\Statex \textcolor{gray}{\textit{\# Contrast Attention Gate}}
\State $\mathbf{g} \gets \text{Conv}_{1 \times 1}^{\text{proj}}(\mathbf{x})$           \Comment{learned grayscale, $(B, 1, H, W)$}
\State $\mathbf{e} \gets \mathbf{g} * \mathbf{K}_{\text{lap}}$                         \Comment{fixed Laplacian, no grad}
\State $\hat{\mathbf{e}} \gets (\mathbf{e} - \min\mathbf{e}) \,/\, (\max\mathbf{e} - \min\mathbf{e} + \epsilon)$ \Comment{per-sample normalize}
\State $\mathbf{G}_{\text{ct}} \gets \sigma\!\big(\text{Conv}_{3 \times 3}^{\text{refine}}(\hat{\mathbf{e}})\big)$ \Comment{$(B, 1, H, W)$}

\Statex
\Statex \textcolor{gray}{\textit{\# Multiplicative Fusion}}
\State $\mathbf{x}_{\text{att}} \gets \mathbf{x} \odot \mathbf{G}_{\text{ch}} \odot \mathbf{G}_{\text{sp}} \odot \mathbf{G}_{\text{ct}}$ \Comment{broadcast multiply}

\Statex
\Statex \textcolor{gray}{\textit{\# Depthwise-Separable Projection (PW is zero-init)}}
\State $\mathbf{h} \gets \text{DWConv}_{3 \times 3}(\mathbf{x}_{\text{att}})$          \Comment{groups $= C$}
\State $\mathbf{h} \gets \text{ReLU}\!\big(\text{GroupNorm}(\mathbf{h})\big)$
\State $\Phi(\mathbf{x}) \gets \text{PWConv}_{1 \times 1}(\mathbf{h})$                 \Comment{$\mathbf{W}_{\text{PW}}$ init $= \mathbf{0}$}

\Statex
\Statex \textcolor{gray}{\textit{\# Gated Residual Addition}}
\State $\mathbf{x}_{\text{out}} \gets \text{Block}(\mathbf{x}) + \sigma(\gamma) \cdot \Phi(\mathbf{x})$ \Comment{frozen SAM block + \lca{}}

\Statex
\State \Return $\mathbf{x}_{\text{out}}$
\end{algorithmic}
\end{algorithm}

\subsection{PLAP: Operations and Severity Configuration}
\label{sec:supp_plap}

\subsubsection{Image-Space Operations}
\label{image_space_operations}

The following operations form the PLAP library. All transforms are physically motivated and operate in floating-point precision before conversion back to \texttt{uint8}.

\begin{itemize}

\item \textbf{Exposure (EV adjustment):}
Images are converted from sRGB to linear light space, scaled by $2^{\text{EV}}$, then converted back. This mirrors how real cameras adjust exposure.

\item \textbf{Directional Shadow:}
A smooth gradient mask, parameterized by angle $\theta$, strength $s$,
and sharpness $\xi$, darkens one side of the image to approximate the
effect of a strong directional light source being partially occluded.
The sharpness parameter scales with severity level, from a soft gradient at
mild to a near-knife-edge boundary at severe.

\item \textbf{Color Temperature (Warm/Cool):}
Channel-wise multiplication shifts the red-blue balance.
Warm boosts red and suppresses blue. Cool does the reverse.
These are physically opposite effects and are therefore treated as mutually
exclusive in the conflict-aware sampler.

\item \textbf{Vignetting:}
A radial power-law falloff mask progressively darkens the image toward its
edges. The center position is sampled slightly off-center to increase realism.

\item \textbf{Contrast:}
Each channel is scaled linearly around its own mean:
$I_{\text{out}} = (I - \mu_c) \cdot f + \mu_c$.
Operating in per-channel mode avoids introducing unwanted color shifts.

\item \textbf{Gamma Correction:}
The pointwise nonlinearity $I_{\text{out}} = I_{\text{in}}^{1/\gamma}$
reshapes the tonal curve, lifting or compressing midtones while leaving
pure black and white unchanged.

\item \textbf{Brightness:}
A uniform scale factor $(1 + p/100)$ approximates a global change in
illumination level without altering contrast or color balance.
Treated as mutually exclusive with contrast adjustment.

\item \textbf{Film Grain:}
Additive Gaussian noise drawn independently per pixel and per channel
introduces the speckle pattern typical of high-ISO capture or
underlit sensors.

\item \textbf{Atmospheric Haze:}
The image is blended toward a fixed pale haze color via
$I_{\text{out}} = I \cdot (1 - a) + C_{\text{haze}} \cdot a$,
producing the washed-out appearance of fog, mist, or smog.
Treated as mutually exclusive with contrast adjustment and with lens flare.



\end{itemize}

\subsubsection{Conflict-Aware Sampling}
\label{sec:supp_conflicts}

The \plap{} sampler uses three mutual-exclusion groups to avoid contradictory or redundant operations:

\begin{enumerate}
    \item \texttt{warm} $\leftrightarrow$ \texttt{cool}, opposite color temperatures.
    \item \texttt{exposure} $\leftrightarrow$ \texttt{brightness}, redundant luminance adjustments.
    \item \texttt{haze} $\leftrightarrow$ \texttt{contrast}, haze inherently flattens contrast.
\end{enumerate}

\subsubsection{Severity Level Configuration}
\label{sec:supp_severity_config}

\tableref{tab:severity} lists the parameter ranges and maximum operation counts for each severity level. Higher severity levels use wider or stronger parameter ranges and allow more operations to be composed, with some parameter ranges overlapping across levels. Shadow sharpness $\xi$ also scales with severity, making the compound difficulty grow faster than the parameter values.

\subsection{Severity Levels}\label{severity_levels}
\begin{table}[htbp]
\centering
\caption{Severity level configuration. Each level controls the parameter range for each operation and the maximum number of simultaneously applied operations.}
\label{tab:severity}
\small
\setlength{\tabcolsep}{4pt}
\begin{tabular}{@{}lccc@{}}
\toprule
 & \textbf{Severity 1} & \textbf{Severity 2} & \textbf{Severity 3} \\
 & (Mild) & (Moderate) & (Severe) \\
\midrule
Exposure (EV) & $[-0.3, +0.3]$ & $[-0.8, +0.8]$ & $[-1.5, +1.5]$ \\
Brightness (\%) & $[-15, +15]$ & $[-30, +30]$ & $[-45, +45]$ \\
Contrast factor & $[0.9, 1.1]$ & $[0.75, 1.25]$ & $[0.6, 1.4]$ \\
Gamma & $[0.85, 1.15]$ & $[0.7, 1.3]$ & $[0.55, 1.5]$ \\
Color temp. tint & $[0.03, 0.07]$ & $[0.08, 0.14]$ & $[0.15, 0.25]$ \\
Vignette strength & $[0.1, 0.2]$ & $[0.2, 0.4]$ & $[0.4, 0.65]$ \\
Shadow strength & $[0.2, 0.35]$ & $[0.35, 0.55]$ & $[0.55, 0.75]$ \\
Grain intensity & $[0.01, 0.02]$ & $[0.02, 0.04]$ & $[0.04, 0.07]$ \\
Haze amount & $[0.05, 0.10]$ & $[0.10, 0.20]$ & $[0.20, 0.35]$ \\
\midrule
Max operations & 1 & 2 & 3 \\
\bottomrule
\end{tabular}
\end{table}

\subsubsection{Image Quality Validation}
\label{sec:supp_imgquality}

\tableref{tab:image_quality} reports perceptual image quality metrics between clean and the lighting augmentation across all four datasets. Across Cityscapes, COCO, and VOC, augmented images maintain low FID scores and high SSIM/FSIM values, which confirms that PLAP introduces photometric variation while preserving structural and perceptual similarity to the original images. The Unity synthetic dataset also shows higher FID and lower SSIM/FSIM values, reflecting the broad distributional shift introduced by unity stimulation engine rather than image-space operations. Despite this, SSIM and FSIM remain relatively high, that shows that structural scene content is preserved.

\begin{table}[htbp]
\centering
\caption{%
    Perceptual image quality metrics between clean and lighting-augmented
    pairs across four benchmarks. Unity yields higher FID due to physically
    rendered scene variation rather than image-space augmentation.
    (FID $\downarrow$: Fr\'{e}chet Inception Distance;
     SSIM $\uparrow$: Structural Similarity Index;
     FSIM $\uparrow$: Feature Similarity Index)%
}
\label{tab:image_quality}
\begin{tabular}{llccc}
\toprule
\textbf{Dataset} & \textbf{Split}
    & \textbf{FID} $\downarrow$
    & \textbf{SSIM} $\uparrow$
    & \textbf{FSIM} $\uparrow$ \\
\midrule
\multirow{2}{*}{Cityscapes} & train &  4.1753 & 0.8486 & 0.9022 \\
                            & val   & 11.7663 & 0.8449 & 0.8993 \\
\midrule
\multirow{2}{*}{COCO}       & train &  3.1833 & 0.8632 & 0.9211 \\
                            & val   & 10.2894 & 0.8636 & 0.9224 \\
\midrule
\multirow{2}{*}{VOC}        & train &  6.3118 & 0.8649 & 0.9208 \\
                            & val   &  6.5617 & 0.8674 & 0.9237 \\
\midrule
\multirow{2}{*}{Unity}      & train & 22.6932 & 0.7447 & 0.8782 \\
                            & val   & 58.5673 & 0.7299 & 0.8713 \\
\bottomrule
\end{tabular}
\end{table}

\subsection{Unity Dataset Details}
\label{unity_dataset}

\subsubsection{Scene Construction}
We construct our dataset using pre-existing indoor apartment environments within Unity. Each simulation contains a realistic array of household items including furniture, appliances, and decorations. We combine multiple
distinct apartment layouts and capture a wide variety of spatial configurations in our training data. The pipeline is illustrated in~\figref{fig:unity-pipeline}. The simulation environment enables automatic generation of precise instance masks for every object. The property summary of the dataset is shown in~\tableref{tab:unity_dataset}. 

\subsubsection{Lighting Rig}
For each scene, we design lighting configurations to produce lighting variants that are aligned with our previous three-level severity framework. To generate these complex environments, we manipulate the physical properties of light sources, such as intensity, color temperature, and angular direction. This allows us to model a wide range of realistic illumination changes. Each configuration is also saved alongside the variant image, so the capture can be reproduced.   
At the \textbf{mild} level, we adjust the main directional light by $\pm$20 to 30\% and modify its color temperature slightly warm or cool. The rendered images look subtly different. At the \textbf{moderate} level, we start adding or removing additional lights, simulating overcast conditions.  At the \textbf{severe} level, we drop the scene into darkness with a single harsh point light, or increase the lighting intensity to an extreme level. The boundaries between objects become difficult to distinguish due to severe contrast degradation.

\noindent\textbf{Camera Strategy}
We employ a camera rotation strategy for viewpoint diversity. The camera is placed at fixed anchor positions within each scene and performs systematic rotations. This rotational strategy aims to maximize coverage of each scene, resulting in a variety of synthetic indoor capture scenarios. For training, each viewpoint provides a clean image paired with one randomly selected lighting variant, sharing identical ground-truth masks.

\begin{table}[htbp]
\centering
\caption{Unity synthetic dataset summary.}
\label{tab:unity_dataset}
\small
\begin{tabular}{@{}lc@{}}
\toprule
\textbf{Property} & \textbf{Value} \\
\midrule
Apartment scenes          & 10 \\
Camera positions per iteration & 16 \\
Available lighting settings & 4 (reference + 3 severity) \\
Clean-variant pairs (train + val) & 1{,}649 \\
Instance annotations      & 13{,}361 \\
Resolution                & 1920$\times$1080 \\
Annotation format         & COCO JSON (masks + boxes) \\
Rendering pipeline        & Unity HDRP + Perception \\
\bottomrule
\end{tabular}
\end{table}

\subsection{Additional Results}
\label{additional_results}

In this section, we present the implementation details and additional results to validate and support our proposed methods.

\subsubsection{\revised{Full Main-Results Grid}}
\label{sec:supp_full_grid}

\revised{\tableref{tab:main_results} in the main paper shows two backbones paired with
\lca{}. \tableref{tab:t2full_miou} to \tableref{tab:t2full_map} give the complete grid: nine
configurations over eight evaluation sets, one table per metric. Beyond the main-paper columns this adds
the frozen SAM and SAM~2.1 references, the decoder-only (\textbf{Dec}) and
\lca{}-only baselines that isolate where the gain originates, and the four ACDC conditions
evaluated in domain. \textbf{Dec} trains the mask decoder alone. \lca{} trains the adapter
with the decoder frozen. \lca{}+Dec trains both. The image encoder is frozen throughout.
Frozen reference columns are released weights in a single deterministic pass and carry no
seed spread.}

\revised{The in-domain ACDC rows here train and evaluate on the same condition, and are a
different experiment from the COCO-to-ACDC transfer reported in the main paper. Within ACDC
the clean-to-variant pair is synthetic \plap{} re-lighting applied on top of an already
adverse photograph, so the \textbf{C} column is the meaningful number on those rows.}


\begin{table*}[t]
\centering
\caption{Extended results, mIoU. Full grid of nine configurations over eight evaluation sets under clean (C) and lighting-variant (V) streams. Trained models are mean\,$\pm$\,population sd over three seeds; frozen released models are single deterministic passes and carry no spread. \textbf{Bold} marks the best value per row and stream.}
\label{tab:t2full_miou}
\scriptsize
\setlength{\tabcolsep}{3pt}
\renewcommand{\arraystretch}{1.1}
\begin{tabular}{@{}llccccccccc@{}}
\toprule
 & & \multicolumn{4}{c}{SAM} & \multicolumn{2}{c}{HQ-SAM} & \multicolumn{1}{c}{SAM 2.1} & \multicolumn{2}{c}{SAM 3} \\
\cmidrule(lr){3-6}\cmidrule(lr){7-8}\cmidrule(lr){9-9}\cmidrule(lr){10-11}
\textbf{Dataset} & & SAM & Dec & LCA & LCA+Dec & HQ-SAM & LCA+Dec & SAM\,2.1 & SAM\,3 & LCA+Dec \\
\midrule
\multirow{2}{*}{Cityscapes} & C & 0.5416 & 0.7700\,{\tiny$\pm$.0003} & 0.7574\,{\tiny$\pm$.0006} & 0.7845\,{\tiny$\pm$.0011} & 0.4737 & \textbf{0.7851}\,{\tiny$\pm$.0006} & 0.6084 & 0.6496 & 0.7662\,{\tiny$\pm$.0042} \\
 & V & 0.5359 & 0.7662\,{\tiny$\pm$.0003} & 0.7542\,{\tiny$\pm$.0010} & 0.7819\,{\tiny$\pm$.0011} & 0.4713 & \textbf{0.7820}\,{\tiny$\pm$.0008} & 0.6035 & 0.6486 & 0.7634\,{\tiny$\pm$.0042} \\
\addlinespace[2pt]
\multirow{2}{*}{COCO} & C & 0.6627 & 0.8122\,{\tiny$\pm$.0003} & 0.7946\,{\tiny$\pm$.0004} & 0.8166\,{\tiny$\pm$.0009} & 0.6585 & 0.8178\,{\tiny$\pm$.0011} & 0.7425 & 0.7607 & \textbf{0.8207}\,{\tiny$\pm$.0006} \\
 & V & 0.6584 & 0.8089\,{\tiny$\pm$.0001} & 0.7912\,{\tiny$\pm$.0007} & 0.8139\,{\tiny$\pm$.0008} & 0.6539 & 0.8153\,{\tiny$\pm$.0012} & 0.7397 & 0.7602 & \textbf{0.8196}\,{\tiny$\pm$.0006} \\
\addlinespace[2pt]
\multirow{2}{*}{VOC} & C & 0.6130 & 0.7342\,{\tiny$\pm$.0020} & 0.7129\,{\tiny$\pm$.0015} & 0.7575\,{\tiny$\pm$.0016} & 0.6131 & 0.7561\,{\tiny$\pm$.0012} & 0.6759 & 0.6918 & \textbf{0.7873}\,{\tiny$\pm$.0015} \\
 & V & 0.6076 & 0.7265\,{\tiny$\pm$.0016} & 0.7056\,{\tiny$\pm$.0013} & 0.7508\,{\tiny$\pm$.0012} & 0.6078 & 0.7498\,{\tiny$\pm$.0012} & 0.6726 & 0.6909 & \textbf{0.7838}\,{\tiny$\pm$.0014} \\
\addlinespace[2pt]
\multirow{2}{*}{Unity} & C & 0.6047 & 0.8226\,{\tiny$\pm$.0003} & 0.8419\,{\tiny$\pm$.0008} & 0.8764\,{\tiny$\pm$.0027} & 0.6047 & \textbf{0.8816}\,{\tiny$\pm$.0017} & 0.6880 & 0.7325 & 0.8679\,{\tiny$\pm$.0007} \\
 & V & 0.5967 & 0.8138\,{\tiny$\pm$.0003} & 0.8308\,{\tiny$\pm$.0005} & 0.8661\,{\tiny$\pm$.0024} & 0.6018 & \textbf{0.8722}\,{\tiny$\pm$.0003} & 0.6752 & 0.7284 & 0.8591\,{\tiny$\pm$.0004} \\
\midrule
\multirow{2}{*}{ACDC-fog} & C & 0.6589 & 0.8443\,{\tiny$\pm$.0005} & 0.8338\,{\tiny$\pm$.0007} & 0.8547\,{\tiny$\pm$.0007} & 0.6654 & \textbf{0.8585}\,{\tiny$\pm$.0009} & 0.7075 & 0.7393 & 0.8481\,{\tiny$\pm$.0021} \\
 & V & 0.6575 & 0.8414\,{\tiny$\pm$.0007} & 0.8299\,{\tiny$\pm$.0006} & 0.8505\,{\tiny$\pm$.0004} & 0.6639 & \textbf{0.8541}\,{\tiny$\pm$.0008} & 0.7039 & 0.7375 & 0.8455\,{\tiny$\pm$.0015} \\
\addlinespace[2pt]
\multirow{2}{*}{ACDC-night} & C & 0.4807 & 0.7943\,{\tiny$\pm$.0009} & 0.7785\,{\tiny$\pm$.0021} & 0.8043\,{\tiny$\pm$.0027} & 0.4995 & \textbf{0.8049}\,{\tiny$\pm$.0021} & 0.5942 & 0.6623 & 0.7875\,{\tiny$\pm$.0030} \\
 & V & 0.4822 & 0.7929\,{\tiny$\pm$.0009} & 0.7771\,{\tiny$\pm$.0017} & \textbf{0.8040}\,{\tiny$\pm$.0027} & 0.4943 & 0.8036\,{\tiny$\pm$.0015} & 0.5948 & 0.6613 & 0.7868\,{\tiny$\pm$.0027} \\
\addlinespace[2pt]
\multirow{2}{*}{ACDC-rain} & C & 0.5904 & 0.7888\,{\tiny$\pm$.0021} & 0.7801\,{\tiny$\pm$.0008} & 0.8036\,{\tiny$\pm$.0003} & 0.5957 & \textbf{0.8058}\,{\tiny$\pm$.0012} & 0.6644 & 0.6961 & 0.8005\,{\tiny$\pm$.0025} \\
 & V & 0.5890 & 0.7878\,{\tiny$\pm$.0020} & 0.7770\,{\tiny$\pm$.0016} & 0.8000\,{\tiny$\pm$.0014} & 0.5918 & \textbf{0.8024}\,{\tiny$\pm$.0012} & 0.6604 & 0.6943 & 0.7994\,{\tiny$\pm$.0022} \\
\addlinespace[2pt]
\multirow{2}{*}{ACDC-snow} & C & 0.6039 & 0.8329\,{\tiny$\pm$.0003} & 0.8164\,{\tiny$\pm$.0015} & 0.8417\,{\tiny$\pm$.0009} & 0.6083 & \textbf{0.8439}\,{\tiny$\pm$.0021} & 0.6724 & 0.7027 & 0.8260\,{\tiny$\pm$.0025} \\
 & V & 0.6001 & 0.8305\,{\tiny$\pm$.0010} & 0.8144\,{\tiny$\pm$.0008} & 0.8403\,{\tiny$\pm$.0002} & 0.6033 & \textbf{0.8416}\,{\tiny$\pm$.0022} & 0.6704 & 0.7041 & 0.8249\,{\tiny$\pm$.0032} \\
\bottomrule
\end{tabular}
\end{table*}

\begin{table*}[t]
\centering
\caption{Extended results, pmAP$_{50}$. Full grid of nine configurations over eight evaluation sets under clean (C) and lighting-variant (V) streams. Trained models are mean\,$\pm$\,population sd over three seeds; frozen released models are single deterministic passes and carry no spread. \textbf{Bold} marks the best value per row and stream.}
\label{tab:t2full_ap50}
\scriptsize
\setlength{\tabcolsep}{3pt}
\renewcommand{\arraystretch}{1.1}
\begin{tabular}{@{}llccccccccc@{}}
\toprule
 & & \multicolumn{4}{c}{SAM} & \multicolumn{2}{c}{HQ-SAM} & \multicolumn{1}{c}{SAM 2.1} & \multicolumn{2}{c}{SAM 3} \\
\cmidrule(lr){3-6}\cmidrule(lr){7-8}\cmidrule(lr){9-9}\cmidrule(lr){10-11}
\textbf{Dataset} & & SAM & Dec & LCA & LCA+Dec & HQ-SAM & LCA+Dec & SAM\,2.1 & SAM\,3 & LCA+Dec \\
\midrule
\multirow{2}{*}{Cityscapes} & C & 0.5982 & 0.9050\,{\tiny$\pm$.0023} & 0.8973\,{\tiny$\pm$.0020} & \textbf{0.9178}\,{\tiny$\pm$.0009} & 0.5224 & 0.9143\,{\tiny$\pm$.0026} & 0.6914 & 0.7623 & 0.9076\,{\tiny$\pm$.0053} \\
 & V & 0.5902 & 0.9023\,{\tiny$\pm$.0034} & 0.8933\,{\tiny$\pm$.0017} & \textbf{0.9177}\,{\tiny$\pm$.0002} & 0.5166 & 0.9112\,{\tiny$\pm$.0021} & 0.6794 & 0.7627 & 0.9030\,{\tiny$\pm$.0090} \\
\addlinespace[2pt]
\multirow{2}{*}{COCO} & C & 0.7796 & 0.9627\,{\tiny$\pm$.0012} & 0.9522\,{\tiny$\pm$.0006} & 0.9663\,{\tiny$\pm$.0014} & 0.7751 & \textbf{0.9674}\,{\tiny$\pm$.0011} & 0.8971 & 0.9264 & 0.9663\,{\tiny$\pm$.0004} \\
 & V & 0.7740 & 0.9609\,{\tiny$\pm$.0009} & 0.9498\,{\tiny$\pm$.0014} & 0.9648\,{\tiny$\pm$.0010} & 0.7692 & 0.9661\,{\tiny$\pm$.0003} & 0.8938 & 0.9242 & \textbf{0.9666}\,{\tiny$\pm$.0010} \\
\addlinespace[2pt]
\multirow{2}{*}{VOC} & C & 0.6935 & 0.8361\,{\tiny$\pm$.0029} & 0.8092\,{\tiny$\pm$.0023} & 0.8653\,{\tiny$\pm$.0025} & 0.6952 & 0.8653\,{\tiny$\pm$.0006} & 0.7742 & 0.7930 & \textbf{0.9000}\,{\tiny$\pm$.0018} \\
 & V & 0.6861 & 0.8260\,{\tiny$\pm$.0018} & 0.7984\,{\tiny$\pm$.0026} & 0.8598\,{\tiny$\pm$.0030} & 0.6841 & 0.8578\,{\tiny$\pm$.0027} & 0.7722 & 0.7927 & \textbf{0.8939}\,{\tiny$\pm$.0025} \\
\addlinespace[2pt]
\multirow{2}{*}{Unity} & C & 0.6393 & 0.9055\,{\tiny$\pm$.0004} & 0.9163\,{\tiny$\pm$.0012} & 0.9456\,{\tiny$\pm$.0004} & 0.6427 & \textbf{0.9493}\,{\tiny$\pm$.0022} & 0.7586 & 0.8192 & 0.9358\,{\tiny$\pm$.0004} \\
 & V & 0.6322 & 0.8990\,{\tiny$\pm$.0013} & 0.9055\,{\tiny$\pm$.0026} & 0.9404\,{\tiny$\pm$.0012} & 0.6398 & \textbf{0.9440}\,{\tiny$\pm$.0018} & 0.7395 & 0.8177 & 0.9303\,{\tiny$\pm$.0028} \\
\midrule
\multirow{2}{*}{ACDC-fog} & C & 0.8087 & 0.9725\,{\tiny$\pm$.0015} & 0.9725\,{\tiny$\pm$.0032} & 0.9751\,{\tiny$\pm$.0013} & 0.8134 & \textbf{0.9756}\,{\tiny$\pm$.0037} & 0.8694 & 0.9114 & 0.9751\,{\tiny$\pm$.0025} \\
 & V & 0.8180 & 0.9720\,{\tiny$\pm$.0025} & 0.9689\,{\tiny$\pm$.0022} & 0.9715\,{\tiny$\pm$.0015} & 0.8134 & \textbf{0.9736}\,{\tiny$\pm$.0046} & 0.8647 & 0.9114 & 0.9720\,{\tiny$\pm$.0013} \\
\addlinespace[2pt]
\multirow{2}{*}{ACDC-night} & C & 0.5161 & 0.9504\,{\tiny$\pm$.0008} & 0.9498\,{\tiny$\pm$.0015} & 0.9546\,{\tiny$\pm$.0030} & 0.5394 & \textbf{0.9558}\,{\tiny$\pm$.0022} & 0.6971 & 0.8351 & 0.9492\,{\tiny$\pm$.0030} \\
 & V & 0.5215 & 0.9528\,{\tiny$\pm$.0034} & 0.9486\,{\tiny$\pm$.0022} & 0.9558\,{\tiny$\pm$.0047} & 0.5179 & \textbf{0.9576}\,{\tiny$\pm$.0030} & 0.7079 & 0.8315 & 0.9546\,{\tiny$\pm$.0008} \\
\addlinespace[2pt]
\multirow{2}{*}{ACDC-rain} & C & 0.6645 & 0.9288\,{\tiny$\pm$.0049} & 0.9371\,{\tiny$\pm$.0021} & 0.9480\,{\tiny$\pm$.0059} & 0.6763 & \textbf{0.9498}\,{\tiny$\pm$.0041} & 0.8087 & 0.8545 & 0.9463\,{\tiny$\pm$.0028} \\
 & V & 0.6592 & 0.9284\,{\tiny$\pm$.0034} & 0.9336\,{\tiny$\pm$.0061} & 0.9450\,{\tiny$\pm$.0039} & 0.6632 & \textbf{0.9484}\,{\tiny$\pm$.0025} & 0.8021 & 0.8467 & 0.9471\,{\tiny$\pm$.0022} \\
\addlinespace[2pt]
\multirow{2}{*}{ACDC-snow} & C & 0.6996 & 0.9650\,{\tiny$\pm$.0020} & 0.9550\,{\tiny$\pm$.0006} & 0.9706\,{\tiny$\pm$.0021} & 0.7008 & \textbf{0.9723}\,{\tiny$\pm$.0026} & 0.8058 & 0.8432 & 0.9618\,{\tiny$\pm$.0021} \\
 & V & 0.6948 & 0.9610\,{\tiny$\pm$.0015} & 0.9582\,{\tiny$\pm$.0080} & 0.9650\,{\tiny$\pm$.0017} & 0.6912 & \textbf{0.9694}\,{\tiny$\pm$.0040} & 0.8118 & 0.8504 & 0.9590\,{\tiny$\pm$.0036} \\
\bottomrule
\end{tabular}
\end{table*}

\begin{table*}[t]
\centering
\caption{Extended results, pmAP$_{75}$. Full grid of nine configurations over eight evaluation sets under clean (C) and lighting-variant (V) streams. Trained models are mean\,$\pm$\,population sd over three seeds; frozen released models are single deterministic passes and carry no spread. \textbf{Bold} marks the best value per row and stream.}
\label{tab:t2full_ap75}
\scriptsize
\setlength{\tabcolsep}{3pt}
\renewcommand{\arraystretch}{1.1}
\begin{tabular}{@{}llccccccccc@{}}
\toprule
 & & \multicolumn{4}{c}{SAM} & \multicolumn{2}{c}{HQ-SAM} & \multicolumn{1}{c}{SAM 2.1} & \multicolumn{2}{c}{SAM 3} \\
\cmidrule(lr){3-6}\cmidrule(lr){7-8}\cmidrule(lr){9-9}\cmidrule(lr){10-11}
\textbf{Dataset} & & SAM & Dec & LCA & LCA+Dec & HQ-SAM & LCA+Dec & SAM\,2.1 & SAM\,3 & LCA+Dec \\
\midrule
\multirow{2}{*}{Cityscapes} & C & 0.2874 & 0.6405\,{\tiny$\pm$.0042} & 0.6231\,{\tiny$\pm$.0041} & \textbf{0.6807}\,{\tiny$\pm$.0029} & 0.2483 & 0.6782\,{\tiny$\pm$.0048} & 0.3157 & 0.4049 & 0.6262\,{\tiny$\pm$.0057} \\
 & V & 0.2785 & 0.6321\,{\tiny$\pm$.0020} & 0.6127\,{\tiny$\pm$.0079} & 0.6693\,{\tiny$\pm$.0018} & 0.2412 & \textbf{0.6718}\,{\tiny$\pm$.0044} & 0.3140 & 0.4053 & 0.6213\,{\tiny$\pm$.0070} \\
\addlinespace[2pt]
\multirow{2}{*}{COCO} & C & 0.4624 & 0.7652\,{\tiny$\pm$.0026} & 0.7201\,{\tiny$\pm$.0031} & 0.7778\,{\tiny$\pm$.0020} & 0.4512 & 0.7790\,{\tiny$\pm$.0031} & 0.5895 & 0.6342 & \textbf{0.7895}\,{\tiny$\pm$.0026} \\
 & V & 0.4505 & 0.7566\,{\tiny$\pm$.0014} & 0.7102\,{\tiny$\pm$.0011} & 0.7712\,{\tiny$\pm$.0017} & 0.4399 & 0.7695\,{\tiny$\pm$.0026} & 0.5816 & 0.6389 & \textbf{0.7873}\,{\tiny$\pm$.0020} \\
\addlinespace[2pt]
\multirow{2}{*}{VOC} & C & 0.4383 & 0.6150\,{\tiny$\pm$.0036} & 0.5773\,{\tiny$\pm$.0032} & 0.6384\,{\tiny$\pm$.0018} & 0.4339 & 0.6410\,{\tiny$\pm$.0010} & 0.5552 & 0.5948 & \textbf{0.6945}\,{\tiny$\pm$.0053} \\
 & V & 0.4281 & 0.5997\,{\tiny$\pm$.0043} & 0.5665\,{\tiny$\pm$.0041} & 0.6270\,{\tiny$\pm$.0011} & 0.4313 & 0.6328\,{\tiny$\pm$.0024} & 0.5490 & 0.5916 & \textbf{0.6895}\,{\tiny$\pm$.0045} \\
\addlinespace[2pt]
\multirow{2}{*}{Unity} & C & 0.4480 & 0.7533\,{\tiny$\pm$.0004} & 0.7918\,{\tiny$\pm$.0032} & 0.8511\,{\tiny$\pm$.0032} & 0.4399 & \textbf{0.8569}\,{\tiny$\pm$.0040} & 0.5472 & 0.6202 & 0.8360\,{\tiny$\pm$.0023} \\
 & V & 0.4475 & 0.7395\,{\tiny$\pm$.0030} & 0.7705\,{\tiny$\pm$.0014} & 0.8282\,{\tiny$\pm$.0041} & 0.4447 & \textbf{0.8408}\,{\tiny$\pm$.0013} & 0.5281 & 0.6088 & 0.8259\,{\tiny$\pm$.0017} \\
\midrule
\multirow{2}{*}{ACDC-fog} & C & 0.4090 & 0.8196\,{\tiny$\pm$.0000} & 0.7957\,{\tiny$\pm$.0019} & 0.8419\,{\tiny$\pm$.0007} & 0.4090 & \textbf{0.8445}\,{\tiny$\pm$.0046} & 0.4930 & 0.5801 & 0.8326\,{\tiny$\pm$.0039} \\
 & V & 0.4121 & 0.8082\,{\tiny$\pm$.0041} & 0.7859\,{\tiny$\pm$.0026} & \textbf{0.8305}\,{\tiny$\pm$.0067} & 0.4168 & 0.8258\,{\tiny$\pm$.0034} & 0.4837 & 0.5677 & 0.8279\,{\tiny$\pm$.0041} \\
\addlinespace[2pt]
\multirow{2}{*}{ACDC-night} & C & 0.1882 & 0.6577\,{\tiny$\pm$.0053} & 0.6362\,{\tiny$\pm$.0044} & \textbf{0.6971}\,{\tiny$\pm$.0053} & 0.1989 & 0.6935\,{\tiny$\pm$.0089} & 0.2670 & 0.3978 & 0.6625\,{\tiny$\pm$.0089} \\
 & V & 0.1864 & 0.6523\,{\tiny$\pm$.0077} & 0.6326\,{\tiny$\pm$.0029} & \textbf{0.6947}\,{\tiny$\pm$.0066} & 0.1935 & 0.6822\,{\tiny$\pm$.0166} & 0.2634 & 0.3835 & 0.6637\,{\tiny$\pm$.0118} \\
\addlinespace[2pt]
\multirow{2}{*}{ACDC-rain} & C & 0.3001 & 0.6846\,{\tiny$\pm$.0051} & 0.6654\,{\tiny$\pm$.0034} & 0.7104\,{\tiny$\pm$.0043} & 0.2988 & 0.7090\,{\tiny$\pm$.0011} & 0.4024 & 0.4784 & \textbf{0.7134}\,{\tiny$\pm$.0083} \\
 & V & 0.2988 & 0.6828\,{\tiny$\pm$.0049} & 0.6509\,{\tiny$\pm$.0073} & 0.7003\,{\tiny$\pm$.0071} & 0.2896 & 0.6990\,{\tiny$\pm$.0075} & 0.3958 & 0.4718 & \textbf{0.7086}\,{\tiny$\pm$.0022} \\
\addlinespace[2pt]
\multirow{2}{*}{ACDC-snow} & C & 0.3486 & 0.7853\,{\tiny$\pm$.0045} & 0.7515\,{\tiny$\pm$.0010} & \textbf{0.8066}\,{\tiny$\pm$.0006} & 0.3498 & 0.8054\,{\tiny$\pm$.0037} & 0.4210 & 0.5006 & 0.7696\,{\tiny$\pm$.0052} \\
 & V & 0.3378 & 0.7772\,{\tiny$\pm$.0080} & 0.7511\,{\tiny$\pm$.0015} & \textbf{0.8010}\,{\tiny$\pm$.0081} & 0.3366 & 0.7961\,{\tiny$\pm$.0069} & 0.4089 & 0.5030 & 0.7676\,{\tiny$\pm$.0077} \\
\bottomrule
\end{tabular}
\end{table*}

\begin{table*}[t]
\centering
\caption{Extended results, pmAP. Full grid of nine configurations over eight evaluation sets under clean (C) and lighting-variant (V) streams. Trained models are mean\,$\pm$\,population sd over three seeds; frozen released models are single deterministic passes and carry no spread. \textbf{Bold} marks the best value per row and stream.}
\label{tab:t2full_map}
\scriptsize
\setlength{\tabcolsep}{3pt}
\renewcommand{\arraystretch}{1.1}
\begin{tabular}{@{}llccccccccc@{}}
\toprule
 & & \multicolumn{4}{c}{SAM} & \multicolumn{2}{c}{HQ-SAM} & \multicolumn{1}{c}{SAM 2.1} & \multicolumn{2}{c}{SAM 3} \\
\cmidrule(lr){3-6}\cmidrule(lr){7-8}\cmidrule(lr){9-9}\cmidrule(lr){10-11}
\textbf{Dataset} & & SAM & Dec & LCA & LCA+Dec & HQ-SAM & LCA+Dec & SAM\,2.1 & SAM\,3 & LCA+Dec \\
\midrule
\multirow{2}{*}{Cityscapes} & C & 0.3141 & 0.6137\,{\tiny$\pm$.0008} & 0.5965\,{\tiny$\pm$.0008} & 0.6399\,{\tiny$\pm$.0021} & 0.2757 & \textbf{0.6417}\,{\tiny$\pm$.0004} & 0.3585 & 0.4244 & 0.6097\,{\tiny$\pm$.0077} \\
 & V & 0.3077 & 0.6071\,{\tiny$\pm$.0004} & 0.5909\,{\tiny$\pm$.0021} & 0.6350\,{\tiny$\pm$.0017} & 0.2711 & \textbf{0.6356}\,{\tiny$\pm$.0008} & 0.3528 & 0.4222 & 0.6052\,{\tiny$\pm$.0080} \\
\addlinespace[2pt]
\multirow{2}{*}{COCO} & C & 0.4519 & 0.6823\,{\tiny$\pm$.0006} & 0.6501\,{\tiny$\pm$.0011} & 0.6910\,{\tiny$\pm$.0019} & 0.4473 & 0.6934\,{\tiny$\pm$.0018} & 0.5582 & 0.5895 & \textbf{0.7007}\,{\tiny$\pm$.0008} \\
 & V & 0.4452 & 0.6763\,{\tiny$\pm$.0002} & 0.6442\,{\tiny$\pm$.0015} & 0.6859\,{\tiny$\pm$.0017} & 0.4404 & 0.6882\,{\tiny$\pm$.0019} & 0.5532 & 0.5894 & \textbf{0.6984}\,{\tiny$\pm$.0008} \\
\addlinespace[2pt]
\multirow{2}{*}{VOC} & C & 0.4356 & 0.5803\,{\tiny$\pm$.0025} & 0.5477\,{\tiny$\pm$.0040} & 0.6093\,{\tiny$\pm$.0012} & 0.4318 & 0.6123\,{\tiny$\pm$.0013} & 0.5300 & 0.5584 & \textbf{0.6606}\,{\tiny$\pm$.0023} \\
 & V & 0.4267 & 0.5683\,{\tiny$\pm$.0022} & 0.5370\,{\tiny$\pm$.0031} & 0.5984\,{\tiny$\pm$.0008} & 0.4244 & 0.6019\,{\tiny$\pm$.0021} & 0.5245 & 0.5563 & \textbf{0.6549}\,{\tiny$\pm$.0025} \\
\addlinespace[2pt]
\multirow{2}{*}{Unity} & C & 0.4385 & 0.7279\,{\tiny$\pm$.0012} & 0.7630\,{\tiny$\pm$.0017} & 0.8193\,{\tiny$\pm$.0043} & 0.4356 & \textbf{0.8264}\,{\tiny$\pm$.0034} & 0.5313 & 0.5935 & 0.8043\,{\tiny$\pm$.0012} \\
 & V & 0.4317 & 0.7133\,{\tiny$\pm$.0008} & 0.7441\,{\tiny$\pm$.0002} & 0.8014\,{\tiny$\pm$.0034} & 0.4328 & \textbf{0.8089}\,{\tiny$\pm$.0006} & 0.5165 & 0.5862 & 0.7903\,{\tiny$\pm$.0012} \\
\midrule
\multirow{2}{*}{ACDC-fog} & C & 0.4331 & 0.7432\,{\tiny$\pm$.0012} & 0.7250\,{\tiny$\pm$.0016} & 0.7647\,{\tiny$\pm$.0009} & 0.4384 & \textbf{0.7715}\,{\tiny$\pm$.0019} & 0.5034 & 0.5552 & 0.7555\,{\tiny$\pm$.0030} \\
 & V & 0.4319 & 0.7381\,{\tiny$\pm$.0008} & 0.7177\,{\tiny$\pm$.0011} & 0.7569\,{\tiny$\pm$.0003} & 0.4387 & \textbf{0.7622}\,{\tiny$\pm$.0027} & 0.4988 & 0.5507 & 0.7499\,{\tiny$\pm$.0017} \\
\addlinespace[2pt]
\multirow{2}{*}{ACDC-night} & C & 0.2240 & 0.6423\,{\tiny$\pm$.0010} & 0.6154\,{\tiny$\pm$.0038} & 0.6619\,{\tiny$\pm$.0053} & 0.2364 & \textbf{0.6633}\,{\tiny$\pm$.0035} & 0.3237 & 0.4228 & 0.6388\,{\tiny$\pm$.0065} \\
 & V & 0.2262 & 0.6399\,{\tiny$\pm$.0010} & 0.6114\,{\tiny$\pm$.0036} & \textbf{0.6616}\,{\tiny$\pm$.0054} & 0.2341 & 0.6614\,{\tiny$\pm$.0038} & 0.3224 & 0.4217 & 0.6372\,{\tiny$\pm$.0059} \\
\addlinespace[2pt]
\multirow{2}{*}{ACDC-rain} & C & 0.3394 & 0.6420\,{\tiny$\pm$.0030} & 0.6261\,{\tiny$\pm$.0011} & 0.6678\,{\tiny$\pm$.0003} & 0.3408 & \textbf{0.6721}\,{\tiny$\pm$.0025} & 0.4290 & 0.4798 & 0.6672\,{\tiny$\pm$.0047} \\
 & V & 0.3376 & 0.6394\,{\tiny$\pm$.0033} & 0.6202\,{\tiny$\pm$.0032} & 0.6608\,{\tiny$\pm$.0029} & 0.3367 & \textbf{0.6657}\,{\tiny$\pm$.0027} & 0.4250 & 0.4779 & 0.6643\,{\tiny$\pm$.0033} \\
\addlinespace[2pt]
\multirow{2}{*}{ACDC-snow} & C & 0.3662 & 0.7222\,{\tiny$\pm$.0016} & 0.6936\,{\tiny$\pm$.0027} & 0.7402\,{\tiny$\pm$.0020} & 0.3669 & \textbf{0.7427}\,{\tiny$\pm$.0040} & 0.4431 & 0.4973 & 0.7156\,{\tiny$\pm$.0049} \\
 & V & 0.3620 & 0.7183\,{\tiny$\pm$.0024} & 0.6894\,{\tiny$\pm$.0012} & 0.7366\,{\tiny$\pm$.0007} & 0.3638 & \textbf{0.7385}\,{\tiny$\pm$.0048} & 0.4397 & 0.4987 & 0.7137\,{\tiny$\pm$.0061} \\
\bottomrule
\end{tabular}
\end{table*}

\subsubsection{\revised{Resolution Sweep: AP Metrics}}
\label{sec:supp_resolution}

\revised{\tableref{tab:image_size_ablation} reports mIoU for the resolution sweep. \tableref{tab:resolution_ap} gives the same sweep for pmAP$_{50}$,
pmAP$_{75}$ and pmAP. The AP metrics also generally improve with resolution on both streams, with some fluctuations at higher resolutions. Of the two axes, the encoder axis is the cleaner curve:
on the data axis, the masks are still rasterised at the loader resolution, so part of the
remaining movement is rasterisation fidelity rather than model behaviour.}

\begin{table}[htbp]
\centering
\caption{\revised{Input resolution on VOC, mIoU. The \textbf{data axis} varies the loader while the encoder still receives $4{,}096$ tokens. The \textbf{encoder axis} varies the encoder input, so the token grid shrinks. Validation resolution is fixed, so all rows score the same instances. $^{*}$Encoder-axis 1024 is data-axis 512 by construction. AP metrics in \tableref{tab:resolution_ap}.}}
\label{tab:image_size_ablation}
\footnotesize
\setlength{\tabcolsep}{2.2pt}
\begin{tabular}{@{}l|cccc|cccc@{}}
\toprule
 & \multicolumn{4}{c|}{\revised{Data axis}} & \multicolumn{4}{c}{\revised{Encoder axis}} \\
\cmidrule(lr){2-5}\cmidrule(lr){6-9}
 & \multicolumn{2}{c}{\revised{SAM}} & \multicolumn{2}{c|}{\revised{+\lca{}}} & \multicolumn{2}{c}{\revised{SAM}} & \multicolumn{2}{c}{\revised{+\lca{}}} \\
\cmidrule(lr){2-3}\cmidrule(lr){4-5}\cmidrule(lr){6-7}\cmidrule(lr){8-9}
\revised{Size} & \revised{C} & \revised{V} & \revised{C} & \revised{V} & \revised{C} & \revised{V} & \revised{C} & \revised{V} \\
\midrule
\revised{128} & \revised{0.5520} & \revised{0.5472} & \revised{0.7106} & \revised{0.7047} & \revised{0.3849} & \revised{0.3772} & \revised{0.5673} & \revised{0.5628} \\
\revised{256} & \revised{0.6034} & \revised{0.5984} & \revised{0.7380} & \revised{0.7317} & \revised{0.5471} & \revised{0.5423} & \revised{0.6638} & \revised{0.6605} \\
\revised{512} & \revised{0.6130} & \revised{0.6076} & \revised{0.7456} & \revised{0.7380} & \revised{0.6172} & \revised{0.6122} & \revised{0.7211} & \revised{0.7175} \\
\revised{1024$^{*}$} & \revised{0.6108} & \revised{0.6047} & \revised{0.7429} & \revised{0.7344} & \revised{0.6130} & \revised{0.6076} & \revised{0.7456} & \revised{0.7380} \\
\bottomrule
\end{tabular}
\end{table}

\begin{table*}[t]
\centering
\caption{\revised{Input-resolution sweep on VOC, AP metrics. Same protocol and axes as \tableref{tab:image_size_ablation}, which reports mIoU. Higher resolutions generally improve AP metrics on both axes and both streams, with some fluctuations at higher resolutions. $^{*}$Encoder-axis 1024 is data-axis 512 by construction.}}
\label{tab:resolution_ap}
\footnotesize
\setlength{\tabcolsep}{5pt}
\renewcommand{\arraystretch}{1.05}
\begin{tabular}{@{}ll|cccc|cccc@{}}
\toprule
 & & \multicolumn{4}{c|}{\textbf{Data axis} (loader resolution)} & \multicolumn{4}{c}{\textbf{Encoder axis} (encoder input)} \\
\cmidrule(lr){3-6}\cmidrule(lr){7-10}
 & & \multicolumn{2}{c}{SAM} & \multicolumn{2}{c|}{+\lca{}} & \multicolumn{2}{c}{SAM} & \multicolumn{2}{c}{+\lca{}} \\
\cmidrule(lr){3-4}\cmidrule(lr){5-6}\cmidrule(lr){7-8}\cmidrule(lr){9-10}
\textbf{Metric} & \textbf{Size} & C & V & C & V & C & V & C & V \\
\midrule
\multirow{4}{*}{pmAP$_{50}$}
  & 128 & 0.6124 & 0.6066 & 0.8095 & 0.8027 & 0.3453 & 0.3324 & 0.6289 & 0.6236 \\
  & 256 & 0.6794 & 0.6753 & 0.8406 & 0.8332 & 0.5887 & 0.5769 & 0.7355 & 0.7317 \\
  & 512 & 0.6935 & 0.6861 & 0.8503 & 0.8417 & 0.6870 & 0.6820 & 0.8206 & 0.8136 \\
  & 1024$^{*}$ & 0.6944 & 0.6835 & 0.8447 & 0.8353 & 0.6935 & 0.6861 & 0.8503 & 0.8417 \\
\midrule
\multirow{4}{*}{pmAP$_{75}$}
  & 128 & 0.3306 & 0.3250 & 0.5681 & 0.5578 & 0.1609 & 0.1550 & 0.3647 & 0.3529 \\
  & 256 & 0.4237 & 0.4143 & 0.6130 & 0.6016 & 0.3600 & 0.3541 & 0.5038 & 0.4982 \\
  & 512 & 0.4383 & 0.4281 & 0.6312 & 0.6169 & 0.4595 & 0.4533 & 0.6025 & 0.5919 \\
  & 1024$^{*}$ & 0.4336 & 0.4240 & 0.6318 & 0.6195 & 0.4383 & 0.4281 & 0.6312 & 0.6169 \\
\midrule
\multirow{4}{*}{pmAP}
  & 128 & 0.3414 & 0.3352 & 0.5417 & 0.5321 & 0.1754 & 0.1680 & 0.3653 & 0.3583 \\
  & 256 & 0.4210 & 0.4138 & 0.5840 & 0.5733 & 0.3566 & 0.3518 & 0.4858 & 0.4805 \\
  & 512 & 0.4356 & 0.4267 & 0.5973 & 0.5853 & 0.4459 & 0.4407 & 0.5698 & 0.5619 \\
  & 1024$^{*}$ & 0.4312 & 0.4222 & 0.5957 & 0.5831 & 0.4356 & 0.4267 & 0.5973 & 0.5853 \\
\bottomrule
\end{tabular}
\end{table*}

\subsubsection{\revised{Full Severity Sweep}}
\label{sec:supp_severity}

\revised{\tableref{tab:robustness} (a) in the main paper reports relative degradation at
severity~3. \tableref{tab:severity_full} gives absolute scores at every severity for all
eight evaluation sets and four models. The performance generally declines with increasing severity, with some non-monotonic variation, particularly on Unity. \lca{}+Dec holds
the highest absolute score at every severity on all four core datasets.}

\begin{table*}[t]
\centering
\caption{\revised{Severity sweep, absolute scores (higher is better). \texttt{sev1/2/3} are the validation images re-lit at a fixed mild/moderate/severe \plap{} setting; \textbf{clean} is byte-identical across the three and is reported once per row as a control. Trained models are mean\,$\pm$\,population sd over three seeds; SAM is the frozen released model. Within ACDC the clean-to-variant pair is synthetic re-lighting applied on top of an already adverse photograph, not ACDC's own clear/adverse correspondence, so the \textbf{clean} column is the meaningful number on those rows.}}
\label{tab:severity_full}
\scriptsize
\setlength{\tabcolsep}{3pt}
\renewcommand{\arraystretch}{1.05}
\begin{tabular}{@{}ll|cccc|cccc@{}}
\toprule
 & & \multicolumn{4}{c|}{mIoU} & \multicolumn{4}{c}{pmAP$_{50}$} \\
\cmidrule(lr){3-6}\cmidrule(lr){7-10}
\textbf{Dataset} & \textbf{Method} & clean & sev1 & sev2 & sev3 & clean & sev1 & sev2 & sev3 \\
\midrule
\multirow{4}{*}{Cityscapes} & SAM & 0.5416 & 0.5393 & 0.5391 & 0.5315 & 0.5982 & 0.5920 & 0.5916 & 0.5836 \\
 & Dec & 0.7700\,{\tiny$\pm$.0003} & 0.7692\,{\tiny$\pm$.0002} & 0.7680\,{\tiny$\pm$.0007} & 0.7627\,{\tiny$\pm$.0008} & 0.9050\,{\tiny$\pm$.0023} & 0.9060\,{\tiny$\pm$.0023} & 0.9052\,{\tiny$\pm$.0008} & 0.8993\,{\tiny$\pm$.0007} \\
 & LCA & 0.7574\,{\tiny$\pm$.0006} & 0.7573\,{\tiny$\pm$.0012} & 0.7560\,{\tiny$\pm$.0005} & 0.7517\,{\tiny$\pm$.0003} & 0.8973\,{\tiny$\pm$.0020} & 0.8968\,{\tiny$\pm$.0031} & 0.8973\,{\tiny$\pm$.0024} & 0.8931\,{\tiny$\pm$.0032} \\
 & LCA+Dec & 0.7845\,{\tiny$\pm$.0011} & 0.7850\,{\tiny$\pm$.0013} & 0.7836\,{\tiny$\pm$.0017} & 0.7787\,{\tiny$\pm$.0011} & 0.9178\,{\tiny$\pm$.0009} & 0.9193\,{\tiny$\pm$.0010} & 0.9168\,{\tiny$\pm$.0009} & 0.9152\,{\tiny$\pm$.0013} \\
\midrule
\multirow{4}{*}{COCO} & SAM & 0.6627 & 0.6627 & 0.6593 & 0.6511 & 0.7796 & 0.7812 & 0.7749 & 0.7621 \\
 & Dec & 0.8122\,{\tiny$\pm$.0003} & 0.8117\,{\tiny$\pm$.0002} & 0.8100\,{\tiny$\pm$.0002} & 0.8049\,{\tiny$\pm$.0001} & 0.9627\,{\tiny$\pm$.0012} & 0.9620\,{\tiny$\pm$.0008} & 0.9614\,{\tiny$\pm$.0004} & 0.9575\,{\tiny$\pm$.0007} \\
 & LCA & 0.7946\,{\tiny$\pm$.0004} & 0.7942\,{\tiny$\pm$.0004} & 0.7927\,{\tiny$\pm$.0006} & 0.7871\,{\tiny$\pm$.0008} & 0.9522\,{\tiny$\pm$.0006} & 0.9526\,{\tiny$\pm$.0006} & 0.9509\,{\tiny$\pm$.0010} & 0.9444\,{\tiny$\pm$.0016} \\
 & LCA+Dec & 0.8166\,{\tiny$\pm$.0009} & 0.8162\,{\tiny$\pm$.0009} & 0.8149\,{\tiny$\pm$.0010} & 0.8098\,{\tiny$\pm$.0012} & 0.9663\,{\tiny$\pm$.0014} & 0.9653\,{\tiny$\pm$.0004} & 0.9654\,{\tiny$\pm$.0001} & 0.9612\,{\tiny$\pm$.0008} \\
\midrule
\multirow{4}{*}{VOC} & SAM & 0.6130 & 0.6136 & 0.6114 & 0.6020 & 0.6935 & 0.6935 & 0.6926 & 0.6782 \\
 & Dec & 0.7342\,{\tiny$\pm$.0020} & 0.7334\,{\tiny$\pm$.0021} & 0.7301\,{\tiny$\pm$.0025} & 0.7171\,{\tiny$\pm$.0021} & 0.8361\,{\tiny$\pm$.0029} & 0.8347\,{\tiny$\pm$.0030} & 0.8315\,{\tiny$\pm$.0036} & 0.8137\,{\tiny$\pm$.0036} \\
 & LCA & 0.7129\,{\tiny$\pm$.0015} & 0.7114\,{\tiny$\pm$.0013} & 0.7096\,{\tiny$\pm$.0019} & 0.6999\,{\tiny$\pm$.0016} & 0.8092\,{\tiny$\pm$.0023} & 0.8058\,{\tiny$\pm$.0018} & 0.8061\,{\tiny$\pm$.0016} & 0.7915\,{\tiny$\pm$.0033} \\
 & LCA+Dec & 0.7575\,{\tiny$\pm$.0016} & 0.7567\,{\tiny$\pm$.0012} & 0.7542\,{\tiny$\pm$.0013} & 0.7448\,{\tiny$\pm$.0009} & 0.8653\,{\tiny$\pm$.0025} & 0.8662\,{\tiny$\pm$.0016} & 0.8616\,{\tiny$\pm$.0017} & 0.8528\,{\tiny$\pm$.0006} \\
\midrule
\multirow{4}{*}{Unity} & SAM & 0.6047 & 0.5988 & 0.5920 & 0.5985 & 0.6393 & 0.6365 & 0.6274 & 0.6355 \\
 & Dec & 0.8226\,{\tiny$\pm$.0003} & 0.8172\,{\tiny$\pm$.0011} & 0.8108\,{\tiny$\pm$.0002} & 0.8136\,{\tiny$\pm$.0003} & 0.9055\,{\tiny$\pm$.0004} & 0.9020\,{\tiny$\pm$.0046} & 0.8974\,{\tiny$\pm$.0007} & 0.8982\,{\tiny$\pm$.0018} \\
 & LCA & 0.8419\,{\tiny$\pm$.0008} & 0.8366\,{\tiny$\pm$.0005} & 0.8278\,{\tiny$\pm$.0013} & 0.8329\,{\tiny$\pm$.0011} & 0.9163\,{\tiny$\pm$.0012} & 0.9122\,{\tiny$\pm$.0010} & 0.9036\,{\tiny$\pm$.0018} & 0.9066\,{\tiny$\pm$.0012} \\
 & LCA+Dec & 0.8764\,{\tiny$\pm$.0027} & 0.8704\,{\tiny$\pm$.0027} & 0.8630\,{\tiny$\pm$.0025} & 0.8684\,{\tiny$\pm$.0028} & 0.9456\,{\tiny$\pm$.0004} & 0.9416\,{\tiny$\pm$.0006} & 0.9396\,{\tiny$\pm$.0013} & 0.9416\,{\tiny$\pm$.0014} \\
\midrule
\multirow{4}{*}{ACDC-fog} & SAM & 0.6589 & 0.6614 & 0.6590 & 0.6517 & 0.8087 & 0.8149 & 0.8056 & 0.7994 \\
 & Dec & 0.8443\,{\tiny$\pm$.0005} & 0.8444\,{\tiny$\pm$.0004} & 0.8441\,{\tiny$\pm$.0007} & 0.8392\,{\tiny$\pm$.0003} & 0.9725\,{\tiny$\pm$.0015} & 0.9746\,{\tiny$\pm$.0015} & 0.9705\,{\tiny$\pm$.0034} & 0.9673\,{\tiny$\pm$.0022} \\
 & LCA & 0.8338\,{\tiny$\pm$.0007} & 0.8337\,{\tiny$\pm$.0006} & 0.8327\,{\tiny$\pm$.0006} & 0.8269\,{\tiny$\pm$.0014} & 0.9725\,{\tiny$\pm$.0032} & 0.9715\,{\tiny$\pm$.0019} & 0.9730\,{\tiny$\pm$.0015} & 0.9689\,{\tiny$\pm$.0025} \\
 & LCA+Dec & 0.8547\,{\tiny$\pm$.0007} & 0.8537\,{\tiny$\pm$.0003} & 0.8533\,{\tiny$\pm$.0003} & 0.8488\,{\tiny$\pm$.0008} & 0.9751\,{\tiny$\pm$.0013} & 0.9730\,{\tiny$\pm$.0007} & 0.9736\,{\tiny$\pm$.0022} & 0.9720\,{\tiny$\pm$.0034} \\
\midrule
\multirow{4}{*}{ACDC-night} & SAM & 0.4807 & 0.4841 & 0.4812 & 0.4801 & 0.5161 & 0.5233 & 0.5143 & 0.5108 \\
 & Dec & 0.7943\,{\tiny$\pm$.0009} & 0.7941\,{\tiny$\pm$.0017} & 0.7920\,{\tiny$\pm$.0020} & 0.7908\,{\tiny$\pm$.0012} & 0.9504\,{\tiny$\pm$.0008} & 0.9504\,{\tiny$\pm$.0045} & 0.9522\,{\tiny$\pm$.0055} & 0.9480\,{\tiny$\pm$.0051} \\
 & LCA & 0.7785\,{\tiny$\pm$.0021} & 0.7798\,{\tiny$\pm$.0020} & 0.7786\,{\tiny$\pm$.0013} & 0.7727\,{\tiny$\pm$.0016} & 0.9498\,{\tiny$\pm$.0015} & 0.9516\,{\tiny$\pm$.0015} & 0.9486\,{\tiny$\pm$.0042} & 0.9474\,{\tiny$\pm$.0017} \\
 & LCA+Dec & 0.8043\,{\tiny$\pm$.0027} & 0.8040\,{\tiny$\pm$.0027} & 0.8032\,{\tiny$\pm$.0029} & 0.8014\,{\tiny$\pm$.0030} & 0.9546\,{\tiny$\pm$.0030} & 0.9564\,{\tiny$\pm$.0045} & 0.9498\,{\tiny$\pm$.0039} & 0.9510\,{\tiny$\pm$.0030} \\
\midrule
\multirow{4}{*}{ACDC-rain} & SAM & 0.5904 & 0.5905 & 0.5896 & 0.5845 & 0.6645 & 0.6632 & 0.6527 & 0.6501 \\
 & Dec & 0.7888\,{\tiny$\pm$.0021} & 0.7888\,{\tiny$\pm$.0025} & 0.7878\,{\tiny$\pm$.0025} & 0.7838\,{\tiny$\pm$.0016} & 0.9288\,{\tiny$\pm$.0049} & 0.9297\,{\tiny$\pm$.0045} & 0.9279\,{\tiny$\pm$.0054} & 0.9249\,{\tiny$\pm$.0051} \\
 & LCA & 0.7801\,{\tiny$\pm$.0008} & 0.7795\,{\tiny$\pm$.0011} & 0.7783\,{\tiny$\pm$.0022} & 0.7719\,{\tiny$\pm$.0019} & 0.9371\,{\tiny$\pm$.0021} & 0.9362\,{\tiny$\pm$.0006} & 0.9336\,{\tiny$\pm$.0033} & 0.9279\,{\tiny$\pm$.0037} \\
 & LCA+Dec & 0.8036\,{\tiny$\pm$.0003} & 0.8027\,{\tiny$\pm$.0001} & 0.8016\,{\tiny$\pm$.0004} & 0.7975\,{\tiny$\pm$.0002} & 0.9480\,{\tiny$\pm$.0059} & 0.9463\,{\tiny$\pm$.0049} & 0.9458\,{\tiny$\pm$.0034} & 0.9419\,{\tiny$\pm$.0033} \\
\midrule
\multirow{4}{*}{ACDC-snow} & SAM & 0.6039 & 0.6041 & 0.6021 & 0.5992 & 0.6996 & 0.6948 & 0.6936 & 0.6876 \\
 & Dec & 0.8329\,{\tiny$\pm$.0003} & 0.8317\,{\tiny$\pm$.0004} & 0.8312\,{\tiny$\pm$.0006} & 0.8302\,{\tiny$\pm$.0009} & 0.9650\,{\tiny$\pm$.0020} & 0.9622\,{\tiny$\pm$.0015} & 0.9622\,{\tiny$\pm$.0015} & 0.9634\,{\tiny$\pm$.0015} \\
 & LCA & 0.8164\,{\tiny$\pm$.0015} & 0.8168\,{\tiny$\pm$.0012} & 0.8175\,{\tiny$\pm$.0007} & 0.8148\,{\tiny$\pm$.0011} & 0.9550\,{\tiny$\pm$.0006} & 0.9590\,{\tiny$\pm$.0020} & 0.9578\,{\tiny$\pm$.0026} & 0.9598\,{\tiny$\pm$.0025} \\
 & LCA+Dec & 0.8417\,{\tiny$\pm$.0009} & 0.8413\,{\tiny$\pm$.0007} & 0.8416\,{\tiny$\pm$.0009} & 0.8406\,{\tiny$\pm$.0008} & 0.9706\,{\tiny$\pm$.0021} & 0.9694\,{\tiny$\pm$.0025} & 0.9694\,{\tiny$\pm$.0025} & 0.9674\,{\tiny$\pm$.0036} \\
\bottomrule
\end{tabular}
\end{table*}

\subsubsection{\revised{Evaluation Protocol and Metrics}}
\label{sec:eval_protocol}

\revised{All results in this paper use ground-truth prompts, and we state the protocol in
full so the numbers are read for what they measure. For each annotated instance we sample one
foreground point uniformly at random from the mask interior and take the tight ground-truth
box. We add a single background point, placed at the image corner farthest from the box
centre. The prompt encoder receives the point pair and the box together. The decoder is run
with multi-mask output, which produces three candidate masks, and we keep the candidate with
the highest score from SAM's own IoU-prediction head. Masks are compared in a fixed
$256{\times}256$ space and binarised at $0.5$, so the encoder-resolution study in
\secref{sec:ablation} changes only the encoder input and nothing downstream. The identical
prompts drive the clean and the variant stream, so any difference between the two streams
comes from the encoder rather than from the prompt.}

\revised{This yields exactly one prediction per ground-truth instance. Predictions are not
ranked by confidence, instances are never proposed or discovered, duplicates cannot arise,
and no category is predicted. The quantities usually folded into detection AP, such as
false negatives from missed instances and class-aware matching, therefore have no role here.
We name the metric \emph{prompted mask AP} (\text{pmAP}) to make the setting explicit, and
define it as the fraction of instances whose decoded mask reaches an IoU threshold,}
\begin{equation}
\label{eq:pmap}
\revised{\begin{aligned}
\text{pmAP}_{t} &= \frac{1}{K}\sum_{k=1}^{K}\mathbf{1}\!\left[\text{IoU}(p_k, m_k) \ge t\right], \\[2pt]
\text{pmAP} &= \frac{1}{10}\sum_{t \in \{0.50,\, 0.55,\, \ldots,\, 0.95\}} \text{pmAP}_{t},
\end{aligned}}
\end{equation}
\revised{averaged over the same ten thresholds COCO AP uses. \text{pmAP}$_{50}$ and
\text{pmAP}$_{75}$ are the single-threshold values, and mIoU is the mean of
$\text{IoU}(p_k, m_k)$ over the same instances. The paper therefore studies
\emph{prompted mask prediction under illumination change}. It measures how well a frozen
backbone segments an object it has already been pointed at, and how much that degrades when
the lighting changes. Instance discovery and classification are outside this setting, so
\text{pmAP} is the appropriate point of comparison for our numbers.}

\subsubsection{\revised{Gate Composition and Loss Weights}}
\label{sec:supp_design}

\revised{\tableref{tab:lsa_version_ablation} compares design variants
with different fusion strategies, projections, and parameter counts.
Sequential and additive variants achieve similar performance.
The conjunctive variant with a depthwise-separable projection achieves
the highest mIoU, with 2.49\,M parameters compared with 151\,K for the
other variants. \tableref{tab:gate_ablation} in the main paper evaluates
the contribution of each gate within the chosen architecture.}

\revised{\tableref{tab:cons_weight_ablation} sweeps the two loss weights. The consistency
weight has a clear optimum. Moving from $\lambda_{\text{cons}}{=}0$ to $5$ gains $0.0051$
mIoU on the lighting-variant stream against $0.0022$ on clean, so the term acts where it is
designed to act. Pushing it further reverses this, and at $\lambda_{\text{cons}}{=}2000$ the
model loses $0.056$ mIoU as stream agreement starts to override segmentation accuracy. The supervised blend $\lambda_{\text{sup}}$ varies by less than $0.004$ across its range, which is expected once the consistency term already couples the streams. We adopt
$\lambda_{\text{cons}}{=}5$ and $\lambda_{\text{sup}}{=}0.5$.}

\begin{table}[htbp]
\centering
\caption{\revised{Gate composition. VOC, \lca{}-only, $N{=}4$, single run. Conjunctive fusion keeps a feature only where all three gates agree.}}
\label{tab:lsa_version_ablation}
\small
\setlength{\tabcolsep}{5pt}
\begin{tabular}{@{}l|c|cc@{}}
\toprule
\revised{Design} & \revised{Params} & \revised{Clean} & \revised{Variant} \\
\midrule
\revised{Sequential gating}  & \revised{151\,K}  & \revised{0.6698} & \revised{0.6626} \\
\revised{Additive fusion}    & \revised{151\,K}  & \revised{0.6686} & \revised{0.6608} \\
\revised{\textbf{Conjunctive fusion}} & \revised{2.49\,M} & \revised{\textbf{0.7027}} & \revised{\textbf{0.6946}} \\
\bottomrule
\end{tabular}
\end{table}

\begin{table}[htbp]
\centering
\caption{\revised{Loss weights, VOC, \lca{}-only, single run per cell. Adopted defaults in \textbf{bold}.}}
\label{tab:cons_weight_ablation}
\small
\setlength{\tabcolsep}{4pt}
\begin{minipage}[t]{0.48\linewidth}
  \centering
  \revised{\textbf{(a)}~$\lambda_{\text{cons}}$ (consistency)}\\[3pt]
  \begin{tabular}{@{}c|cc@{}}
  \toprule
  \revised{$\lambda_{\text{cons}}$} & \revised{Clean} & \revised{Variant} \\
  \midrule
  \revised{0}    & \revised{0.7030} & \revised{0.6941} \\
  \revised{0.1}  & \revised{0.7027} & \revised{0.6946} \\
  \revised{1}    & \revised{0.7013} & \revised{0.6949} \\
  \revised{\textbf{5}} & \revised{\textbf{0.7052}} & \revised{\textbf{0.6992}} \\
  \revised{10}   & \revised{0.7035} & \revised{0.6954} \\
  \revised{20}   & \revised{0.7036} & \revised{0.6972} \\
  \revised{100}  & \revised{0.6709} & \revised{0.6661} \\
  \revised{2000} & \revised{0.6477} & \revised{0.6429} \\
  \bottomrule
  \end{tabular}
\end{minipage}
\hfill
\begin{minipage}[t]{0.48\linewidth}
  \centering
  \revised{\textbf{(b)}~$\lambda_{\text{sup}}$ (supervised blend)}\\[3pt]
  \begin{tabular}{@{}c|cc@{}}
  \toprule
  \revised{$\lambda_{\text{sup}}$} & \revised{Clean} & \revised{Variant} \\
  \midrule
  \revised{0}   & \revised{0.6999} & \revised{0.6925} \\
  \revised{0.1} & \revised{0.7002} & \revised{0.6922} \\
  \revised{0.3} & \revised{0.6995} & \revised{0.6924} \\
  \revised{\textbf{0.5}} & \revised{\textbf{0.7004}} & \revised{\textbf{0.6922}} \\
  \revised{0.7} & \revised{0.7014} & \revised{0.6941} \\
  \revised{0.9} & \revised{0.7021} & \revised{0.6949} \\
  \revised{1}   & \revised{0.7026} & \revised{0.6954} \\
  \bottomrule
  \end{tabular}
\end{minipage}
\end{table}

\begin{figure}[htbp]
    \centering
        \includegraphics[width=\linewidth]{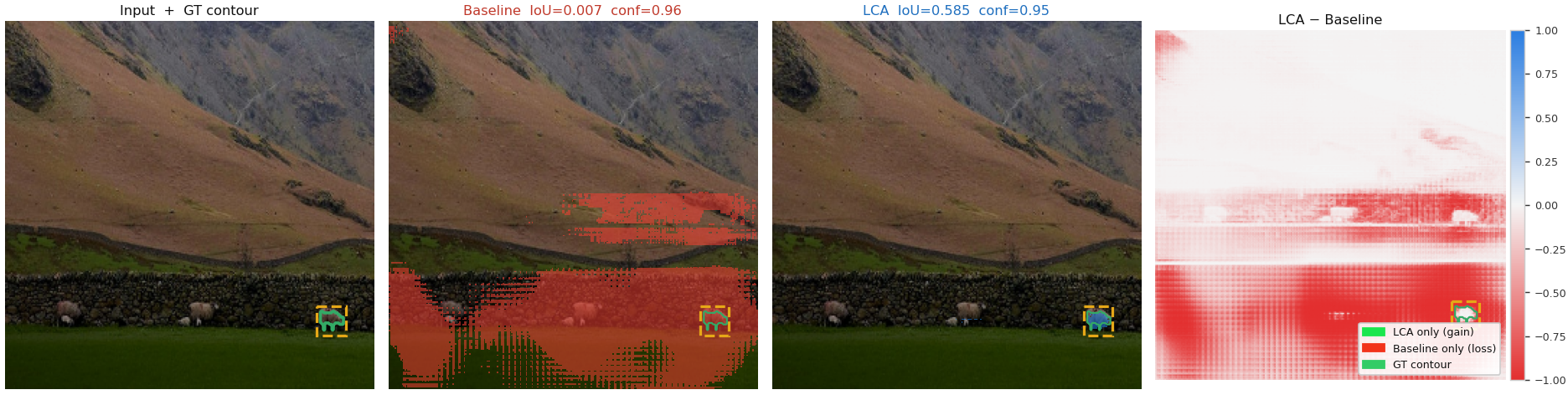}
        \caption{\revised{Sheep in a low-contrast scene. The baseline floods the hillside with false positives (IoU 0.007). \lca{} isolates the target (IoU 0.585).}}


    \label{fig:qualitative}
\end{figure}

\subsubsection{\revised{Positioning against Low-Light Instance Segmentation}}
\label{sec:supp_positioning}

\revised{\emph{Instance Segmentation in the Dark}~\cite{chen2023instance} pursues a
disturbance-invariant objective that is conceptually parallel to our consistency loss, but it
treats high-frequency content in the opposite way. It \emph{suppresses} high-frequency
disturbance, because noise dominates that band in RAW low-light capture. We \emph{amplify}
it, because edges dominate that band at moderate illumination. The two positions are
consistent once the noise regime is made explicit, and which treatment is correct is an
empirical question about the sensor noise floor rather than a disagreement about
architecture. We did not evaluate on LIS, which requires a RAW processing pipeline we do not
implement, or on Dark Zurich~\cite{sakaridis2019guided}, which provides only semantic
rather than instance annotations.}

\subsubsection{Implementation Details}

We build on the SAM checkpoint and attach \lca{} modules to the last \revised{$N{=}4$ transformer blocks for SAM and the last $N{=}10$ for HQ-SAM and SAM~3}. The image encoder and prompt encoder are frozen. We train with AdamW using three parameter groups: \lca{} weights at learning rate $1 \times 10^{-4}$ with weight decay $1 \times 10^{-2}$, gate scalars at $2 \times 10^{-3}$ with no weight decay, and the mask decoder at $5 \times 10^{-5}$ with weight decay $5 \times 10^{-2}$. \revised{Training runs for 20 epochs. Batch size is set to what fits one card for each backbone family, which is 32 for SAM, 16 for HQ-SAM and 2 for SAM~3.} We use automatic mixed precision throughout and clip gradient norms to 1.0. \revised{The supervised and consistency loss weights are $\lambda_s = 0.5$ and $\lambda_c = 5$ unless otherwise noted.}

\subsubsection{Computational Overhead}



\tableref{tab:computation} summarizes the parameter counts for SAM with $N{=}4$ patched blocks.
\lca{} adds 2.49\,M trainable parameters, increasing the total model size
from 89.67\,M to 92.17\,M, a 2.8\% overhead.
Combined with the mask decoder, the trainable parameters
total 6.55\,M, about 7.1\% of the full model. The deployment cost of each
configuration, including the deeper $N{=}10$ setting used for HQ-SAM and SAM~3,
is reported in \tableref{tab:compute_main}.

\begin{table}[htbp]
\centering
\caption{%
    Computational comparison for SAM with $N{=}4$. Total parameters are reported for the full model, including the frozen SAM encoder and the mask decoder. Trainable parameters are those that receive gradients.%
}
\label{tab:computation}
\small
\begin{tabular}{@{}lcc@{}}
\toprule
\textbf{Model} & \textbf{Params (M)} & \textbf{Trainable (M)} \\
\midrule
SAM                    & 89.67 & n/a  \\
+ Decoder training     & 89.67 & 4.06 \\
+ \textbf{\lca{}} only & 92.17 & 2.49 \\
+ \textbf{\lca{}+Dec}  & 92.17 & 6.55 \\
\bottomrule
\end{tabular}
\end{table}

\subsubsection{Robustness Analysis}
\label{sec:supp_robustness}

\begin{figure*}[htbp]
    \centering
    \includegraphics[width=0.70\textwidth]{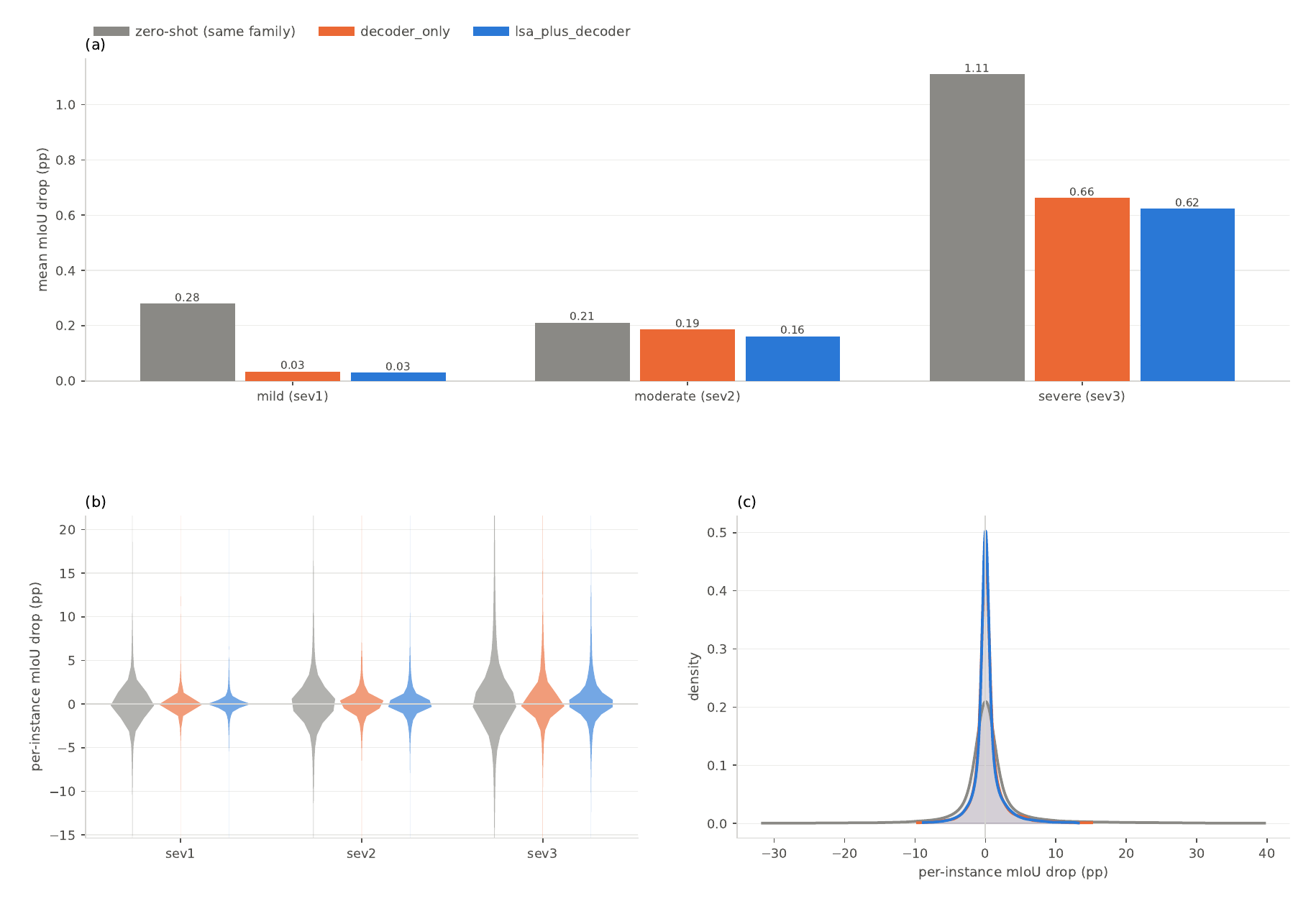}
    \caption{\revised{Robustness analysis on COCO, SAM family, across the three
        severity levels. \textbf{(a)} Mean absolute mIoU drop (clean $-$ variant) per
        severity. At severe the frozen backbone loses 1.11\,pp, decoder tuning 0.66\,pp and
        \lca{}+Dec 0.62\,pp. \textbf{(b)} Per-instance drop distributions, where a wider spread
        means more variable degradation. \textbf{(c)} Pooled density over all severities, where a
        sharper peak at zero indicates greater lighting robustness.}}
    \label{fig:robustness_coco}
\end{figure*}

\begin{figure*}[htbp]
    \centering
    \includegraphics[width=0.70\textwidth]{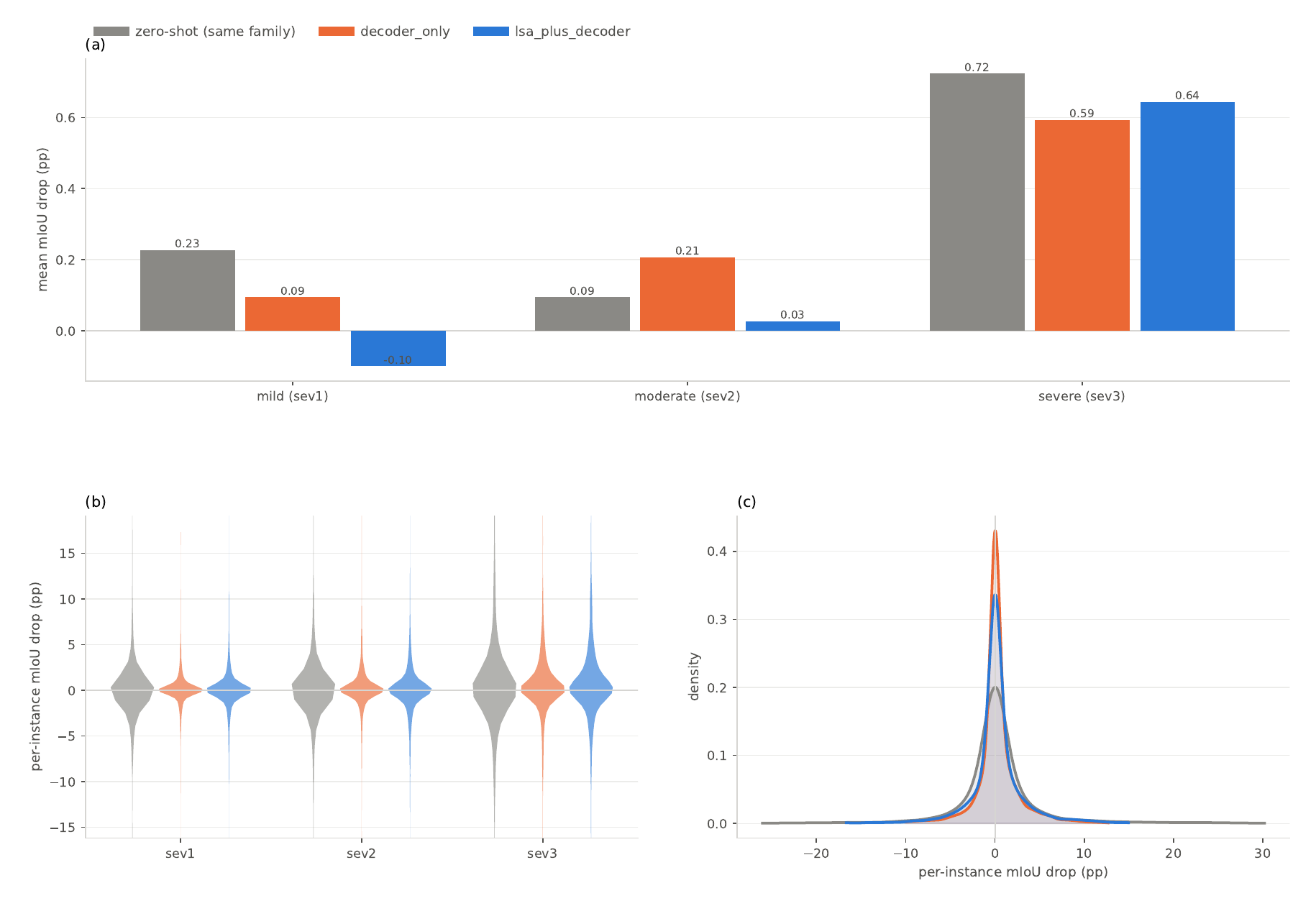}
    \caption{\revised{Robustness analysis on Cityscapes, SAM family, across the three
        severity levels. \textbf{(a)} Mean absolute mIoU drop (clean $-$ variant) per
        severity. At severe the frozen backbone loses 1.02\,pp, decoder tuning 0.73\,pp and
        \lca{}+Dec 0.58\,pp. \textbf{(b)} Per-instance drop distributions, where a wider spread
        means more variable degradation. \textbf{(c)} Pooled density over all severities, where a
        sharper peak at zero indicates greater lighting robustness.}}
    \label{fig:robustness_cityscapes}
\end{figure*}

\begin{figure*}[htbp]
    \centering
    \includegraphics[width=0.70\textwidth]{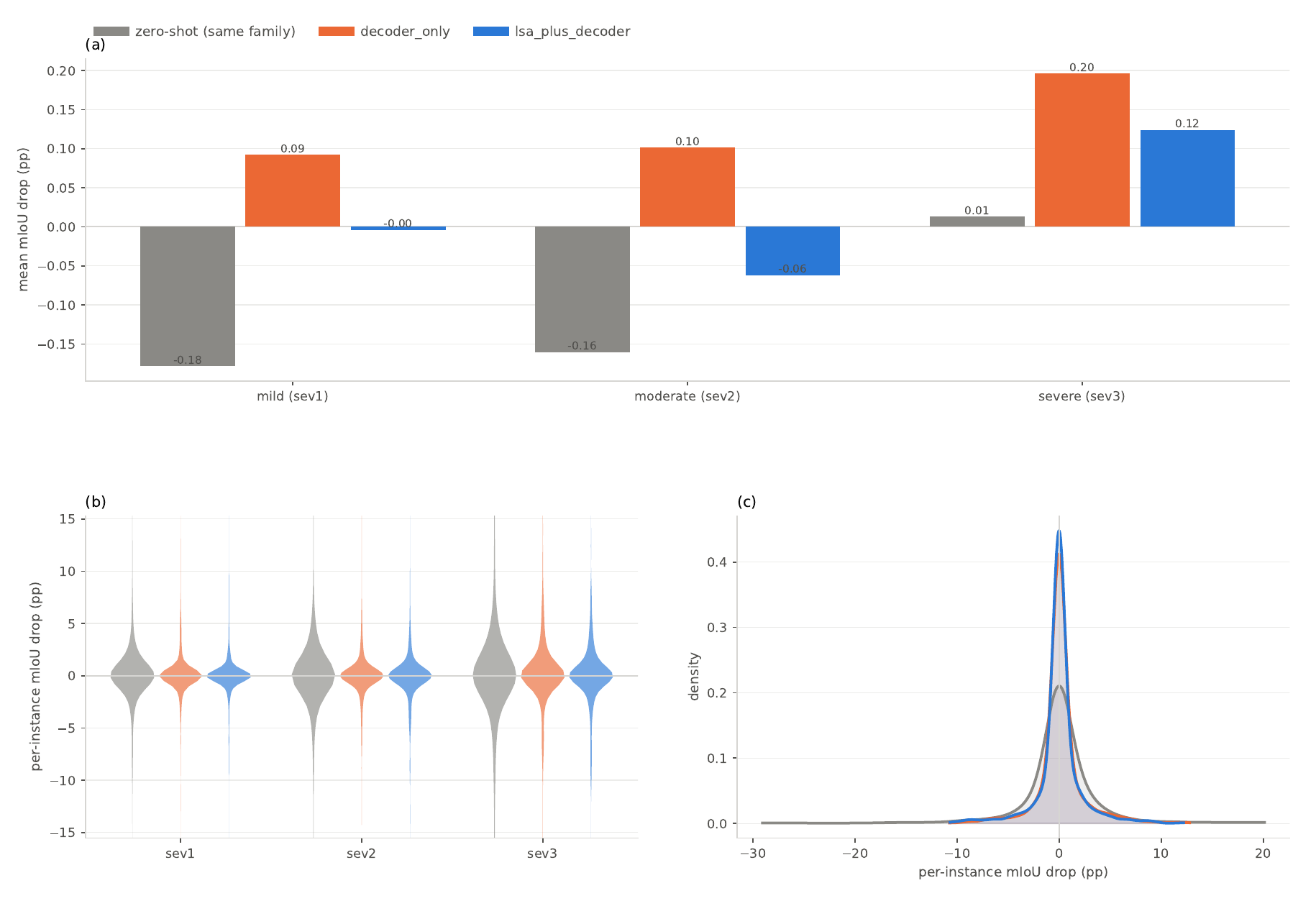}
    \caption{\revised{Robustness analysis on real ACDC snow, SAM family, across the three
        severity levels. \textbf{(a)} Mean absolute mIoU drop (clean $-$ variant) per
        severity. At severe the frozen backbone loses 0.47\,pp, decoder tuning 0.27\,pp and
        \lca{}+Dec 0.11\,pp. \textbf{(b)} Per-instance drop distributions, where a wider spread
        means more variable degradation. \textbf{(c)} Pooled density over all severities, where a
        sharper peak at zero indicates greater lighting robustness.}}
    \label{fig:robustness_snow}
\end{figure*}

\revised{\figref{fig:robustness_coco} to \figref{fig:robustness_snow} show how much each
model degrades under increasing lighting severity, on the three settings where \lca{}+Dec holds
the smallest absolute drop. Each figure reports \textbf{absolute} mIoU drop in percentage
points, which is not the same quantity as the relative drop in
\tableref{tab:robustness}(a). A model that scores higher on clean has more to lose in absolute
terms, so the two views can disagree, and the relative figure is the fairer summary across all
datasets. We show the absolute view here because it is what the per-instance distributions in
panels~(b) and~(c) are built from.}

\revised{The pattern is consistent across the three. At mild severity all models behave alike,
as expected when the perturbation is small. Differences emerge at moderate severity and are
clearest at severe: on COCO the frozen backbone loses $1.11$\,pp while \lca{}+Dec limits the
loss to $0.62$\,pp, and on Cityscapes $1.02$\,pp against $0.58$\,pp. Panel~(b) shows this is
not driven by a few outliers: the frozen model develops a long tail of failures as severity
rises, while the adapted models stay concentrated. Panel~(c) pools all severities, where
\lca{}+Dec is the most tightly peaked at zero, indicating fewer large failures overall.}

\subsubsection{Per-Dataset IoU Analysis}
\label{sec:supp_iou_analysis}

\begin{figure*}[tp]
    \centering
    \begin{subfigure}[t]{0.614\linewidth}
        \includegraphics[width=\linewidth]{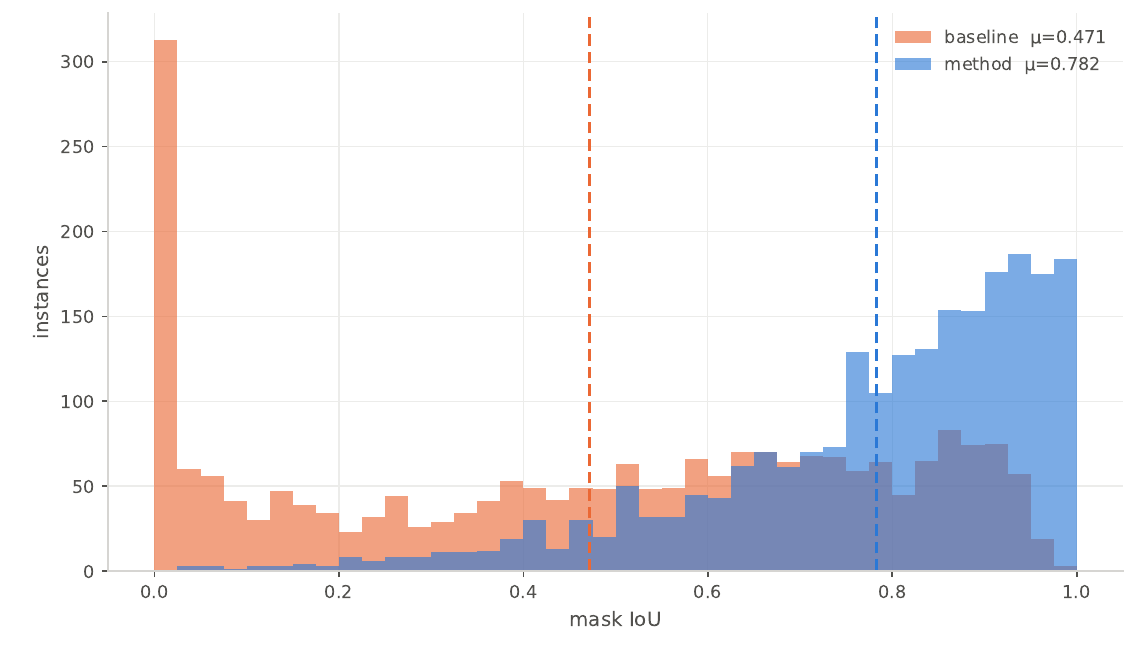}
        \caption{\revised{IoU distribution. \lca{} shifts the mass rightward
        ($\mu{=}0.782$) against the frozen backbone ($\mu{=}0.471$).}}
        \label{fig:iou_histogram}
    \end{subfigure}
    \hfill
    \begin{subfigure}[t]{0.356\linewidth}
        \includegraphics[width=\linewidth]{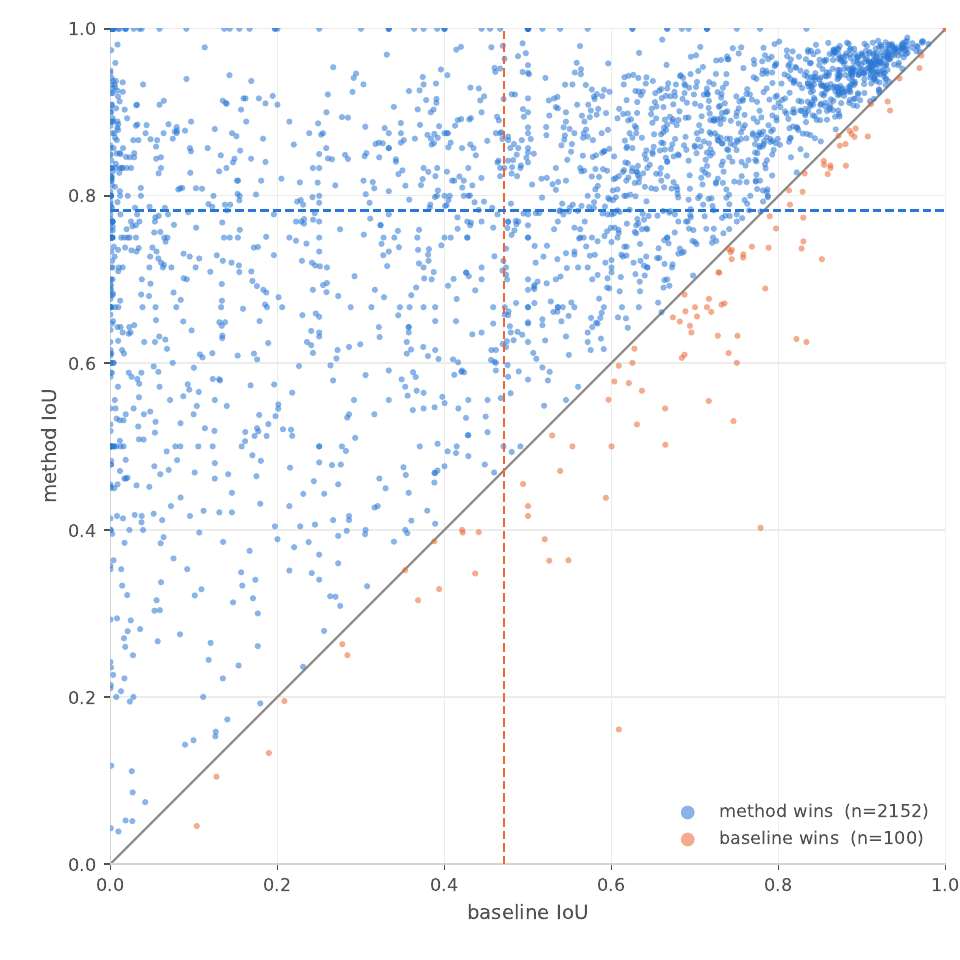}
        \caption{\revised{Per-instance scatter. \lca{} wins on 2{,}152 of 2{,}252
        instances (95.6\%).}}
        \label{fig:iou_scatter}
    \end{subfigure}
    \caption{\revised{Per-instance IoU on the Cityscapes lighting-variant stream, HQ-SAM frozen against HQ-SAM\,+\,\lca{}+Dec over the full validation set. Points above $y{=}x$ are instances \lca{} improves.}}
    \label{fig:iou_analysis}
\end{figure*}

\begin{figure*}[tp]
    \centering
    \begin{subfigure}[t]{0.614\linewidth}
        \includegraphics[width=\linewidth]{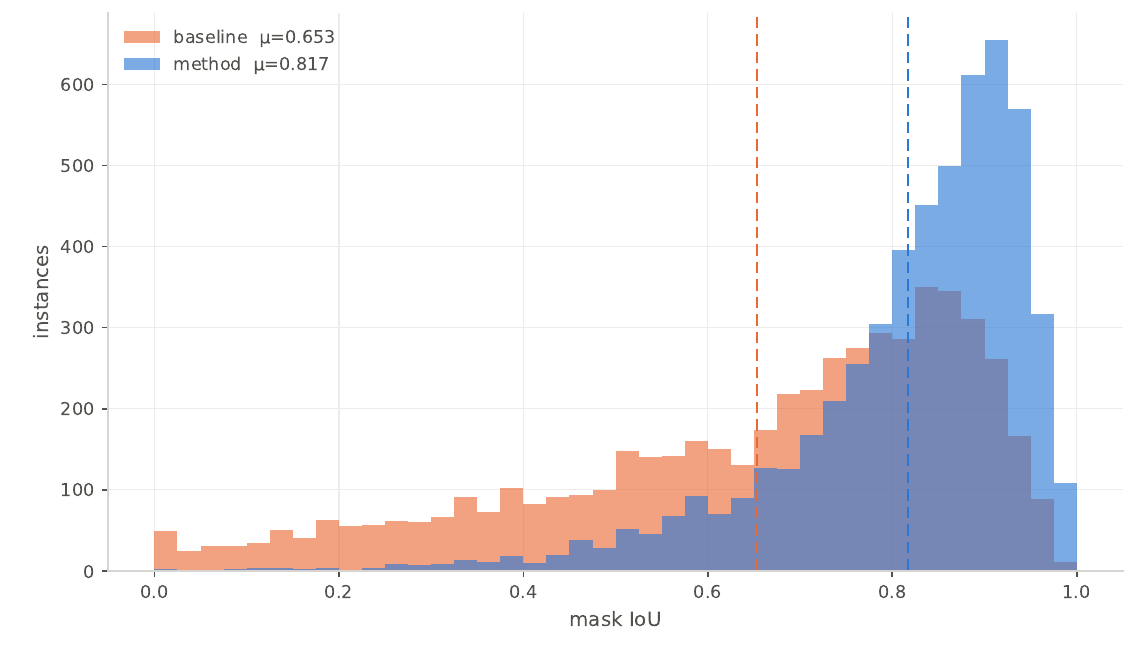}
        \caption{\revised{IoU histogram, COCO. $\mu$ 0.653 $\rightarrow$ 0.817.}}
        \label{fig:supp_iou_coco_hist}
    \end{subfigure}
    \hfill
    \begin{subfigure}[t]{0.356\linewidth}
        \includegraphics[width=\linewidth]{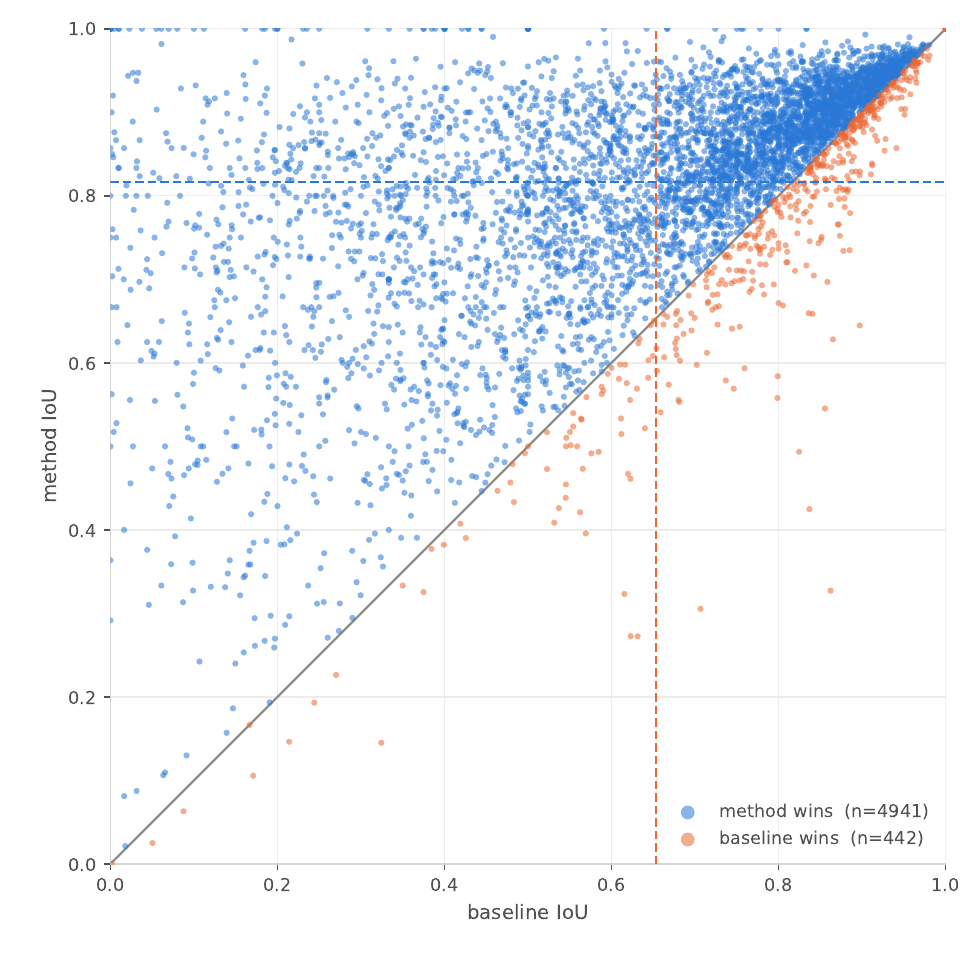}
        \caption{\revised{Per-instance scatter, COCO. \lca{} wins 4,941 of 5,383 (91.8\%).}}
        \label{fig:supp_iou_coco_scatter}
    \end{subfigure}
    \caption{%
        IoU analysis on COCO under lighting variant conditions.
        Left: distribution shift between SAM baseline and \lca{}.
        Right, per-instance comparison, where points above $y{=}x$ indicate \lca{} improvement.%
    }
    \label{fig:supp_iou_coco}
\end{figure*}

\begin{figure*}[tp]
    \centering
    \begin{subfigure}[t]{0.614\linewidth}
        \includegraphics[width=\linewidth]{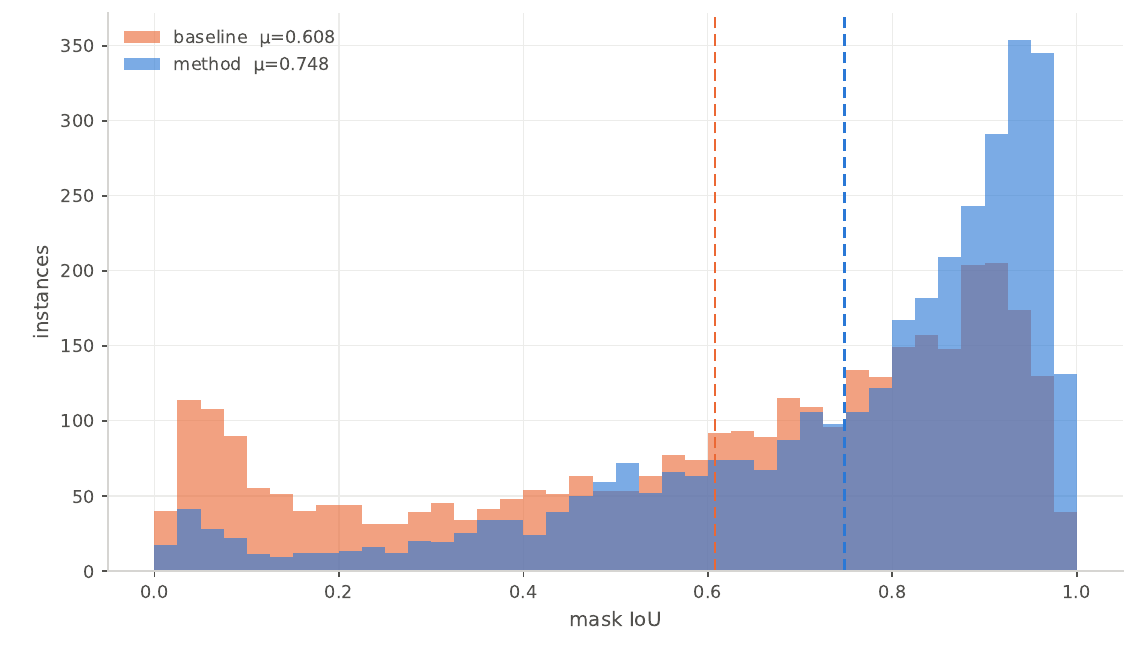}
        \caption{\revised{IoU histogram, VOC. $\mu$ 0.608 $\rightarrow$ 0.748.}}
        \label{fig:supp_iou_voc_hist}
    \end{subfigure}
    \hfill
    \begin{subfigure}[t]{0.356\linewidth}
        \includegraphics[width=\linewidth]{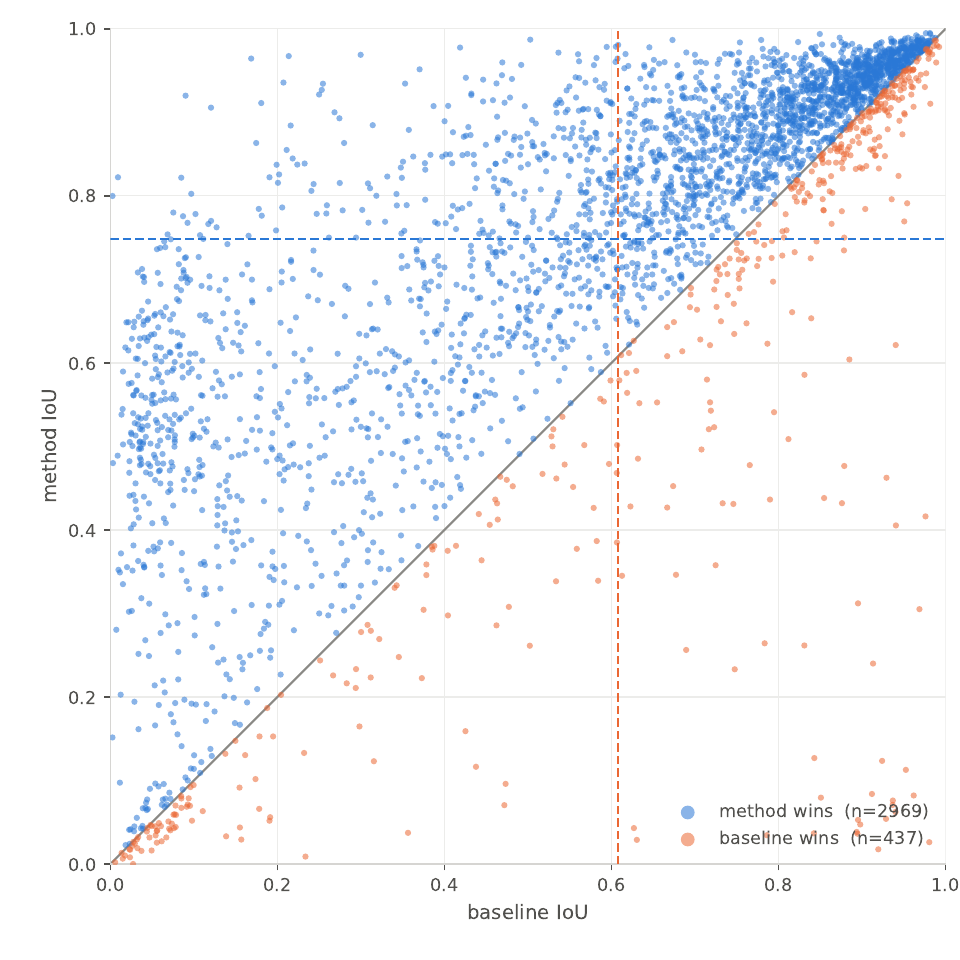}
        \caption{\revised{Per-instance scatter, VOC. \lca{} wins 2,969 of 3,406 (87.2\%).}}
        \label{fig:supp_iou_voc_scatter}
    \end{subfigure}
    \caption{%
        IoU analysis on VOC under lighting variant conditions.%
    }
    \label{fig:supp_iou_voc}
\end{figure*}

\begin{figure*}[tp]
    \centering
    \begin{subfigure}[t]{0.614\linewidth}
        \includegraphics[width=\linewidth]{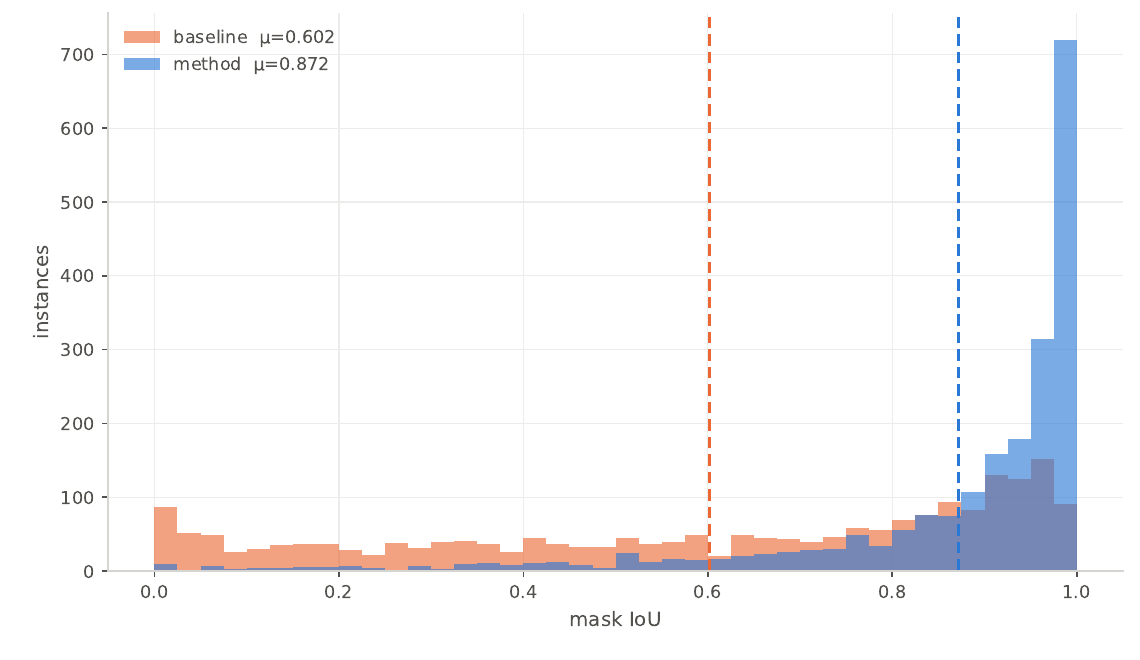}
        \caption{\revised{IoU histogram, Unity. $\mu$ 0.602 $\rightarrow$ 0.872.}}
        \label{fig:supp_iou_unity_hist}
    \end{subfigure}
    \hfill
    \begin{subfigure}[t]{0.356\linewidth}
        \includegraphics[width=\linewidth]{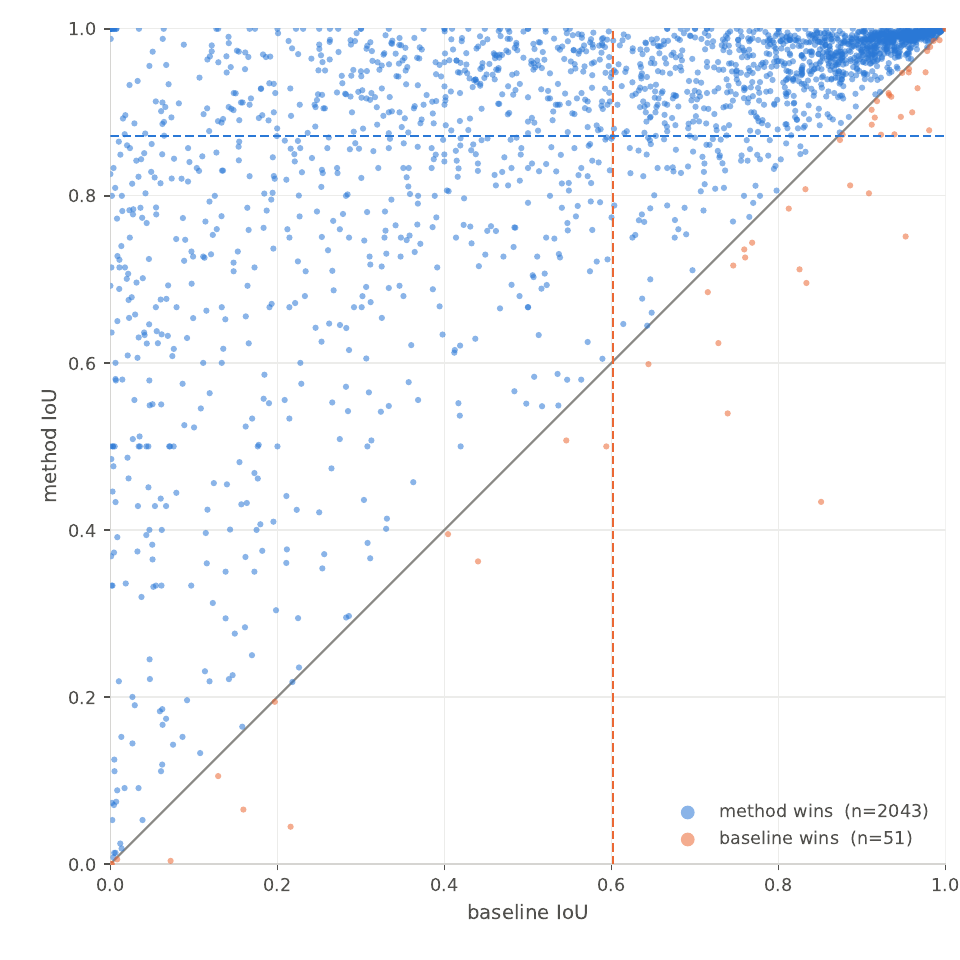}
        \caption{\revised{Per-instance scatter, Unity. \lca{} wins 2,043 of 2,094 (97.6\%).}}
        \label{fig:supp_iou_unity_scatter}
    \end{subfigure}
    \caption{%
        IoU analysis on the Unity synthetic dataset under physically rendered
        lighting variant conditions.
        Because Unity lighting is rendered rather than augmented, the
        performance gap is larger, reflecting the additional domain shift.%
    }
    \label{fig:supp_iou_unity}
\end{figure*}

\figref{fig:iou_analysis} shows the IoU histogram and scatter plot for Cityscapes under lighting variant conditions.
Here we provide the same analysis for the three remaining benchmarks
(\figref{fig:supp_iou_coco}-\figref{fig:supp_iou_unity}). In all cases, the histogram shift to the right and the majority of scatter
points above the $y{=}x$ diagonal confirm that \lca{} consistently
improves per-instance IoU, with the largest gains concentrated in
low-baseline instances where lighting degradation most severely disrupts
the encoder.

\begin{figure*}[tp]
    \centering
    \includegraphics[width=\textwidth]{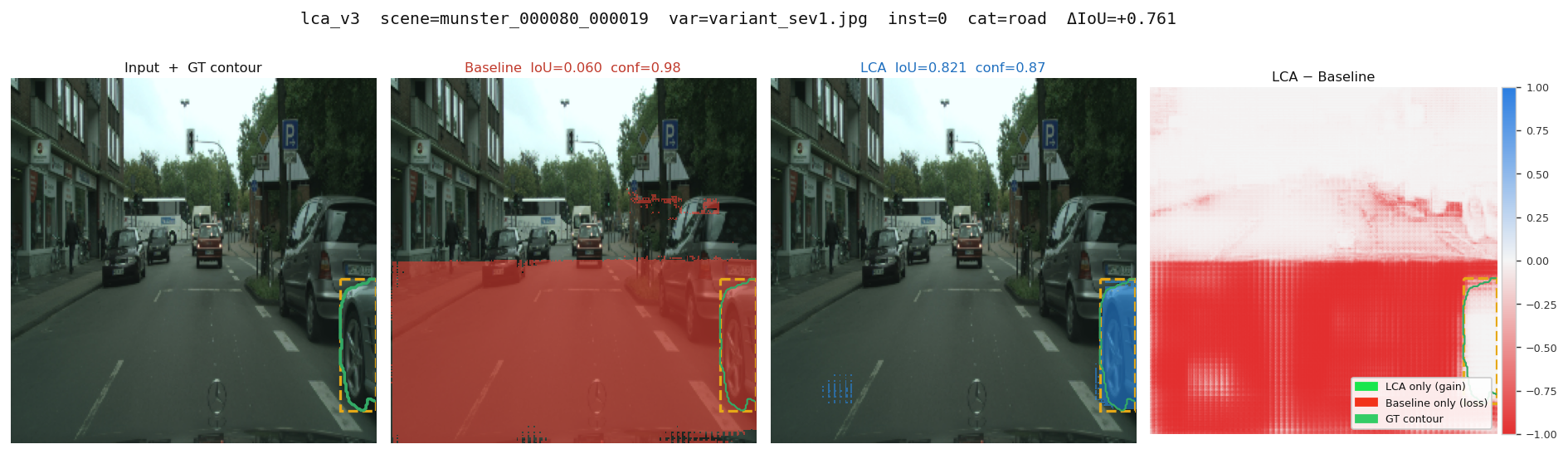}
    \caption{%
        \textbf{Cityscapes} (road, sev1, $\Delta$IoU\,=\,+0.761).
        The baseline floods the entire road surface with high confidence
        (IoU\,=\,0.060, conf\,=\,0.98). \lca{} recovers a compact mask
        tightly aligned with the target lane segment (IoU\,=\,0.821).%
    }
    \label{fig:hg_city1}
\end{figure*}

\begin{figure*}[tp]
    \centering
    \includegraphics[width=\textwidth]{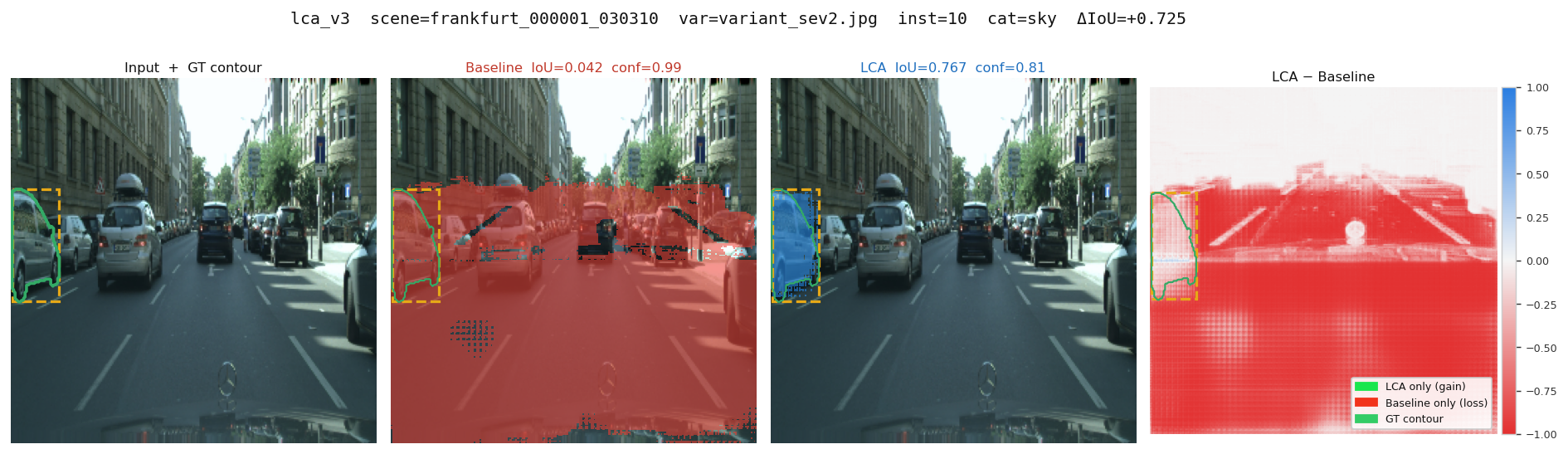}
    \caption{%
        \textbf{Cityscapes} (sky, sev2, $\Delta$IoU\,=\,+0.725).
        Under moderate lighting, the baseline misidentifies large building
        façades as sky (IoU\,=\,0.042). \lca{} correctly isolates the
        visible sky region (IoU\,=\,0.767).%
    }
    \label{fig:hg_city2}
\end{figure*}

\begin{figure*}[tp]
    \centering
    \includegraphics[width=\textwidth]{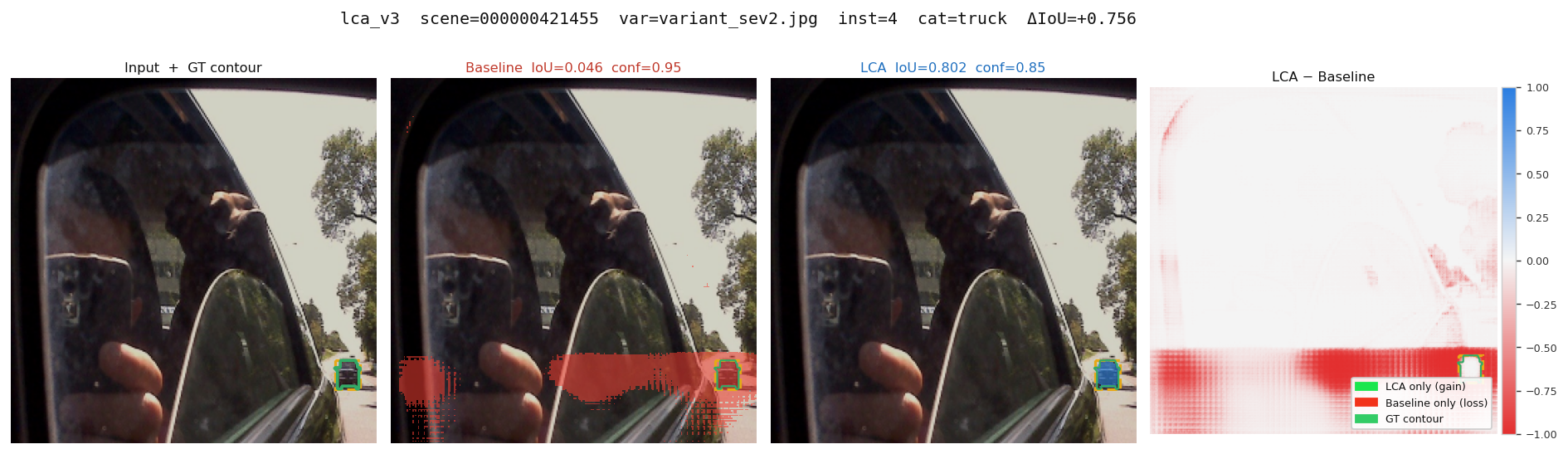}
    \caption{%
        \textbf{COCO} (truck, sev2, $\Delta$IoU\,=\,+0.756).
        The baseline activates on the lower-half background rather than the
        vehicle body (IoU\,=\,0.046). \lca{} produces a well-bounded truck
        mask (IoU\,=\,0.802), demonstrating robustness to the
        high-contrast reflective window surface.%
    }
    \label{fig:hg_coco1}
\end{figure*}

\begin{figure*}[tp]
    \centering
    \includegraphics[width=\textwidth]{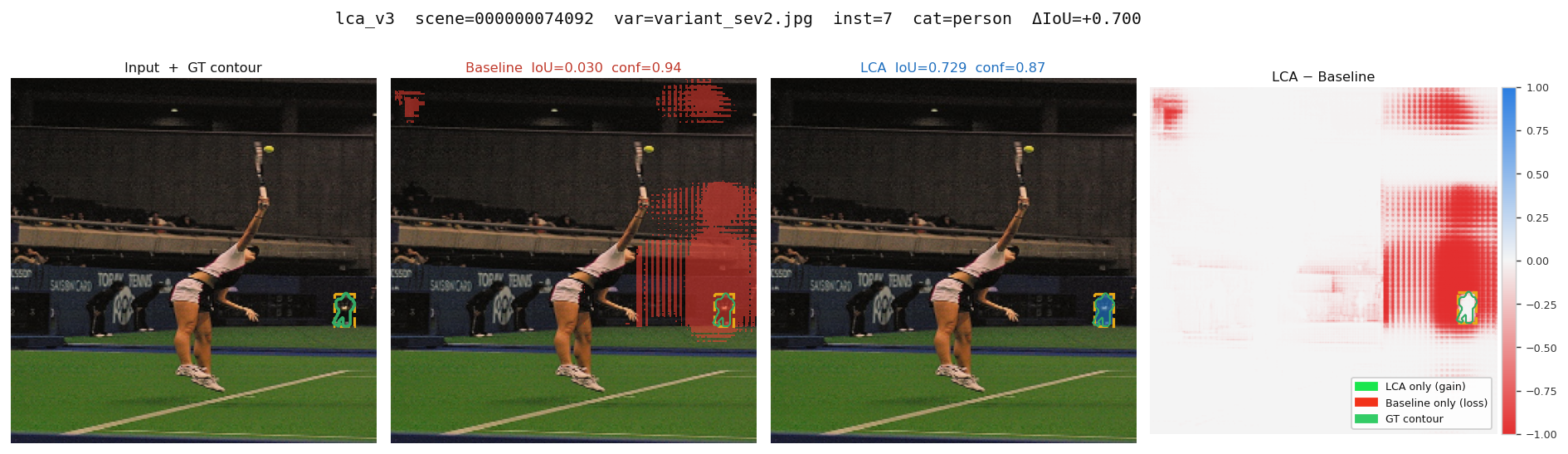}
    \caption{%
        \textbf{COCO} (person, sev2, $\Delta$IoU\,=\,+0.700).
        Low ambient stadium lighting causes the baseline to flood a
        rectangular region of the court (IoU\,=\,0.030). \lca{} correctly
        segments the athlete's silhouette (IoU\,=\,0.729).%
    }
    \label{fig:hg_coco2}
\end{figure*}

\begin{figure*}[tp]
    \centering
    \includegraphics[width=\textwidth]{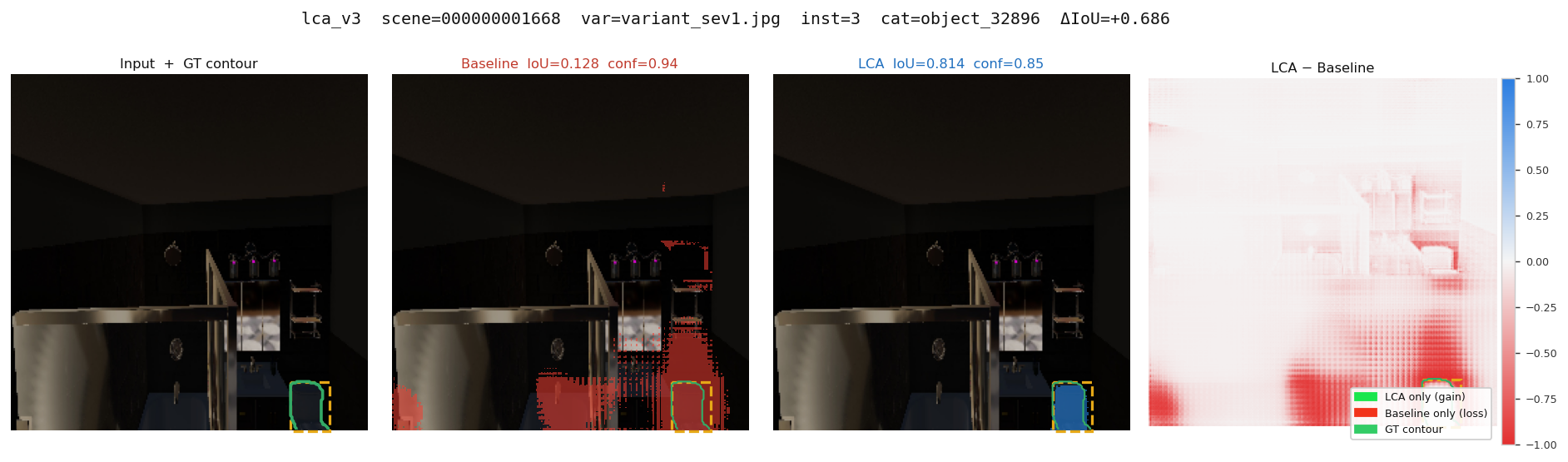}
    \caption{%
        \textbf{Unity} (object, sev1, $\Delta$IoU\,=\,+0.686).
        In a near-dark synthetic indoor scene, the baseline produces
        a large false-positive region unrelated to the target object
        (IoU\,=\,0.128). \lca{} isolates the dimly-lit target using
        structural edge cues (IoU\,=\,0.814).%
    }
    \label{fig:hg_unity1}
\end{figure*}

\begin{figure*}[tp]
    \centering
    \includegraphics[width=\textwidth]{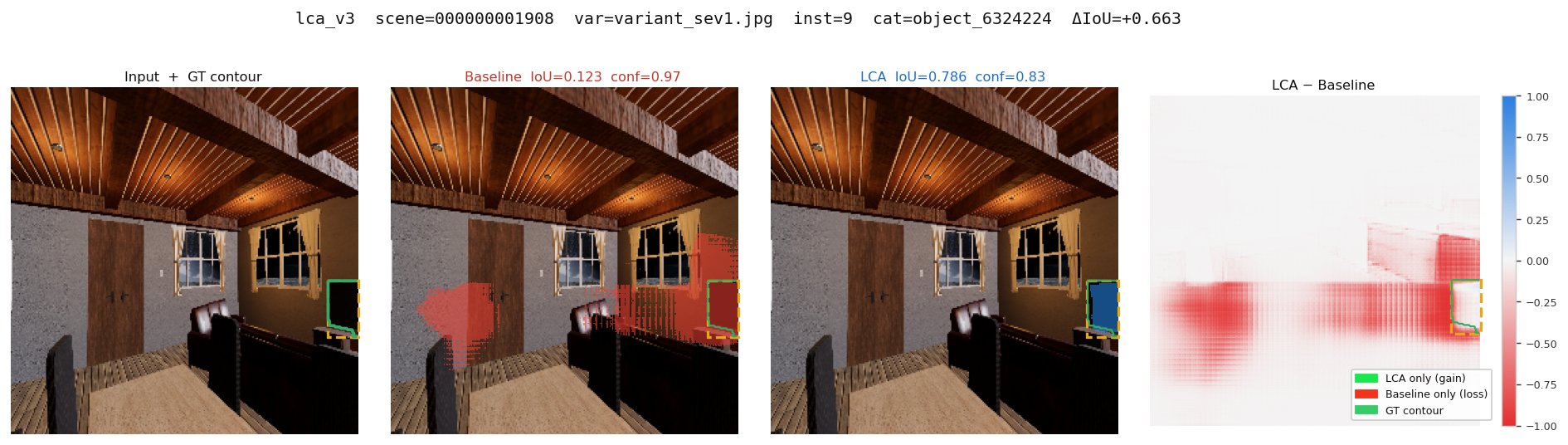}
    \caption{%
        \textbf{Unity} (object, sev1, $\Delta$IoU\,=\,+0.663).
        Warm ceiling lighting creates a strong false edge that misleads
        the baseline into activating across the wall and furniture
        (IoU\,=\,0.123). \lca{} suppresses the photometric distraction
        and recovers the target boundary (IoU\,=\,0.786).%
    }
    \label{fig:hg_unity2}
\end{figure*}

\subsection{Examples of Data}
\label{sec:supp_qualitative}

We provide examples across the all evaluation benchmark. For each of the scenes, the top rows show the clean image and are followed by its lighting variants across all three different severity levels (mild, moderate, severe). The severity level increases from the left to the right. The bottom row shows the corresponding ground-truth instance masks.

\subsubsection{\revised{Robustness on the Remaining Datasets}}
\label{sec:supp_robustness_rest}

\revised{For completeness we report the same analysis on the datasets not shown above. On
these, \lca{}+Dec does \emph{not} hold the smallest absolute drop, and we include them rather
than select only the favourable cases. The reason is visible in the numbers. absolute drop
scales with how much a model has to lose. On VOC the frozen backbone starts at $0.613$ mIoU and
loses $1.10$\,pp, while \lca{}+Dec starts at $0.758$ and loses $1.28$\,pp. Yet in relative
terms \lca{}+Dec degrades least ($1.69\%$ against $1.80\%$, \tableref{tab:robustness}(a)).
The same holds on Unity. On the ACDC conditions the frozen backbone barely degrades at all,
because synthetic re-lighting applied on top of an already adverse photograph changes little.
Those rows say more about the mildness of the perturbation than about robustness. The
clean column of \tableref{tab:severity_full} is the meaningful number there.}

\begin{figure*}[htbp]
    \centering
    \includegraphics[width=0.70\textwidth]{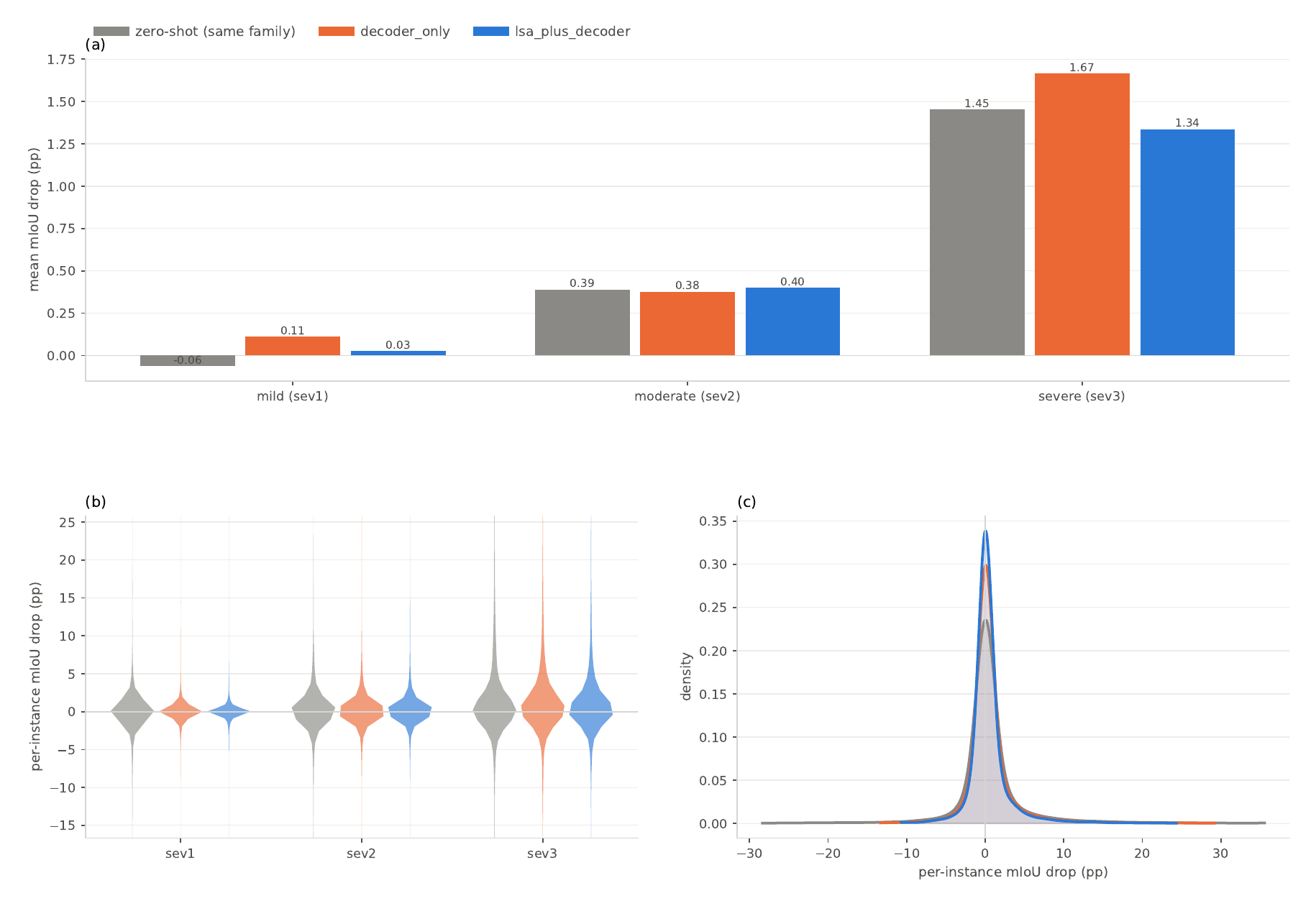}
    \caption{\revised{Robustness analysis on VOC, SAM family. Mean absolute mIoU
        drop at severe: frozen 1.10\,pp, decoder tuning 1.71\,pp, \lca{}+Dec 1.28\,pp.
        Panels as in \figref{fig:robustness_coco}.}}
    \label{fig:robustness_voc}
\end{figure*}

\begin{figure*}[htbp]
    \centering
    \includegraphics[width=0.70\textwidth]{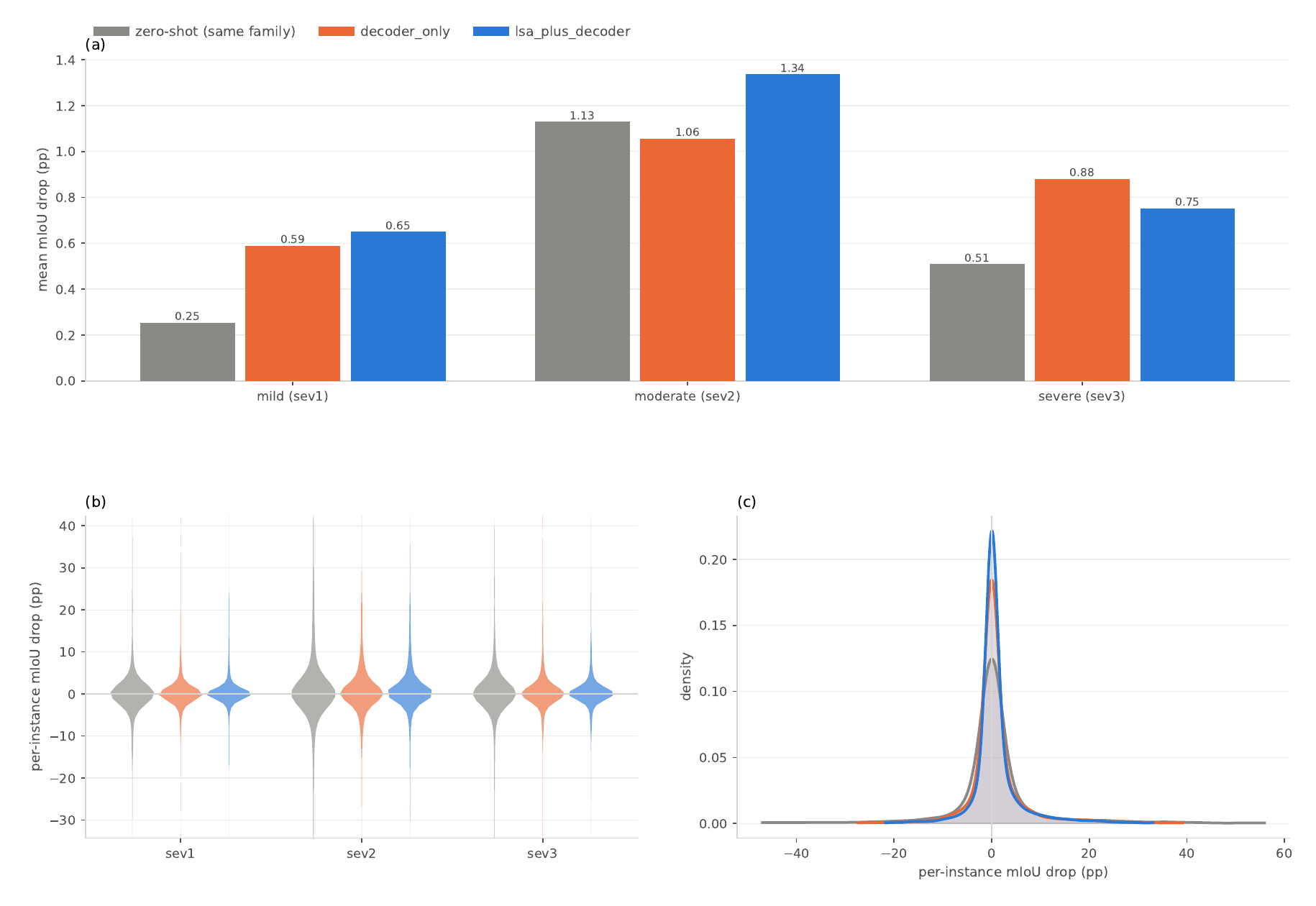}
    \caption{\revised{Robustness analysis on Unity, SAM family. Mean absolute mIoU
        drop at severe: frozen 0.62\,pp, decoder tuning 0.90\,pp, \lca{}+Dec 0.81\,pp.
        Panels as in \figref{fig:robustness_coco}.}}
    \label{fig:robustness_unity}
\end{figure*}

\begin{figure*}[htbp]
    \centering
    \includegraphics[width=0.70\textwidth]{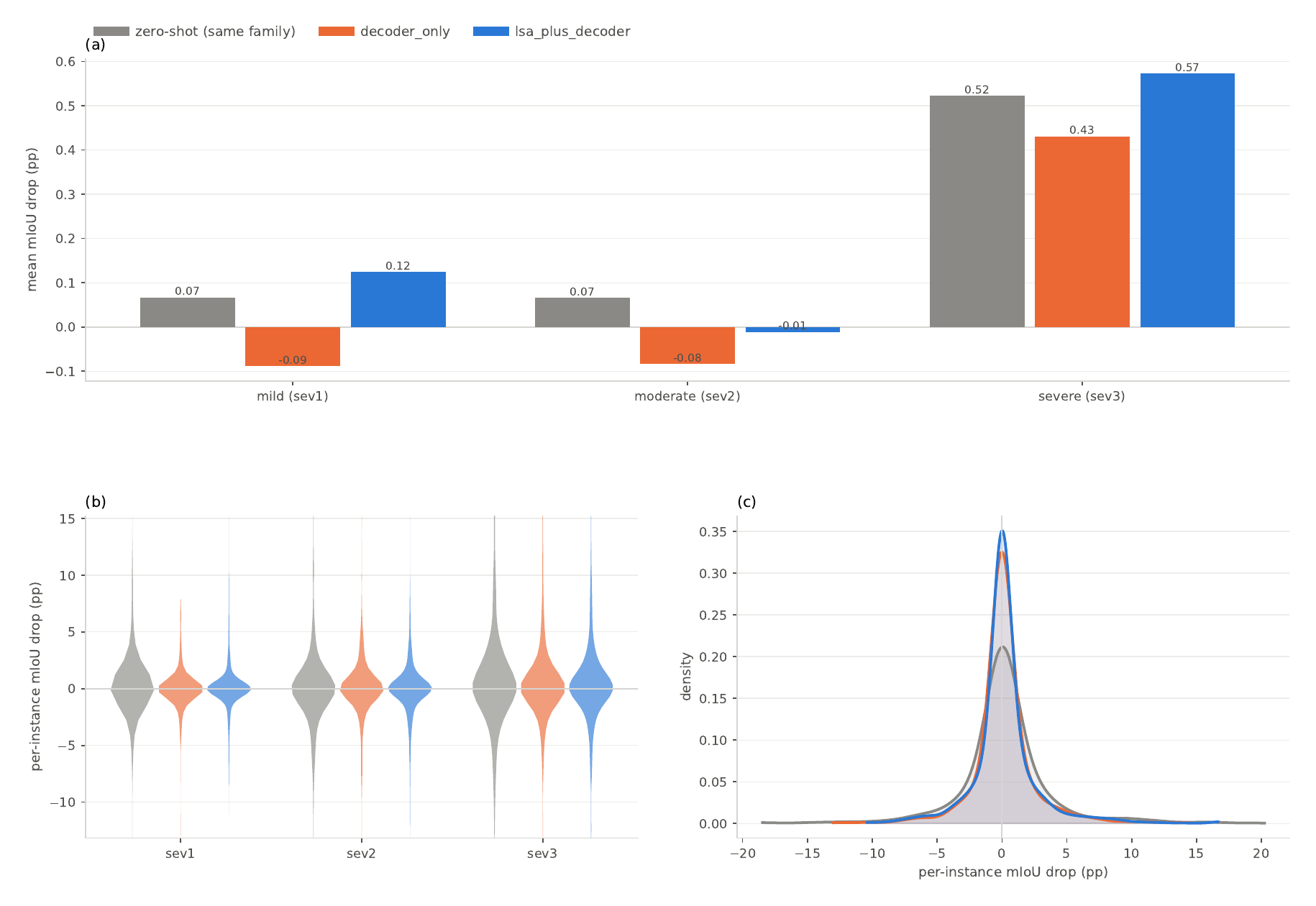}
    \caption{\revised{Robustness analysis on real ACDC fog, SAM family. Mean absolute mIoU
        drop at severe: frozen 0.72\,pp, decoder tuning 0.51\,pp, \lca{}+Dec 0.59\,pp.
        Panels as in \figref{fig:robustness_coco}.}}
    \label{fig:robustness_acdc_fog}
\end{figure*}

\begin{figure*}[htbp]
    \centering
    \includegraphics[width=0.70\textwidth]{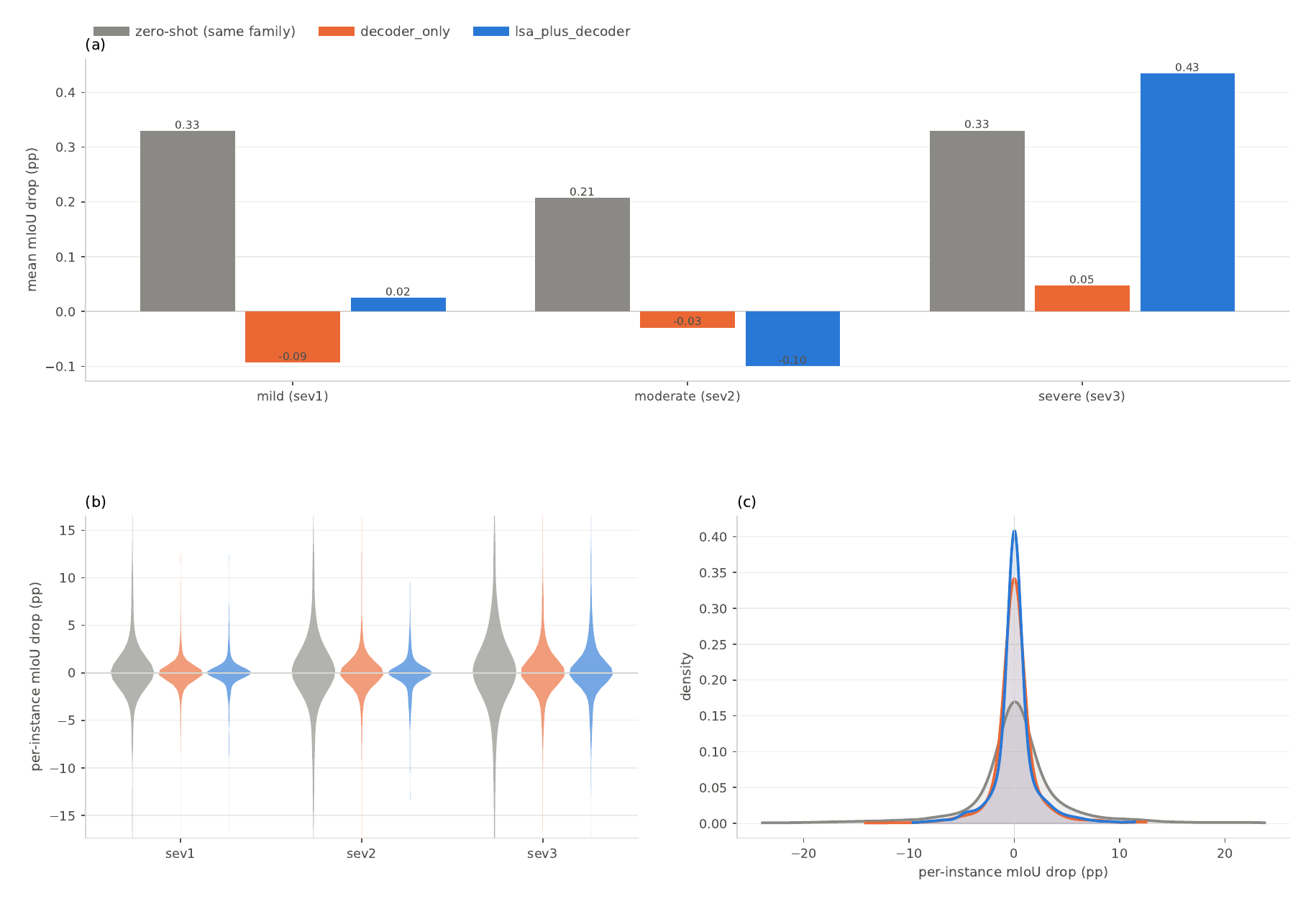}
    \caption{\revised{Robustness analysis on real ACDC night, SAM family. Mean absolute mIoU
        drop at severe: frozen 0.07\,pp, decoder tuning 0.36\,pp, \lca{}+Dec 0.29\,pp.
        Panels as in \figref{fig:robustness_coco}.}}
    \label{fig:robustness_acdc_night}
\end{figure*}

\begin{figure*}[htbp]
    \centering
    \includegraphics[width=0.70\textwidth]{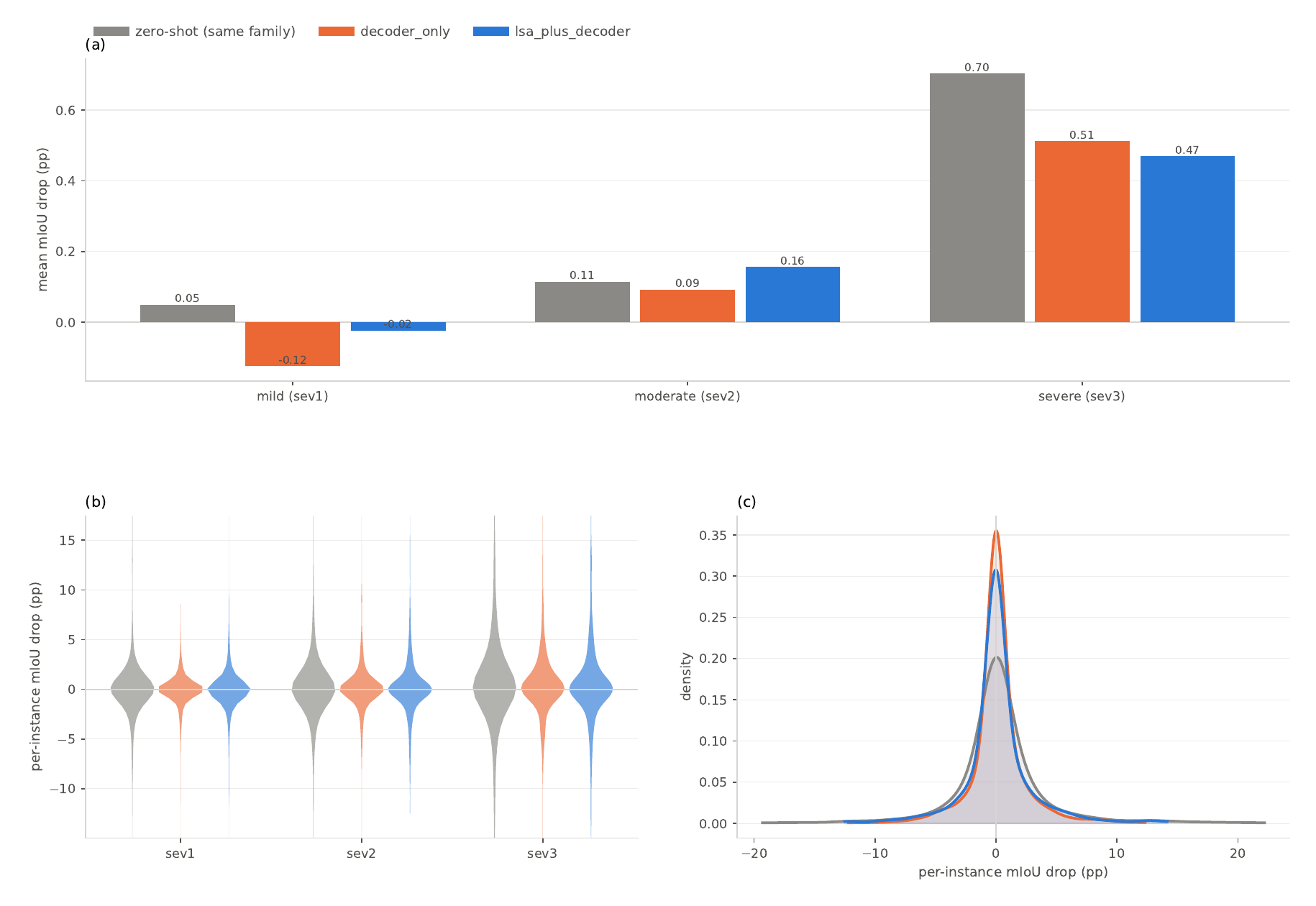}
    \caption{\revised{Robustness analysis on real ACDC rain, SAM family. Mean absolute mIoU
        drop at severe: frozen 0.59\,pp, decoder tuning 0.49\,pp, \lca{}+Dec 0.61\,pp.
        Panels as in \figref{fig:robustness_coco}.}}
    \label{fig:robustness_acdc_rain}
\end{figure*}

\subsubsection{\revised{Per-Instance IoU on Real Adverse Imagery}}
\label{sec:supp_iou_acdc}

\revised{The per-dataset analyses above use synthetically re-lit imagery. The figures below
repeat it on the four real ACDC conditions with SAM~3 as the frozen reference, so the
distribution shift can be read on genuine adverse captures rather than on \plap{} output.
The pattern is the same on real data: the mass of low-IoU instances left by the frozen
backbone is pulled towards $1.0$, and \lca{} wins the large majority of instances in every
condition. The night condition has both the lowest frozen mean ($0.661$) and the largest remaining tail.}

\begin{figure*}[tp]
    \centering
    \begin{subfigure}[t]{0.614\linewidth}
        \includegraphics[width=\linewidth]{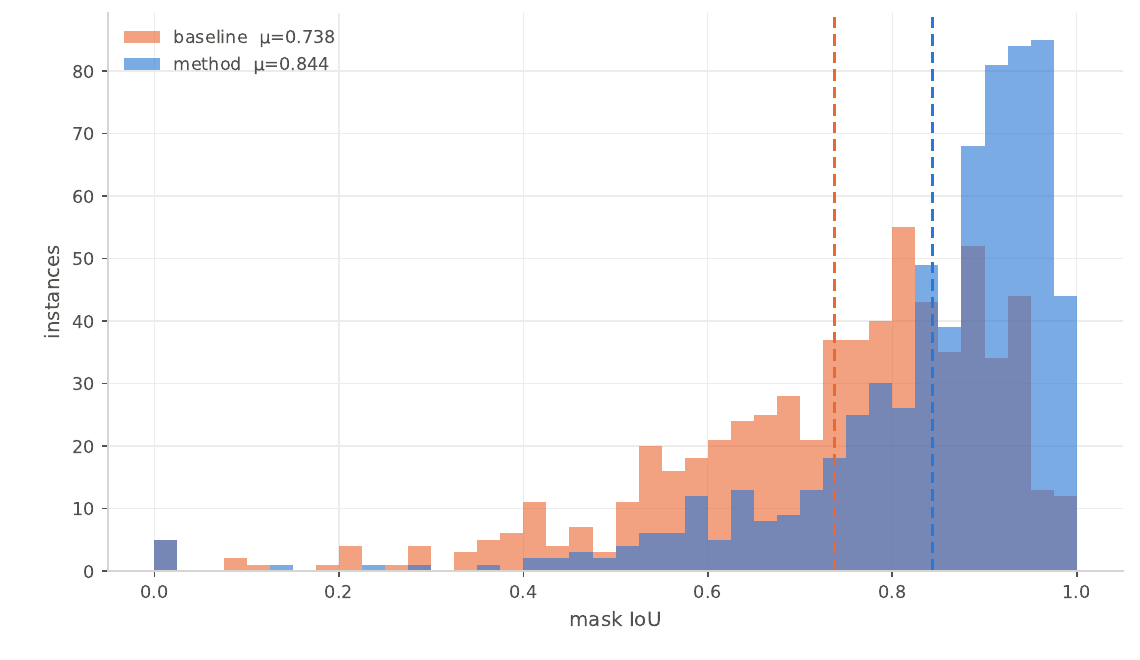}
        \caption{\revised{IoU histogram, ACDC-fog. $\mu$ 0.738 $\rightarrow$ 0.844.}}
        \label{fig:supp_iou_acdc_fog_hist}
    \end{subfigure}
    \hfill
    \begin{subfigure}[t]{0.356\linewidth}
        \includegraphics[width=\linewidth]{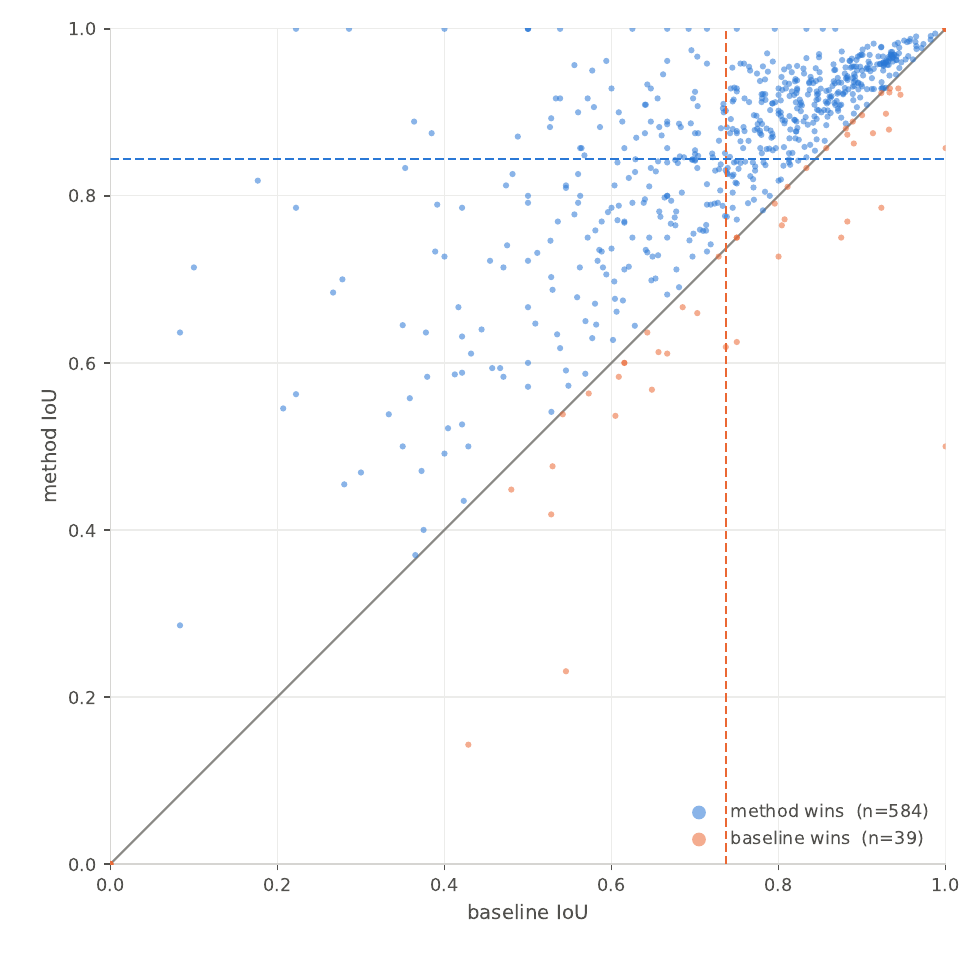}
        \caption{\revised{Per-instance scatter, ACDC-fog. \lca{} wins 584 of 623 (93.7\%).}}
        \label{fig:supp_iou_acdc_fog_scatter}
    \end{subfigure}
    \caption{\revised{Per-instance IoU on real ACDC fog, SAM~3 frozen against
    SAM~3\,+\,\lca{}+Dec.}}
    \label{fig:supp_iou_acdc_fog}
\end{figure*}

\begin{figure*}[tp]
    \centering
    \begin{subfigure}[t]{0.614\linewidth}
        \includegraphics[width=\linewidth]{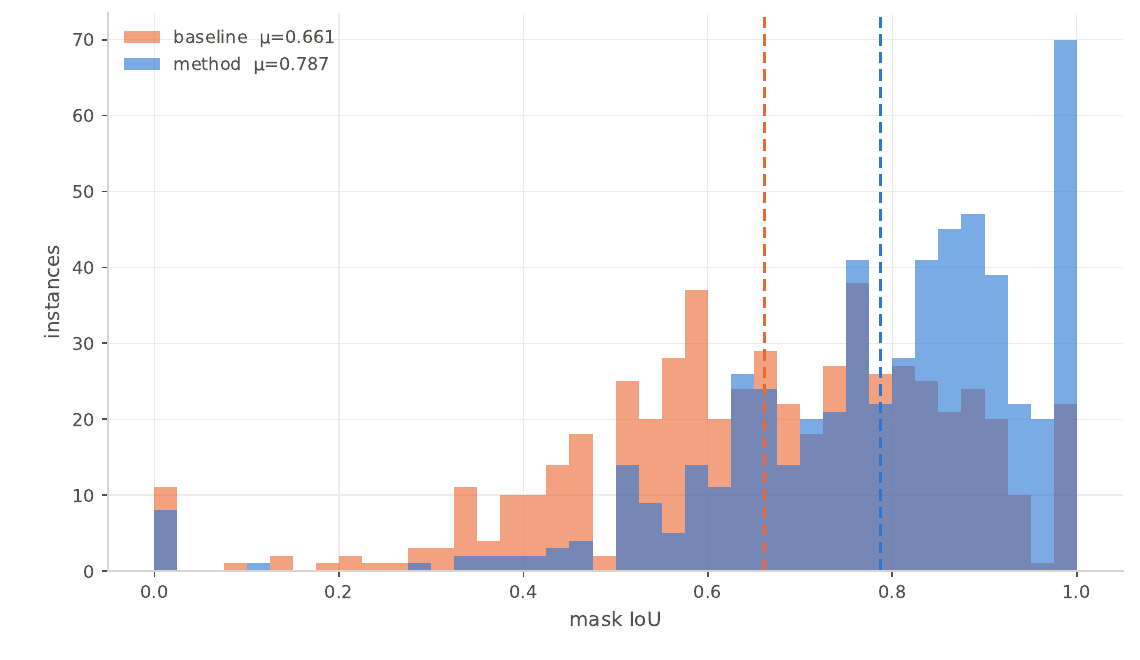}
        \caption{\revised{IoU histogram, ACDC-night. $\mu$ 0.661 $\rightarrow$ 0.787.}}
        \label{fig:supp_iou_acdc_night_hist}
    \end{subfigure}
    \hfill
    \begin{subfigure}[t]{0.356\linewidth}
        \includegraphics[width=\linewidth]{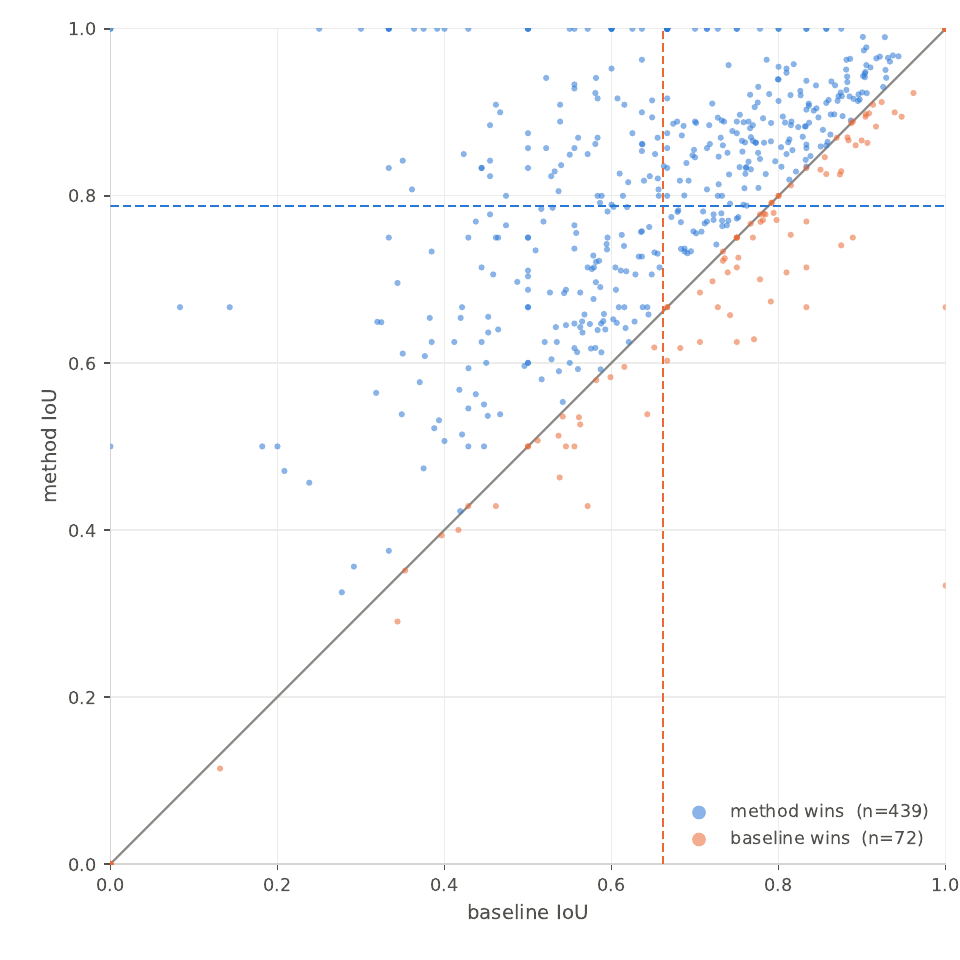}
        \caption{\revised{Per-instance scatter, ACDC-night. \lca{} wins 439 of 511 (85.9\%).}}
        \label{fig:supp_iou_acdc_night_scatter}
    \end{subfigure}
    \caption{\revised{Per-instance IoU on real ACDC night, SAM~3 frozen against
    SAM~3\,+\,\lca{}+Dec.}}
    \label{fig:supp_iou_acdc_night}
\end{figure*}

\begin{figure*}[tp]
    \centering
    \begin{subfigure}[t]{0.614\linewidth}
        \includegraphics[width=\linewidth]{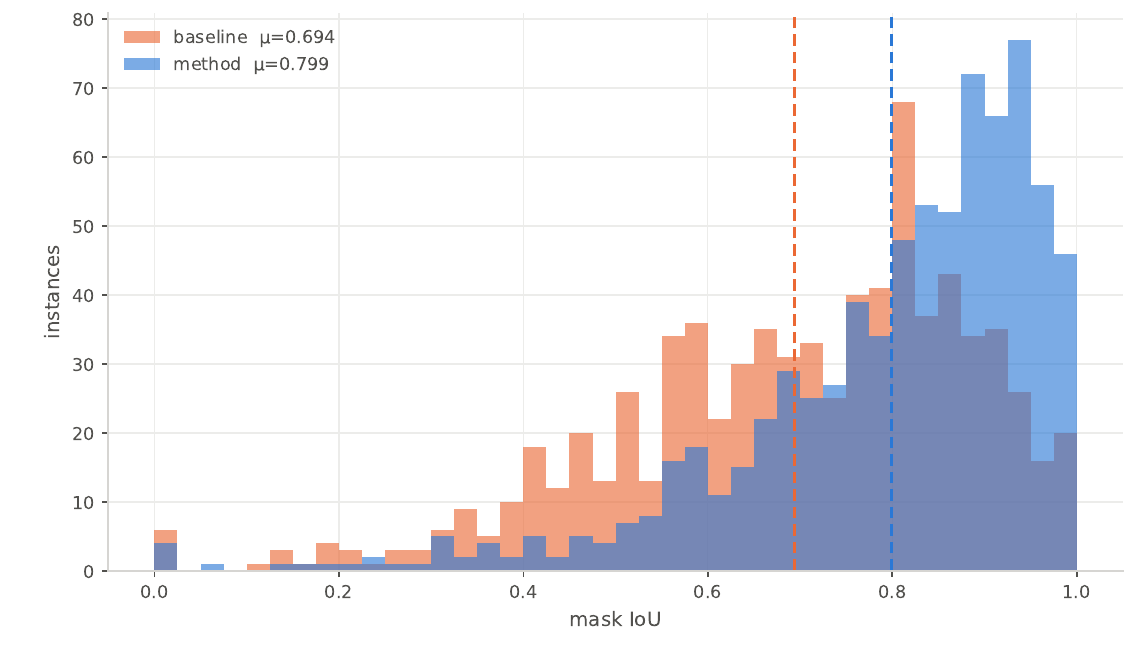}
        \caption{\revised{IoU histogram, ACDC-rain. $\mu$ 0.694 $\rightarrow$ 0.799.}}
        \label{fig:supp_iou_acdc_rain_hist}
    \end{subfigure}
    \hfill
    \begin{subfigure}[t]{0.356\linewidth}
        \includegraphics[width=\linewidth]{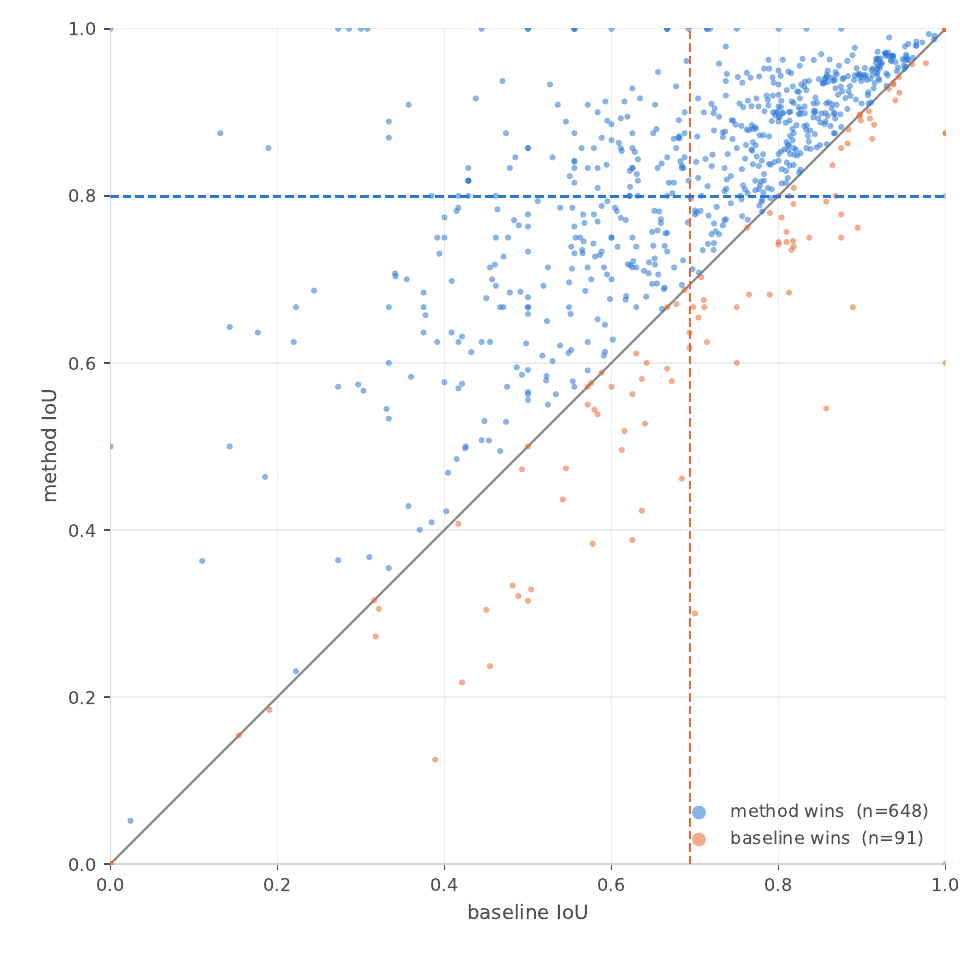}
        \caption{\revised{Per-instance scatter, ACDC-rain. \lca{} wins 648 of 739 (87.7\%).}}
        \label{fig:supp_iou_acdc_rain_scatter}
    \end{subfigure}
    \caption{\revised{Per-instance IoU on real ACDC rain, SAM~3 frozen against
    SAM~3\,+\,\lca{}+Dec.}}
    \label{fig:supp_iou_acdc_rain}
\end{figure*}

\begin{figure*}[tp]
    \centering
    \begin{subfigure}[t]{0.614\linewidth}
        \includegraphics[width=\linewidth]{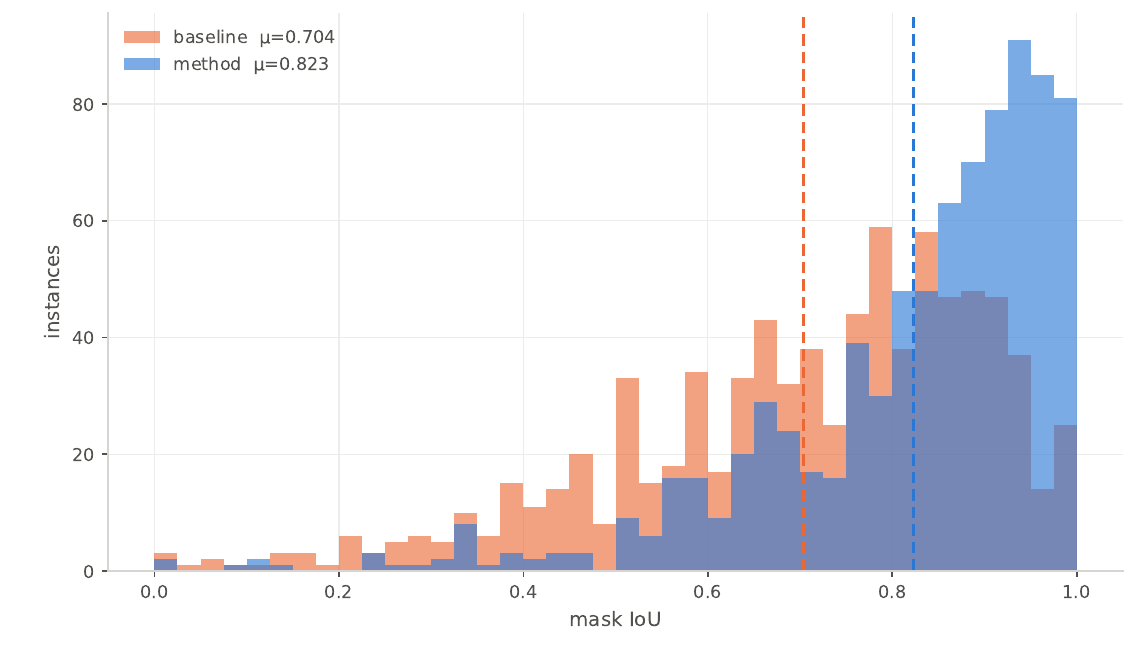}
        \caption{\revised{IoU histogram, ACDC-snow. $\mu$ 0.704 $\rightarrow$ 0.823.}}
        \label{fig:supp_iou_acdc_snow_hist}
    \end{subfigure}
    \hfill
    \begin{subfigure}[t]{0.356\linewidth}
        \includegraphics[width=\linewidth]{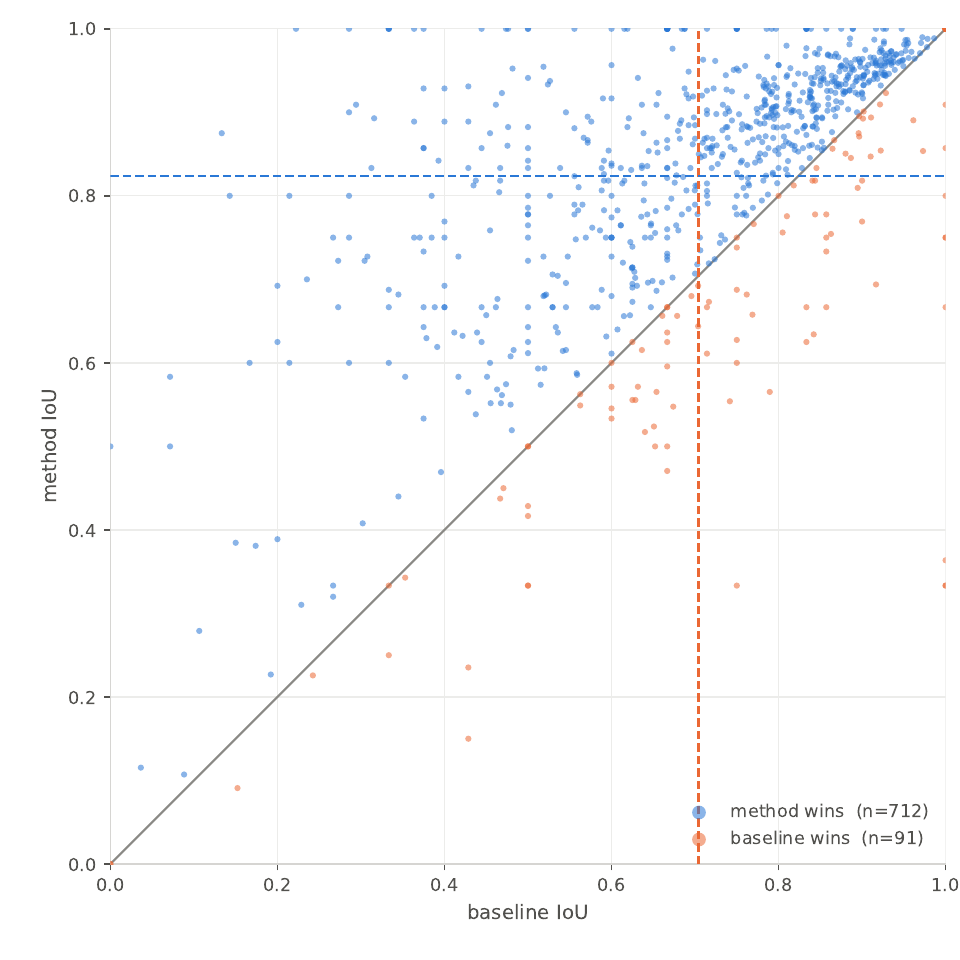}
        \caption{\revised{Per-instance scatter, ACDC-snow. \lca{} wins 712 of 803 (88.7\%).}}
        \label{fig:supp_iou_acdc_snow_scatter}
    \end{subfigure}
    \caption{\revised{Per-instance IoU on real ACDC snow, SAM~3 frozen against
    SAM~3\,+\,\lca{}+Dec.}}
    \label{fig:supp_iou_acdc_snow}
\end{figure*}

\subsubsection*{LCA vs.\ Baseline}
\label{sec:supp_highgain}

The examples shown Figures~\figref{fig:hg_city1}-\figref{fig:hg_city2} for cityscape dataset, Figures~\figref{fig:hg_coco1}-\figref{fig:hg_coco2} for coco dataset, and Figures~\figref{fig:hg_unity1}-\figref{fig:hg_unity2} for unity dataset where the baseline fails ($\text{IoU} < 0.13$) but \lca{} is able to recover a well-localized mask ($\text{IoU} > 0.72$). These cases cover all three evaluation domains. In each panel we show: (1) the input image with the ground-truth contour, (2) the baseline prediction, (3) the \lca{} prediction, and (4) a signed difference map (red marks false positives from the baseline that \lca{} removes, while blue shows regions recovered only by \lca{}). Looking across the examples, a similar behavior appears in all three domains. The baseline often produces confident predictions, but the activations spread beyond the object and bleed into the surrounding area. With \lca{}, these off-boundary activations are largely suppressed, and the mask becomes more focused around the object edges. The contrast gating helps emphasize these high-contrast boundaries and pull the prediction back toward the correct region. We see the same behavior whether the lighting degradation comes from image-space augmentations (COCO and Cityscapes) or from physically rendered lighting changes in Unity. This suggests that the Laplacian branch is not simply adapting to artifacts from a specific dataset, but instead learns a more general structural signal related to object boundaries.

\subsubsection*{Real-World Scenes: COCO and Cityscapes}

\begin{figure*}[tp]
    \centering
    \includegraphics[width=0.62\textwidth]{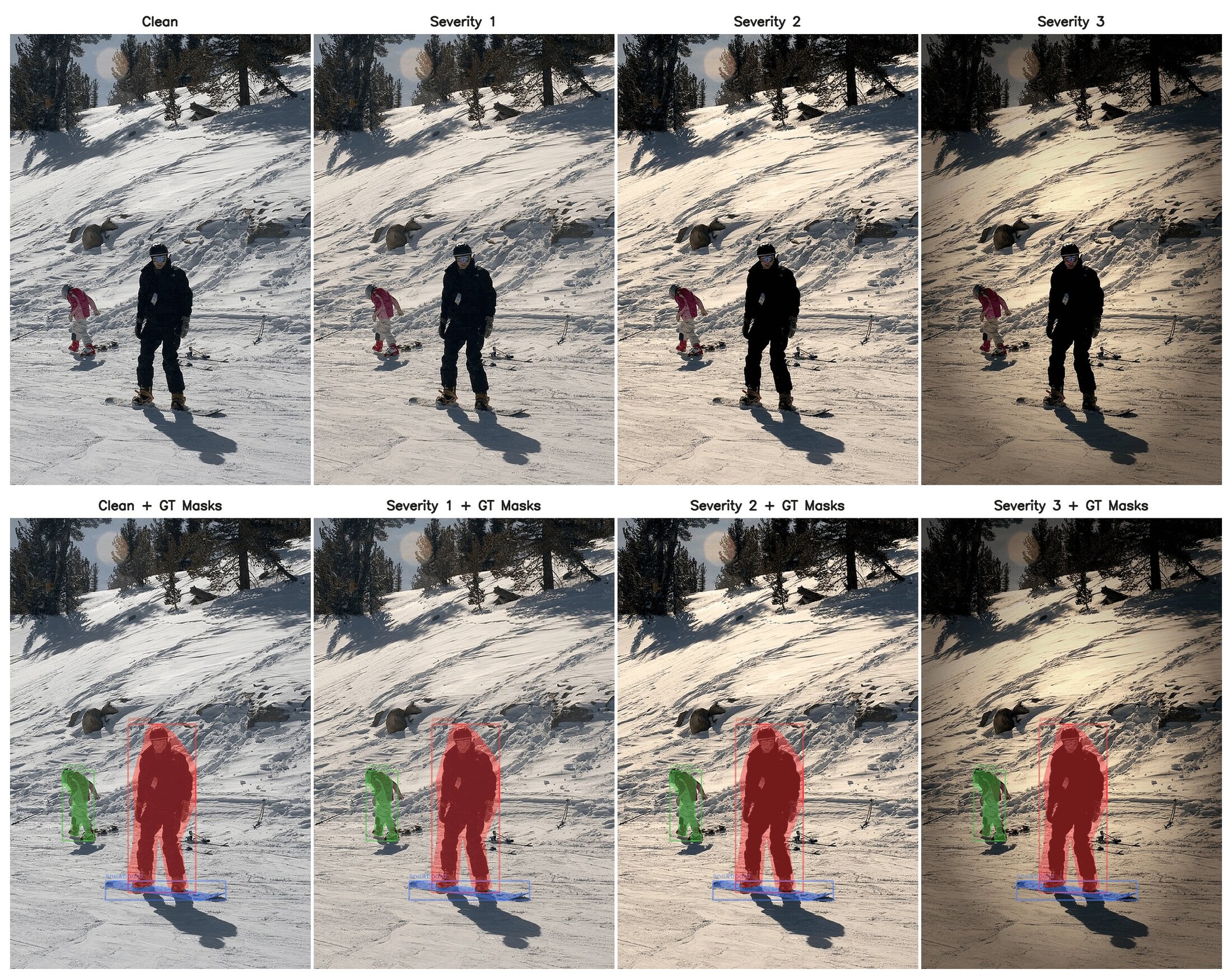}
    \caption{%
        COCO scene.
        Row 1: clean image and three severity levels (mild, moderate, severe).
        Row 2: corresponding ground-truth instance masks.%
    }
    \label{fig:supp_coco_1}
\end{figure*}

\begin{figure*}[tp]
    \centering
    \includegraphics[width=0.62\textwidth]{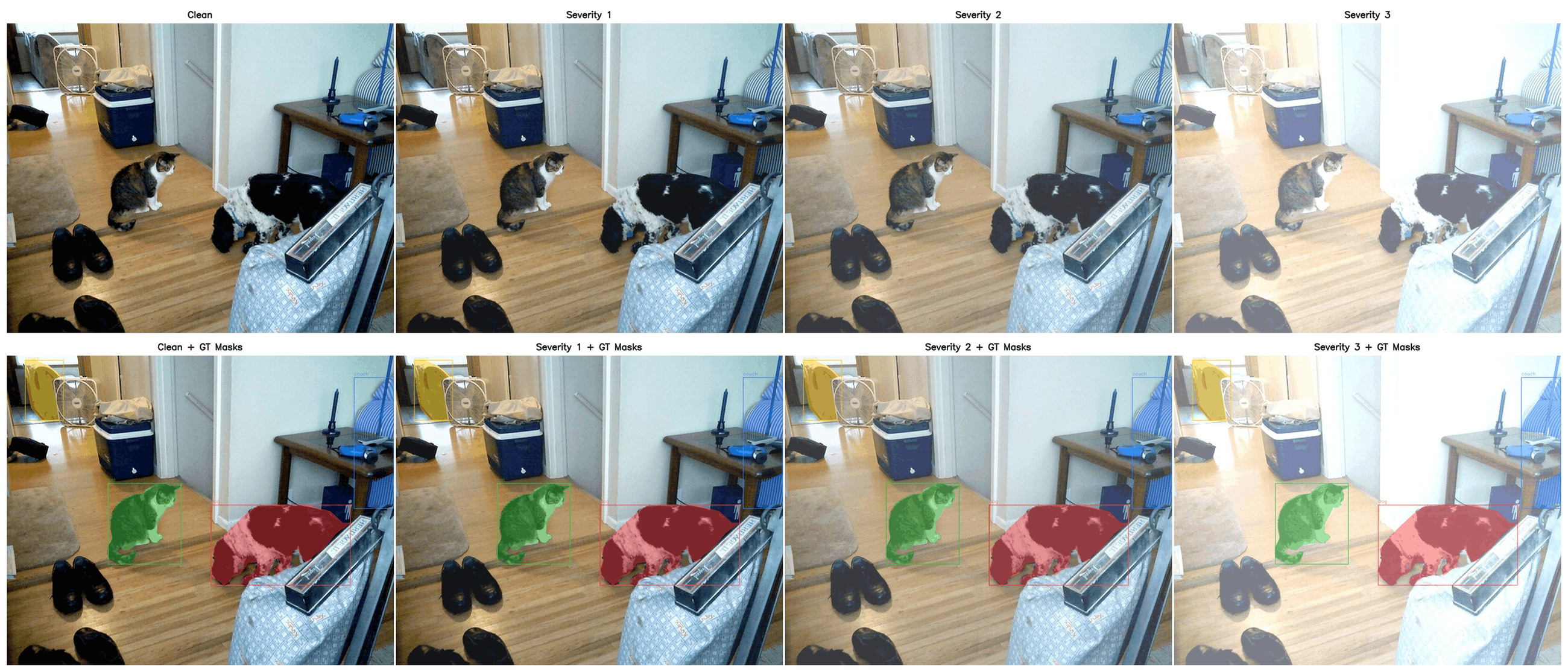}
    \caption{COCO scene (second example).}
    \label{fig:supp_coco_2}
\end{figure*}

\begin{figure*}[tp]
    \centering
    \includegraphics[width=0.62\textwidth]{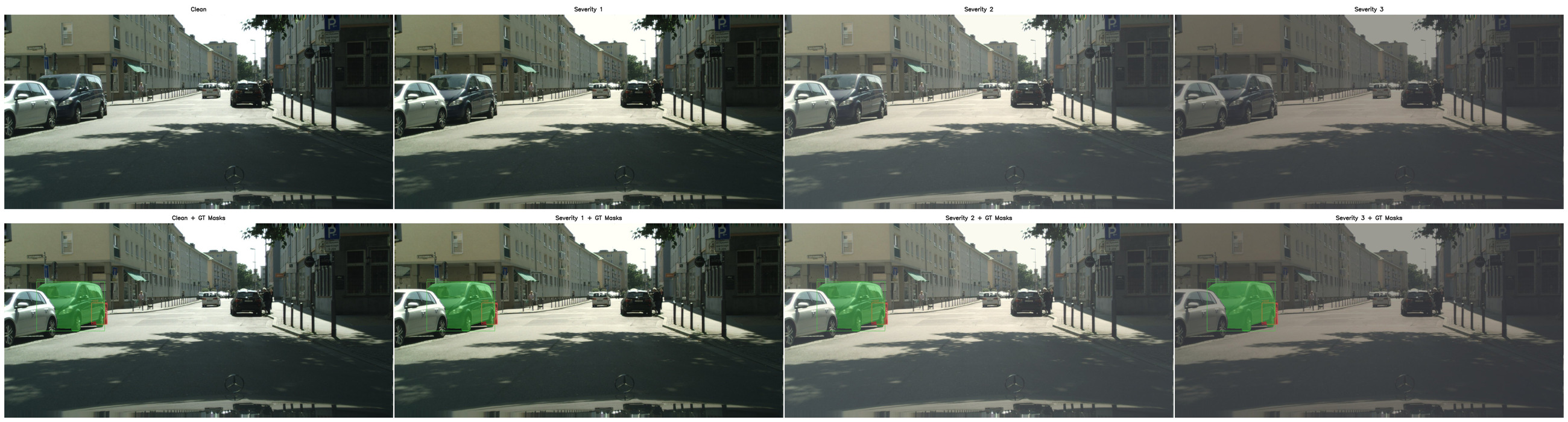}
    \caption{Cityscapes scene.}
    \label{fig:supp_city_1}
\end{figure*}

\begin{figure*}[tp]
    \centering
    \includegraphics[width=0.62\textwidth]{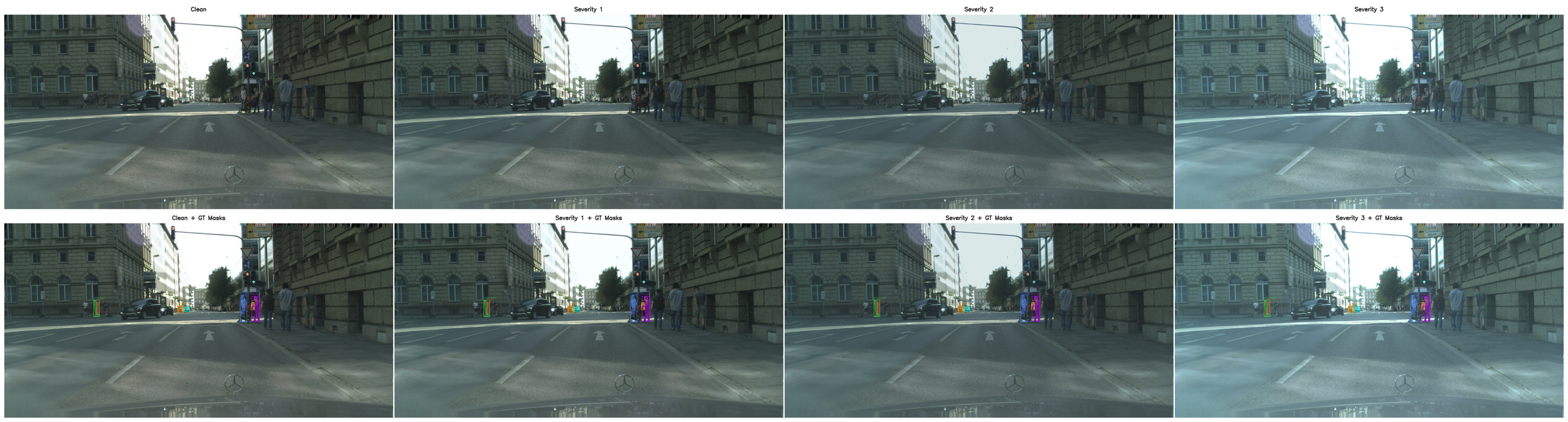}
    \caption{Cityscapes scene (second example).}
    \label{fig:supp_city_2}
\end{figure*}

Figures~\figref{fig:supp_coco_1}-\figref{fig:supp_coco_2} and Figures~\figref{fig:supp_city_1}-\figref{fig:supp_city_2} show examples 
drawn from COCO and Cityscapes using the \plap{} to generate the lighting variants. The original datasets reflect the diversity of real-world capture conditions background, heavy occlusion, and wide differences in instances. Lighting variants are produced by \plap{} applied to the original images. Across both datasets, as the severity keeps increasing, the degradation becomes most evident at instance boundaries and on small or low-contrast objects, which are exactly the scenarios our method is designed to handle.

\subsubsection*{Synthetic Scenes: Unity}

\begin{figure*}[tp]
    \centering
    \includegraphics[width=0.62\textwidth]{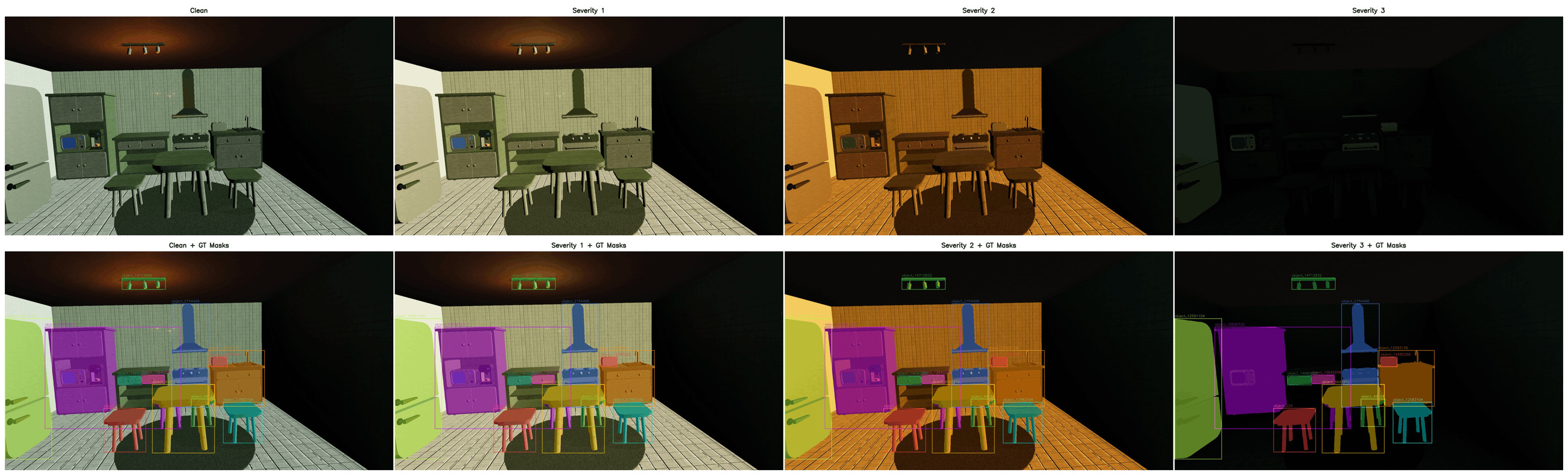}
    \caption{Unity scene.}
    \label{fig:supp_unity_1}
\end{figure*}

\begin{figure*}[tp]
    \centering
    \includegraphics[width=0.62\textwidth]{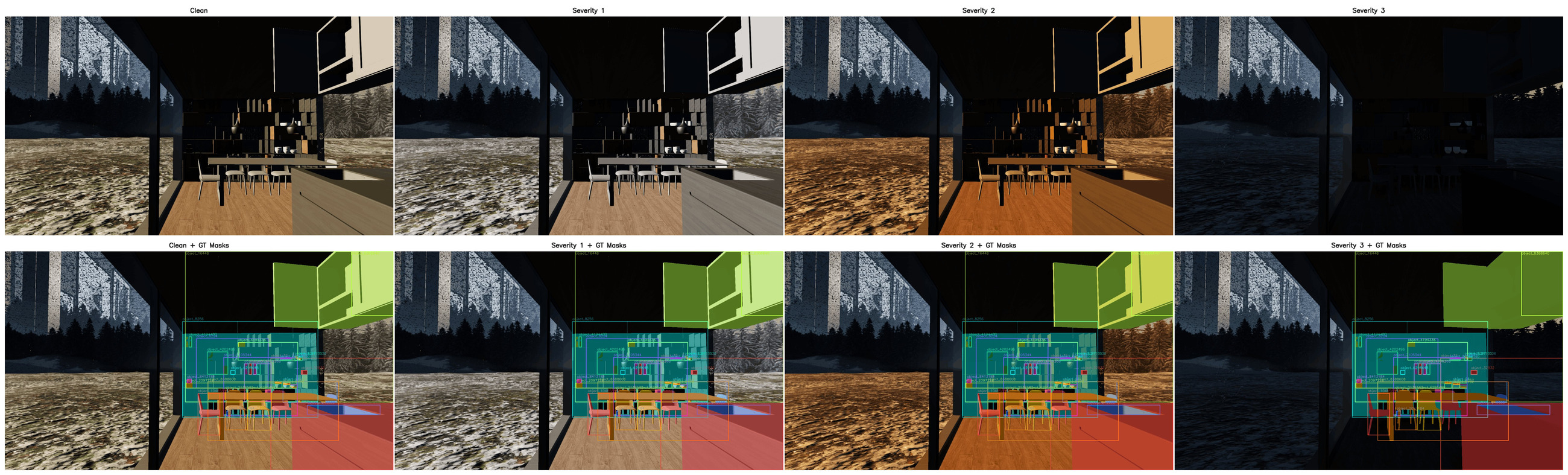}
    \caption{Unity scene (second example).}
    \label{fig:supp_unity_2}
\end{figure*}

Figures~\figref{fig:supp_unity_1}-\figref{fig:supp_unity_2} are the examples from our Unity synthetic dataset. The key distinction from the real-world benchmarks is that lighting here is not a image level operation. It is rendered physically by the Unity engine through our~\secref{unity_dataset} configuration across the three severity levels.  As a result, reflections, and surface materials all respond to the lighting change in a way that image space augmentation cannot replicate. Ground-truth instance masks are captured natively by the Unity 
Perception package at render time, which also gives perfect annotations. Together, these properties make the synthetic dataset a controlled but non-trivial complement to the real-world evaluation.

\end{document}